\documentclass{article}

\usepackage{yx}

\title{Towards Optimal Problem Dependent Generalization Error Bounds in Statistical Learning Theory}

\author{Yunbei Xu\footnote{Graduate School of Business, New York, NY; Email: \texttt{yunbei.xu@gsb.columbia.edu}.} \\ Columbia University
\and
Assaf Zeevi\footnote{ Graduate School of Business, New York, NY; Email: \texttt{assaf@gsb.columbia.edu}.} \\ Columbia University
}

\date{}

\begin{document}

\maketitle

\begin{abstract}
  We study problem-dependent rates, i.e.,  generalization errors that scale near-optimally with the variance, the effective loss, or the gradient norms evaluated at the ``best hypothesis."  We introduce a principled framework dubbed ``uniform localized convergence," and characterize sharp problem-dependent rates for central statistical learning problems.
   From a methodological viewpoint, our framework resolves several fundamental limitations of existing uniform convergence and localization analysis approaches. It also provides improvements and some level of  unification in the study  of localized complexities, one-sided uniform inequalities, and sample-based iterative algorithms. 
    In the so-called ``slow rate" regime, we  provides the first (moment-penalized) estimator that  achieves the optimal variance-dependent rate for general ``rich" classes; we also establish improved loss-dependent rate for standard empirical risk minimization. In the  ``fast rate" regime, we establish finite-sample problem-dependent bounds that are comparable to precise asymptotics. In addition, we show that iterative algorithms like gradient descent  and  first-order Expectation-Maximization  can achieve optimal generalization error in several representative problems across the areas of non-convex learning, stochastic optimization, and learning with missing data.

\end{abstract}

\setcounter{tocdepth}{2}
\tableofcontents

\newpage
\section{Introduction}\label{sec intro}
\subsection{Background}
\paragraph{Problem statement.} Consider the following statistical learning setting. Assume that a random sample $z$ follows an unknown distribution $\P$ with support $\Z$. For each realization of $z$, let $\ell(\cdot;z)$ be a real-valued {\it loss function}, defined over the {\it hypothesis class} $\Hy$. Let $h^*\in\Hy$ be the optimal hypothesis that minimizes the \textit{population risk} 
    \begin{align*}
 \mathbb{P}\ell(h;z) := {\mathbb{E}} [\ell(h; z)].
\end{align*}
 Given $n$ i.i.d. samples $\{z_i\}_{i=1}^n$ drawn from $\P$, our goal, roughly speaking, is to ``learn" a hypothesis $\hh\in \Hy$ that makes the  {\it generalization error} 
\begin{align*}
    \mathcal{E}(\hh):=\P\ell(\hh;z)-\P\ell(h^*;z)
\end{align*} as small as possible. This pursuit is ubiquitous in machine learning, statistics and stochastic optimization.

We study {\it problem-dependent rates}, i.e., finite-sample generalization errors that scale tightly with  problem-dependent parameters, such as  the variance, the effective loss, or the gradient norms at the optimal hypothesis $h^*$.  While the direct dependence of $\mathcal{E}(\hh)$ on the sample size $n$ is often well-understood, it typically only reflects an ``asymptotic" perspective, placing less emphasis on the scale of problem-dependent parameters. Existing literature leaves several outstanding challenges in deriving problem-dependent rates. These can be broadly categorized into the  so-called ``slow rate" and ``fast rate" regimes, as described below.

\paragraph{Main challenges in the ``slow rate" regime.} 
In the absence of strong convexity and margin conditions, the direct dependence of 
$\mathcal{E}(\hh)$ on the sample size $(n)$  is typically no faster than $O(n^{-\frac{1}{2}})$. These settings are  referred to as the ``slow rate" regime. Here, the central challenge is to simultaneously achieve optimal dependence on  problem-dependent parameters  (the variance or the effective loss at the optimal $h^*$) and the sample size ($n$). To the best of our knowledge, this has not been achieved in previous literature for ``rich" classes (to be explained shortly).

Perhaps the most widely used framework to study problem-dependent rates is the traditional ``local Rademacher complexity" analysis \cite{bartlett2005local, koltchinskii2000rademacher, wainwright2019high}, which has become a standard tool in learning theory. However, as we will discuss later, this  analysis leads to a dependence on the sample size ($n$) which is sub-optimal for essentially all ``rich" classes (with the exception of parametric classes). The absence of more precise localization analysis also challenges the design of more refined estimation procedures. For example, designing estimators to achieve variance-dependent rates  requires penalizing the empirical second moment to achieve the ``right" bias-variance trade-off. Most antecedent work is predicated on either the traditional ``local Rademacher complexity" analysis \cite{namkoong2017variance, foster2019orthogonal} or coarser approaches \cite{maurer2009empirical, swaminathan2015counterfactual}. Thus, to the best of our knowledge, the question of optimal problem-dependent rates for general rich classes is still open.

\paragraph{Main challenges in the ``fast rate" regime.}
When assuming suitable curvature or so-called margin conditions,  the direct dependence of $\Eh$ on $n$ is typically faster than $O(n^{-\frac{1}{2}})$; for that reason  we refer to this as the ``fast rate" regime. Here, the traditional localization analysis  often provides correct dependence on the sample size ($n$), but the complexity parameters (e.g., norms of gradients, Lipchitz constants, etc.) characterizing these generalization errors are not localized and hence are not problem-dependent. 

Much progress on problem-dependent rates has been made under particular formulations, such as regression with structured strongly convex cost (e.g., square cost, Huber cost) 
\cite{mendelson2014learning, mendelson2018learning, liang2015learning}, and binary classification under margin conditions \cite{gine2006concentration, zhivotovskiy2016localization}
. In particular, \cite{mendelson2014learning, mendelson2018learning} 
point out that the traditional ``local Rademacher complexity" analysis fails to provide parameter localization {for unbounded regression problems}, and propose the so-called ``learning without concentration" methodology to obtain problem-dependent rates that do not scale with the worst-case parameters. 
We are able to {recover these} results using a more direct concentration-based analysis and remove certain restrictions.

We will also focus on parametric models in the ``fast rate" regime, which covers more ``modern" non-convex learning problems { whose analysis is not aligned with traditional generalization error analysis. 
For example, in non-convex learning problems one wants generalization error bounds for iterative optimization algorithms; and traditional localization frameworks (which mostly focus on supervised learning) do not cover general stochastic optimization and missing data problems. For many representative  non-convex learning, stochastic optimization and missing data problems,   it remains open to provide algorithmic solutions and  problem-dependent generalization error bounds that are comparable to  optimal asymptotic results.}

\subsection{Contributions and organization of the paper}
 The paper provides contributions both in the framework it develops, as well as its application to improving existing results in several problem areas. In particular, it suggests guidelines for designing estimators and learning algorithms and provides analysis tools to study them. In addition, it provides some level of unification across problem areas. Specifically, the main contributions are as follows.
 
 (1) We introduce  a principled framework to study localization in statistical learning, dubbed  the ``uniform localized convergence" procedure, which  simultaneously provides optimal ``direct dependence" on the sample size, correct scaling with problem-dependent parameters, { and flexibility to unify various problem settings.  Section \ref{sec theoretical foundations} provides a  description of the framework, and explains how it addresses several fundamental challenges. }

(2) In the ``slow rate" regime,   we employ the above ideas to design the first estimator that achieves optimal variance-dependent rates for general function classes. The derivation is based on a novel  procedure that {optimally penalizes the empirical (centered) second moment}. We also establish improved loss-dependent rates for standard empirical risk minimization, which has computational advantages. Section \ref{sec slow rate regime}   presents these theoretical results (see Section \ref{sec general loss without curvature} for the loss-dependent rate and Section \ref{sec empirical penalization} for the variance-dependent rate); and Section \ref{sec application} will illustrate our improvements in non-parametric classes  and VC classes.

(3) In the ``fast rate" regime, we establish a ``uniform localized convergence of gradient" framework for parametrized models, and characterize optimal problem-dependent rates for approximate stationary points and iterative optimization algorithms such as gradient descent and first-order Expectation-Maximization. Our results scale tightly with the gradient norms at the optimal parameter, and improve previous guarantees in  non-convex learning, stochastic optimization  and missing data problems. 
See Section \ref{sec parametric} for the theoretical results and Sections \ref{sec application fast rate regime}-\ref{sec em} for illustrations of improvements over previous results. 

(4) In the ``fast rate" regime, we also study supervised learning with  structured convex cost, where the hypothesis class can be  non-parametric and heavy-tailed. This part of the work has direct relationship to a stream of literature known as ``learning without concentration."   Our contributions in this setting lie in technical refinements of the generalization error bounds and some unification between one-sided uniform inequalities and concentration of truncated functions; for this reason we defer its treatment  to Section \ref{sec supervised learning strongly convex}.

\section{The ``uniform localized convergence" procedure}\label{sec theoretical foundations}

\subsection{The current blueprint}  \label{subsec current blueprint}
Denote the {\it empirical risk} 
\begin{align*}\mathbb{P}_n \ell(h;z):=\frac{1}{n}\sum_{i=1}^n \ell(h; z_i),\end{align*}
and consider the following straightforward decomposition of the generalization error
\begin{align}\label{eq: decomposition}
   \mathcal{E}(\hh)= (\P-\Pn)\ell(\hh;z)+\big(\Pn\ell(\hh;z)-\Pn\ell(h^*;z)\big)+(\Pn-\P)\ell(h^*;z).
\end{align} The main difficulty in studying  $\mathcal{E}(\hh)$ comes from bounding the first term $(\P-\Pn)\ell(\hh;z)$, since  $\hh$ depends on the $n$ samples. 
The simplest approach, which does not achieve  problem-dependent rates, is to bound the {uniform error} $$
    \sup_{h\in\Hy}(\P-\Pn)\ell(h;z)
$$ over the {\it entire} hypothesis class $\Hy$.
In order to obtain problem-dependent rates, a natural modification is to consider uniform convergence  over   {\it localized} subsets of $\Hy$. 

We first give an overview of the traditional ``local Rademacher complexity" analysis \cite{bartlett2005local, koltchinskii2011oracle, wainwright2019high}. 
 Consider a generic function class $\F$ that we wish to concentrate, which consists of real-valued functions defined on $\pazocal{Z}$  (e.g., one can set $f(z)=\ell(h;z)$). Denote $$\P f:=\E[f(z)],\quad \Pn f:=\frac{1}{n}\sum_{i=1}^n f(z_i).$$  The notation $r>0$ will serve as a localization parameter, and $\delta>0$ will serve for high probability arguments. Let $\psi(r;\delta)$ denote a surrogate function that  upper bounds the uniform error within a localized region $\{f\in\F: T(f)\leq r\}$, where we call $T:\F\rightarrow \R_{+}$ the ``measurement functional." Formally, let $\psi$ be a function that maps $[0,\infty)\times (0,1)$ to $[0,\infty)$, which possibly depends on the observed samples $\{z_i\}_{i=1}^n$. Assume $\psi$ satisfies   for arbitrary fixed $\delta,r$, with probability at least $1-\delta$,
\begin{align}\label{eq: localized uniform convergence}
    \sup_{f\in\F:T(f)\leq r}(\P-\Pn)f\leq \psi(r;\delta).
\end{align}
 By default we ask $\psi(r;\delta)$ to be a non-decreasing and non-negative function.\footnote{Here and in what follows we will assume that such suprema are measurable, namely, the required regularity conditions on the underlying function classes are met (see, e.g., [\citealp{pollard2012convergence}, Appendix C], [\citealp{van1996weak}, Section 1.7]).} The main result of the traditional localization analysis can be {summarized} as follows. {(The statement is obtained by adapting the proof from [\citealp{bartlett2005local}, Section 3.2]; itself more general than the traditional meta-results \cite{bartlett2005local, koltchinskii2011oracle, wainwright2019high}.)}

\begin{statement}[{\bf current blueprint}]\label{state current blueprint}
Assume that $\psi$ is a sub-root function, i.e., $\psi(r;\delta)/\sqrt{r}$ is non-increasing  with respect to $r\in\R_{+}$.  Assume the Bernstein condition $T(f)\leq B_e\P f$, {$B_e>0$, } $\forall f\in\F$.  Then with probability at least $1-\delta$, for all $f\in\F$ and $K>1$,
 \begin{align}\label{eq: local rademacher complexity}
     (\P-\Pn) f\leq  \frac{1}{K}\P f+ \frac{100(K-1) r^*}{B_e},
 \end{align}
 where $r^*$ is the ``fixed point" solution of the equation $r=B_e\psi(r;\delta)$.
\end{statement}
    Since its inception, Statement \ref{state current blueprint} has become a standard tool in learning theory. However, it requires a rather technical proof, and it appears to be loose when compared {with} the original assumption \eqref{eq: localized uniform convergence}---ideally, we would like to directly extend  \eqref{eq: localized uniform convergence} to hold uniformly without sacrificing any accuracy. Moreover, some assumptions in the statement are restrictive and might not be necessary.

\subsection{The  ``uniform localized convergence" principle}\label{subsec refined analysis}

We provide a surprisingly simple analysis that greatly improves and simplifies Statement \ref{state current blueprint}. 
 {Unlike Statement \ref{state current blueprint}, the following proposition does not require the concentrated functions $g_f$ to satisfy the Bernstein condition, and the surrogate function $\psi$ need not to be ``sub-root." Despite placing less restrictions, Proposition \ref{prop peeling} is able to establish results that are typically ``sharper" than the current blueprint.} Note that in the proposition, both the measurement functional $T$ as well as the surrogate function $\psi$ are allowed to be sample-dependent.

\begin{proposition}[\bf the ``uniform localized convergence" argument]\label{prop peeling}
For a function class  $\G=\{g_f:f\in\F\}$ and   functional $T:\F\rightarrow [0,R]$, assume there is a function  $\psi(r;\delta)$, which is non-decreasing with respect to $r$ and satisfies that $\forall \delta\in(0,1)$, $\forall r\in[0,R]$, with probability at least $1-\delta$,
\begin{align}\label{eq: surrogate}
\sup_{f\in\F: T(f)\leq r}(\P-\Pn)g_f\leq \psi(r;\delta).
\end{align}
 Then, given any $\delta\in(0,1)$ and $r_0\in (0,R]$, with probability at least $1-\delta$,  for all $f\in \F$, either $T(f)\leq {r_0}$ or
\begin{align}\label{eq: peeling}
    (\P-\Pn) g_f\leq  \psi\left(2T(f);\delta\left({\log_2\frac{2R}{r_0}}\right)^{-1}\right).
\end{align}
\end{proposition}

  The key intuition behind Proposition \ref{prop peeling} is that the uniform restatement of the ``localized" argument \eqref{eq: surrogate} is nearly cost-free, because the deviations $(\P-\Pn)g_f$  can be controlled solely by the real valued functional $T(f)$.  As a result, we essentially only require uniform convergence over an interval $[r_0,R]$. The ``cost" of this uniform convergence, namely, the additional $\log_2(\frac{2R}{r_0})$ term in   \eqref{eq: peeling}, will only appear in the form $\log(\delta/\log_2(\frac{2R}{r_0}))$ in high-probability bounds, which is of a negligible $O(\log\log n)$ order  in general.
 
 Formally, we apply a ``peeling" technique: we take $r_k=2^k r_0$, where $k=1,2,\dots, \lceil\log_2\frac{R}{r_0}\rceil$, and we use the union bound to extend \eqref{eq: surrogate} to hold for all these $r_k$. Then for any $f\in\F$ such that $T(f)> r_0$ is true, there exists a non-negative integer $k$  such that $2^k r_0<T(f)\leq 2^{k+1}r_0$. By the  non-decreasing property of the $\psi$ function, we then have
 $$(\P-\Pn)g_f\leq \psi\left(r_{k+1};\delta\left({\log_2\frac{2R}{r_0}}\right)^{-1}\right)\leq \psi\left(2T(f);\delta\left({\log_2\frac{2R}{r_0}}\right)^{-1}\right),$$ which is exactly \eqref{eq: peeling}. Interestingly, the proof of the classical result (Statement \ref{state current blueprint}) relies on a relatively heavy machinery that includes more complicated peeling and re-weighting arguments (see [\citealp{bartlett2005local}, Section 3.1.4]). However,  that analysis obscures the key intuition that we elucidate under inequality \eqref{eq: peeling}.

In this paper, {we prove localized generalization error bounds through a unified principle,} summarized at a high level in the two-step template below.
 
\noindent \textbf{Principle of uniform localized convergence.} {\it First, determine the concentrated functions, the measurement functional and the surrogate $\psi$, and obtain a sharp ``uniform localized convergence" argument. Then,  perform  localization analysis that is customized to the  problem setting and the learning algorithm.}

Distinct from the blueprint \eqref{eq: local rademacher complexity}, the right hand side of our ``uniform localized convergence" argument \eqref{eq: peeling} contains a ``free" variable $T(f)$ rather than a fixed value $r^*$.  {The new principle strictly improves the current blueprint from many aspects, and its merits will be illustrated in the sequel.}

\subsection{Merits of ``uniform localized convergence"}\label{subsec merits}
Our improvements in the ``slow rate" regime  originate from the noticeable gap between  Proposition \ref{prop peeling} and  Statement \ref{state current blueprint}, illustrated by the following (informal) conclusion.  
\begin{statement}[{\bf improvements over the current blueprint--informal statement}]\label{state improvement}
Setting $g_f=f$, then under the assumptions of Statement \ref{state current blueprint}, Proposition \ref{prop peeling} provides a strict improvement over Statement \ref{state current blueprint}. In particular,  the slower $\psi$ grows, the larger the gap between the bounds in the two results, and  the bounds are on pair only when $\psi$ is proportional to $\sqrt{r}$, i.e., when the function class $\F$ is parametric and not ``rich." 
\end{statement}
Formalizing as well as providing rigorous justification for this conclusion is relatively straightforward:  taking the ``optimal choice" of $K$ in Statement \ref{state current blueprint}, we can re-write its conclusion as 
\begin{align*}
    (\P-\Pn)f \leq 20\sqrt{\frac{r^*\P f}{B_e}}-\frac{r^*}{B_e} \quad  [\text{Statement \ref{state current blueprint}}],
\end{align*} where the right hand side is  of order $\sqrt{{r^*\P f}/{B_e}}$ when $\P f< r^*/B_e$, and order $r^*/B_e$ when  $\P f \leq r^*/B_e$.  Our result \eqref{eq: peeling} is also of order $r^*/B_e$ when  $\P f\leq r^*/B_e$. However, for every $f$ such that $\P f> r^*/B_e$, it is straightforward to verify that under the assumptions in Statetment \ref{state current blueprint},
\begin{align}\label{eq: strict improvement}
    \psi(2T(f);\delta)\quad &{\leq}\quad \psi(2B_e\P f;\delta)\quad \text{[Bernstein condition: $T(f)\leq B_e \P f$]} \nonumber\\
    &\leq\quad  \frac{\sqrt{2B_e\P f}}{\sqrt{r^*}}\psi(r^*;\delta) \quad  \text{[$\psi(r;\delta)$ is sub-root]} \nonumber\\ &{\leq}\quad \sqrt{\frac{2 {r^*}\P f}{B_e}} \quad \text{[$r^*$ is the fixed point of $B\psi(r;\delta)$]}.
\end{align}
 Therefore, the argument $\psi(2T(f);\delta)\leq \sqrt{{2r^*\P f}/{B_e}}$ established by \eqref{eq: strict improvement} shows that the ``uniform localized convergence" argument \eqref{eq: peeling}  strictly improves over  Statement \ref{state current blueprint} (ignoring  negligible $O(\log\log n)$ factors). 
 {
 Statement \ref{state improvement} indicates that the folklore use of  fixed values as straightforward complexity control is somewhat questionable.  In the ``slow rate" regime, the key point to achieve optimal problem-dependent rates is to bound the generalization error using the function $\psi$. Otherwise the bounds will have the ``wrong" dependence on the sample size for all ``rich" classes.}

Interestingly, the merits of our approach in the ``fast rate" regime stem from   very different perspectives:  the removal of the ``sub-root" requirement on $\psi$ allows one to achieve parameter localization; and  the added flexibility in the choice of $g_f$ allows one to prove one-sided uniform inequalities and uniform convergence of gradient vectors. 
To better appreciate these, we provide the following informal discussion to help elucidate the key messages. Let $\mu>0$ be the curvature parameter in the ``fast rate" regime (a common example is the strong convexity parameter for the loss function).    The generalization error is often characterized by the fixed point of $\psi(r;\delta)/\mu$, where $\psi$ is the surrogate function that governs the uniform error of excess risk. The removal of the ``sub-root" restriction is crucial because under curvature and smoothness conditions, the uniform error of excess loss typically grows ``faster" than the square root function. Consider  surrogate functions of the form 
\begin{align}\label{eq: decomposition super root}
    \psi(r;\delta)=\underbrace{\sqrt{a_n^* r}}_{\text{problem-dependent}}+\underbrace{c_nr}_{\text{super-root}},
\end{align} where $a_n^*$ is a problem-dependent rate, and $c_n$ satisfies $0<c_n<1/(2\mu)$ when the sample size $n$ satisfies mild conditions. We call $c_n r$ the benign ``super-root" component in the sense that when solving the fixed point equation $ r=\psi(r;\delta)/\mu$, that part can be dropped from both side of the equation. As a result, the fixed point solution only depends on order of the problem-dependent component  $\sqrt{a_n^* r}$, and so one obtains problem-dependent rates. In contrast, worst-case parameters will be unavoidable if one wants to use a ``sub-root" surrogate function to govern \eqref{eq: decomposition super root}. In traditional analysis, a loose ``sub-root" surrogate function is often obtained via two-sided concentration and Lipchitz contraction, making global Lipchitz constants unavoidable.  Furthermore, the added flexibility to choose concentrated functions $g_f$ is also useful. In particular, we will show that: 1) our framework unifies traditional value-based uniform convergence and uniform convergence of gradient vectors, which is crucial to study non-convex learning problems and sample-based iterative algorithms; and 2) simple ``truncated" functions can be used to established one-sided uniform inequalities that are sharper than two-sided ones, which enable recovery of results in unbounded and heavy-tailed regression problems.

\subsection{Unification and improvements over existing localization approaches}\label{subsec history}

Beyond proving new bounds, an important objective of the paper is to  provide some level of unification to the methodology of uniform convergence and localization. Here we present a historical review of uniform convergence and localization, and discuss how the ``uniform localized convergence" principle unifies and improves existing approaches.  We will overview four general settings where localization plays a crucial role in generalization error analysis and algorithm design:  1) the ``slow rate" regime; 2)  classification under margin conditions; 3) regression under curvature conditions; and 4) non-convex learning and  stochastic optimization (the latter three settings belong to the ``fast rate" regime).

\paragraph{The ``slow rate" regime.} The traditional ``local Rademacher complexity" analysis is the standard tool to study localized generalization error bounds in the ``slow rate" regime; it also influences the design of regularization. Here, our framework resolves the fundamental limitation of the traditional analysis, leading to the first optimal loss-dependent and variance-dependent rates for general ``rich" classes (see Section \ref{sec slow rate regime}-\ref{sec application}).

 \paragraph{Classification under margin conditions.}
  One early line of work in the ``fast rate" regime focuses on exploiting the margin conditions to establish fast rates in binary classification (e.g., see [\citealp{koltchinskii2011oracle}, Section 5.3]). It can be shown that the Bernstein condition in Statement \ref{state current blueprint} subsumes standard margin conditions such as Tsybakov's margin condition \cite{tsybakov2003optimal} and Massart's noise condition \cite{massart2006risk}. {Because the loss functions in binary classification are uniformly bounded, the ``current blueprint" (Statement \ref{state current blueprint}) already provides a good framework to study these questions.
  In an orthogonal direction, some recent works study more refined alternatives to the notion ``local Rademacher complexity." While these results are important, they are within the scope of Statement \ref{state current blueprint} from the perspective of localization machinery. For example, one can consider better ``sub-root" relaxation than direct Rademacher symmetrization \cite{zhivotovskiy2016localization}; and one may only need to consider some subset of $\F$ if using ``empirical risk minimization over epsilon net" \cite{zhivotovskiy2017optimal}.} In this setting, the work presented in this paper contributes to unification  rather than improvements  over specific theoretical results.

  \paragraph{Regression under curvature conditions. } The traditional localization analysis has been widely applied to curvatured regression problems to achieve ``fast rates." However, it fails to localize the complexity parameters (e.g., norms of gradients, Lipchitz constants, etc.) in the generalization error bounds. As a consequence, the traditional localization analysis is widely recognized as not being suitable for regression problems with unbounded losses; and it may be  unfavorable even for uniformly bounded problems because it fails to adapt to the ``effective noise level" at the optimal hypothesis. 
Important progress addressing the aforementioned limitations has been made during the last decade. Focusing on supervised learning problems with structured convex costs (square cost, Huber cost, etc.),  the breakthrough works \cite{mendelson2014learning, mendelson2018learning} propose the so-called ``learning without concentration" framework, where the central notions and proof techniques are quite different from the traditional concentration framework. An interesting direction is to recover these results 
through a “one-shot” concentration framework. By exploiting the intrinsic connections between one-sided uniform inequalities and the ``uniform localized convergence" of truncated functions,  we are able to  establish such a  unified  analysis and illustrate systematical refinements of the generalization error bounds; This setting will be discussed in Section \ref{sec supervised learning strongly convex}.

\paragraph{Non-convex learning and stochastic optimization.} Traditional value-based generalization error analysis relies on properties of global (regularized) empirical risk minimizers. However, non-convex learning problems and generalization error analysis of iterative optimization algorithms require one to consider uniform convergence of gradient vectors. And it should be noted that existing localization frameworks mostly  focus on supervised learning problems and are unable to handle general stochastic optimization or missing data problems. Our framework is able to prove ``uniform localized convergence of gradients,"  which improves existing vector-based uniform convergence frameworks \cite{mei2018landscape, foster2018uniform, balakrishnan2017statistical} and provides problem-dependent generalization error bounds for iterative algorithms (See Sections \ref{sec parametric}-\ref{sec em}).

\section{Problem-dependent rates in the ``slow rate" regime}\label{sec slow rate regime}

\subsection{Preliminaries}\label{sec preliminary}

Let  $\pazocal{V}^*$ and $\pazocal{L}^*$ be  the variance and the ``effective loss" at the best hypothesis $h^*$:
\begin{align*}
     \pazocal{V}^*:=\var[\ell(h^*;z)], \quad \pazocal{L}^*:=\P [\ell(h^*;z)-\inf_{\Hy}\ell(h;z)].
\end{align*}
 In this section we study  finite-sample generalization errors that scale tightly with $\V$ or $\LL$, which we call  {problem-dependent rates}, without invoking strong convexity or margin conditions. 

In the slow rate regime, we assume the loss function to be uniformly bounded by $[-B,B]$, i.e., $
    |\ell(h;z)|\leq B
$ for all $h\in\Hy$ and $z\in\Z$.
This is a standard assumption used in almost all previous works that do not invoke curvature conditions or rely on other problem-specific structure; and our results in the slow rate regime essentially only require this boundedness assumption.  Extensions to unbounded targets can be obtained via  truncation techniques (see, e.g. \cite{gyorfi2006distribution}), and our problem-dependent results allow $B$ to be very large, potentially scaling with $n$. 

We represent the complexity through a surrogate function $\psi(r;\delta)$ that satisfies for all $\delta\in(0,1)$,
\begin{align}\label{eq: surrogate preliminary}
    \sup_{f\in\F: \P[f^2]\leq r}(\P-\Pn)f \leq \psi(r;\delta),
\end{align}
with probability at least $1-\delta$, where $\F$ is taken to be the {\it excess loss class}
\begin{align}\label{eq: excess loss class}
 \ell\circ\Hy-\ell\circ h^*:=\{{z\mapsto\left(\ell(h;z)-\ell(h^*;z)\right)}:h\in\Hy\}.
\end{align}
{(From the perspective of Section \ref{subsec refined analysis}, we choose the excess losses as the ``concentrated functions," and use $T(f)=\P[f^2]$ as the ``measurement functional".)}
To achieve non-trivial complexity control (and ensure existence of the fixed point), we only consider ``meaningful" surrogate functions, as stated below. 
\begin{definition}[{\bf meaningful surrogate function}]\label{def fixed point}
 A bivariate function $\psi(r;\delta)$ defined over $[0,\infty)\times(0,1)$ is called a meaningful surrogate function if it is non-decreasing, non-negative and bounded with respect to $r$ for every fixed $\delta\in(0,1)$.
\end{definition}
We note that the above does not place significant   restrictions on the choice of the surrogate function. In particular, for the $\psi$ function defined in \eqref{eq: surrogate preliminary} and the excess loss class in \eqref{eq: excess loss class}, the left hand side of \eqref{eq: surrogate preliminary} is itself non-decreasing and non-negative; and the boundedness requirement can always be met by setting $\psi(r;\delta)=\psi(4B^2;\delta)$ for all $r\geq 4B^2$. We now give the formal definition of fixed points.

 \begin{definition}[{\bf fixed point}]
 Given a non-decreasing, non-negative and bounded function $\varphi(r)$ defined over $[0,\infty)$, we define the fixed point of $\varphi(r)$ to be $\sup\{r>0:\varphi(r)\leq r\}$. 
 \end{definition} 
It is well-known that a non-decreasing, non-negative and bounded function  only has finite discontinuity points, all of which belong to ``discontinuity points of the first kind" [\citealp{rudin1964principles}, Definition 4.26]. Therefore, it is easy to verify that the fixed point of $\varphi(r)$ is the maximal solution to the equation $\varphi(r)=r$.

Given a bounded class $\F$,  empirical process theory provides a general way to construct surrogate function by upper bounding the ``local Rademacher complexity"  $\mathfrak{R}\{f\in\F: \P[f^2]\leq r\}$ (see Lemma \ref{lemma bartlett} in Appendix \ref{appendix auxiliary lemma}). 
We give the definition of  Rademacher complexity for completeness.
   \begin{definition}[{\bf Rademacher complexity}]\label{def Rademacher}
For a function class $\F$ that consists of mappings from $\pazocal{Z}$ to $\R$, define
\begin{align*}
  \mathfrak{R}\F:=\E_{z,\upsilon}\sup_{f\in \F}\frac{1}{n}\sum_{i=1}^n \upsilon_if(z_i), \quad {\mathfrak{R}}_n \F:=\E_{\upsilon}\sup_{f\in \F}\frac{1}{n}\sum_{i=1}^n \upsilon_if(z_i), 
\end{align*}
as the {\it  Rademacher complexiy} and the {\it empirical  Rademacher complexity} of $\F$, respectively, where $\{\upsilon_i\}_{i=1}^n$ are i.i.d. Rademacher variables for which $\text{Prob}(\upsilon_i=1)=\text{Prob}(\upsilon_i=-1)=\frac{1}{2}$. In the above we explicitly denote expectation operators with subscripts that describes the distribution relative to which the expectation is computed. $\E_z$ means taking expectations over $\{z_i\}_{i=1}^n$ and $\E_\upsilon$ means taking expectations over  $\{\upsilon_i\}_{i=1}^n$.
\end{definition}
Furthermore, Dudley's integral bound  provides one general solution to construct a {\it computable} upper bound on the local Rademacher complexity via the covering number of $\F$. We give the definition of covering number and state Dudley's integral bound for completeness as well.
\begin{definition}[{\bf covering number}]
Assume $(\pazocal{M},\metr(\cdot,\cdot))$ is a metric space, and   $\pazocal{F}\subseteq{\pazocal{M}}$. The  $\varepsilon-${\it covering number} of the set $\pazocal{F}$ with respect to a metric $\metr(\cdot,\cdot)$ is the size of its smallest $\varepsilon-$net cover:
\begin{align*}
    \pazocal{N}(\varepsilon,\pazocal{F},\metr) = \min\{m : \exists f_1,...,f_m\in \pazocal{F} \textup{ such that } \pazocal{F} \subseteq \cup_{j=1}^m\pazocal{B}(f_j,\varepsilon)\}.
\end{align*} where $\pazocal{B}(f,\eps):=\{\tilde{f}:\metr(\tilde{f}, f)\leq \eps\}$.
\end{definition}
\noindent We call $\log\pazocal{N}(\eps, \F, \metr)$ the {\it metric entropy} of the set $\F$ with respect to a metric $\metr(\cdot,\cdot)$. Standard metrics include the $L_p(\Pn)$ metric defined by $L_p(\Pn)(f_1,f_2):=\sqrt[p]{\Pn(f_1(z)-f_2(z))^p}$ for  $p>0$.
For function classes characterized by  metric entropy conditions, the surrogate function constructed by Dudley's integral bound is often optimal. We use the $L_2(\Pn)$ metric for simplicity; the result trivially extends to $L_p(\Pn)$ metrics for $p\geq 2$, since   $\log\pazocal{N}(\eps, \F, L_2(\Pn))\leq \log\pazocal{N}(\eps, \F, L_p(\Pn))$ for any set $\F$ and discretization error $\eps>0$.

\begin{lemma}[{\bf Dudley's integral bound, \cite{SridharanNotes}}]\label{lemma dudley Rademacher}
Given $r>0$ and a class $\F$ that consists of functions defined on $\Z$,
\begin{align*}
    \mathfrak{R}_n\{f\in \F: \Pn[f^2]\leq r \}\leq\inf_{\eps_0>0}\left\{ 4\eps_0+12\int_{\eps_0}^{\sqrt{r}}\sqrt{\frac{\log \pazocal{N}(\eps, \F, L_2(\Pn))}{n}}d\eps \right\}.
\end{align*}
\end{lemma}

In what follows, when comparing different complexity parameters, we often use ``$\lor$" to denote the maximum operator, and to interpret correctly, its use should be understood to take precedence over addition but not over multiplication.  Throughout we will find it convenient to use ``big-oh" notation to simplify various expressions and comparisons that capture order of magnitude effects. For two non-negative sequence $\{a_n\}$ and $\{b_n\}$, we write  $a_n=O(b_n)$, if $a_n$ can be upper bounded by $b_n$ up to an absolute constant for sufficiently large $n$. The same expression is often used also in the context of probabilistic statements in which case it is interpreted as holding on an event which has a specified (typically ``high") probability. We write $a_n=\Omega(b_n)$ if $b_n/a_n=O(1)$.

 \subsection{Loss-dependent rates via empirical risk minimization}\label{sec general loss without curvature}

   In this section we are interested in  loss-dependent rates, which should scale tightly with 
   $\LL:=\P[\ell(h^*;z)-\inf_{\Hy}\ell(h;z)]$;
    the best achievable ``effective loss" on $\Hy$. The following theorem characterizes the loss-dependent rate of empirical risk minimization (ERM) via a surrogate function $\psi$, its fixed point $r^*$, the effective loss $\LL$ and $B$.  

\begin{theorem}[\bf loss-dependent rate of ERM]\label{thm slow}
 For the excess loss class  $\F$ in \eqref{eq: excess loss class},  assume there is a meaningful surrogate function $\psi(r;\delta)$ that satisfies $\forall \delta\in(0,1)$ and $\forall r>0$, with probability at least $1-\delta$, 
   \begin{align*}
   \sup_{ f\in\F:\P[f^2]\leq r}(\P-\Pn)f\leq \psi(r;\delta).
\end{align*}
Then the empirical risk minimizer $\hh_{\ERM}\in\argmin_{\Hy}\{\Pn\ell(h;z)\}$ satisfies  for any fixed $\delta\in(0,1)$ and $r_0\in(0,4B^2)$, with probability at least $1-\delta$,
\begin{align*}
    \mathcal{E}(\hh_{\ERM})\leq \psi\left(24B\LL;\frac{\delta}{C_{r_0}}\right)\lor \frac{r^*}{6B}\lor \frac{r_0}{48B},
\end{align*}
where $C_{r_0}=2\log_2\frac{8B^2}{r_0}$, and $r^*$ is the fixed point of $6B\psi\left(8r;\frac{\delta}{C_{r_0}}\right)$.
\end{theorem}

\paragraph{Remarks.} 1) The term $r_0$ is negligible since it can be arbitrarily small. One can simply set $r_0={B^2}/{n^4}$, which will much smaller than $r^*$ in general ($r^*$ is at least  of order $B^2\log\frac{1}{\delta}/n$ in the traditional ``local Rademacher complexity" analysis, because this term is unavoidable in two-sided concentration inequalities). 

\noindent 2) In high-probability bounds, $C_{r_0}$ will only appear in the form $\log ({C_{r_0}}/{\delta}))$, which is of a negligible $O(\log\log n)$ order, so  $C_{r_0}$ can be viewed an absolute constant for all practical purposes. As a result, our generalization error bound can be viewed to be of the order \begin{align}\label{eq: rate of ERM}
\mathcal{E}(\hh_{\ERM})\leq O\left(\psi(B\LL;\delta)\lor \frac{r^*}{B}\right).
\end{align}

\noindent 3) By using the empirical ``effective loss," $\Pn[\ell(\hh_{\ERM};z)-\inf_{\Hy}\ell(h;z)]$, to estimate $\LL$, the loss-dependent rate  can be estimated from data without knowledge of $\LL$. We defer the details to Appendix \ref{subsubsec evaluate loss rate}.

\paragraph{Comparison with existing results. }Under additional restrictions (to be explained later), the traditional analysis  \eqref{eq: local rademacher complexity} leads to a loss-dependent rate of the order \cite{bartlett2005local} \begin{align}\label{eq: previous ERM}
\mathcal{E}(\hh_{\ERM})\leq O\left(\sqrt{\frac{\LL  r^*}{B}}\lor\frac{r^*}{B}\right),\end{align}
which  is strictly worse than our result \eqref{eq: rate of ERM} due to  reasoning following Statement \ref{state improvement}.  When $B\LL\leq O( r^*)$, both \eqref{eq: rate of ERM} and \eqref{eq: previous ERM} are dominated by the order ${r^*}/{B}$ so there is no difference between them. However, when $ B\LL\geq \Omega (r^*)$, our result \eqref{eq: rate of ERM} will be of order $\psi(B\LL;\delta)$ and the previous result \eqref{eq: previous ERM} will be of order $\sqrt{{\LL r^*}/{B}}$. In this case,
the square-root function $\sqrt{{\LL r^*}/{B}}$ is only a coarse relaxation of $\psi(B\LL;\delta)$: as the traditional analysis requires $\psi$ to be sub-root, we can compare  the two orders by
\begin{align}\label{eq: loss sub root}
    \psi\left(B\LL;{\delta}\right)\overset{\text{sub-root}}{\leq} \sqrt{\frac{B\LL}{r^*}}\psi(r^*;\delta)\overset{\text{fixed point}}{=} O\left(\sqrt{\frac{\LL r^*}{B}}\right).
\end{align}
 The ``sub-root" inequality (the first inequality in \eqref{eq: loss sub root})  becomes an equality when  $\psi(r;\delta)=O({\sqrt{d r/n}})$  in the parametric case, where $d$ is the parametric dimension. However, when $\F$ is ``rich," $\psi(r;\delta)/\sqrt{r}$ will be strictly decreasing so that the ``sub-root" inequality can become quite loose. For example, when $\F$ is a non-parametric class we often have $\psi(r;\delta)=O(\sqrt{r^{1-\rho}/n})$ for some $\rho\in(0,1)$. The richer $\F$ is (e.g., the larger $\rho$ is),  the more loose the ``sub-root" inequality. This intuition will be validated via examples in Section \ref{sec application}.

Theorem \ref{thm slow} also applies to broader settings than previous results. For example, in \cite{bartlett2005local} it is assumed that the loss is non-negative, and their original result only adapts to $\P\ell(h^*;z)$ rather than the ``effective loss" $\LL$.  Our  proof (see Appendix \ref{subsec appendix thm slow}) is quite different as we bypass the Bernstein condition  (which  is  traditionally  implied  by non-negativity, but not satisfied by the class used here), bypass the sub-root assumption on $\psi$, and adapt to the ``better"  parameter $\LL$.

\subsection{Variance-dependent rates via moment penalization}\label{sec empirical penalization}

The loss-dependent rate  proved in Theorem \ref{thm slow} {contains} a complexity parameter $B\LL$ within its $\psi$ function, which may still be much larger than the optimal variance $\V$. Despite its prevalent use in practice, standard empirical  risk minimization  is unable to achieve variance-dependent rates in general.  An example is given in \cite{namkoong2017variance} where $\V=0$ and the optimal rate is at most $O({\log n}/{n})$, while $\mathcal{E}(\hh_{\ERM})$ is proved to be slower than $n^{-\frac{1}{2}}$.

We follow the path of penalizing empirical second moments (or variance) \cite{maurer2009empirical, swaminathan2015counterfactual, namkoong2017variance, foster2019orthogonal}  to design an estimator that achieves the ``right" bias-variance trade-off for general, potentially ``rich," classes. Our proposed estimator simultaneously achieves correct scaling on $\V$,  along with minimax-optimal sample dependence ($n$). Besides empirical first and second moments, it
only depends on the boundedness parameter $B$, a computable surrogate function $\psi$, and the confidence parameter $\delta$. All of these quantities  are essentially assumed known in previous works: e.g.,  \cite{maurer2009empirical, swaminathan2015counterfactual} require covering number of the loss class, which implies a computable surrogate $\psi$ via Dudley's integral bound; and estimators in \cite{namkoong2017variance, foster2019orthogonal} rely on the fixed point $r^*$ of a computable surrogate $\psi$.

In order to adapt to $\V$, we use a sample-splitting two-stage estimation procedure ({this idea is inspired by the prior work \cite{foster2019orthogonal}}). Without loss of generality, we assume access to a data set of size $2n$. We split the data set into the ``primary" data set $S$ and the ``auxiliary" data set $S'$, both of which are of size $n$. We denote $\Pn$ the empirical distribution of the ``primary" data set, and $\P_{S'}$ the empirical distribution of the ``auxiliary" data set.

\begin{strategy}[{\bf the two-stage sample-splitting estimation procedure.}]
At the first-stage, we derive a preliminary estimate of $\LL_0:=\P\ell(h^*;z)$ via the ``auxiliary" data set $S'$, which we refer to as $\LLL$.  Then, at the second stage, we perform regularized empirical risk minimization on the ``primal" data set $S$, which penalizes the centered second moment $\Pn[(\ell(h;z)-\LLL)^2]$. 
\end{strategy}

As we will present later, it is rather trivial to obtain a qualified preliminary estimate $\LLL$ via empirical risk minimization. Therefore, we firstly introduce the second-stage moment-penalized estimator, which is more crucial and interesting. 
\begin{strategy}[{\bf the second-stage moment-penalized estimator.}]\label{strategy estimator}
 Consider the excess loss class $\F$ in \eqref{eq: excess loss class}. Let $\psi(r;\delta)$ be a meaningful surrogate function  that satisfies  $\forall \delta\in(0,1)$, $\forall r>0$, with probability at least $1-\delta$,
\begin{align*}
4\mathfrak{R}_n\{f\in\F:\Pn[f^2]\leq 2r\}+\sqrt{\frac{2r\log\frac{8}{\delta}}{n}}+\frac{9B\log\frac{8}{\delta}}{n}\leq \psi(r;\delta).
\end{align*}
Denote $C_n=2\log_2n +5$. Given a fixed $\delta\in(0,1)$, let the estimator $\hh_{\MP}$ be
\begin{align*}
\hh_{\MP}\in\arg\min_{\Hy}\left\{\Pn \ell(h;z)+\psi\left(16\P_n[(\ell(h;z)-\LLL)^2];\frac{\delta}{C_n}\right)\right\}.
\end{align*}  
\end{strategy}

Given an arbitrary preliminary estimate $\LLL\in [-B,B]$, we can prove that the generalization error of the moment-penalized estimator $\hh_{\MP}$ is at most
\begin{align}\label{eq: difficult variance}
    \mathcal{E}(\hh_{\MP})\leq 2\psi\left(c_0 \left[\V\lor (\LLL-\LL_0)^2\lor r^*\right];\frac{\delta}{C_n}\right),
\end{align}
with probability at least $1-\delta$, where $c_0$ is an absolute constant, and $r^*$ is  the fixed point of  $16B\psi(r;\frac{\delta}{C_n})$. Moreover,    the first-stage estimation error will be negligible if 
\begin{align}\label{eq: first stage error}
    (\LLL-\LL_0)^2\leq O\left( r^*\right).
\end{align}
It is rather elementary to show that performing the standard empirical risk minimization on $S'$ suffices to satisfy \eqref{eq: first stage error}, provided an additional assumption that $\psi$ is a ``sub-root" function. We now give our theorem on the generalization error following this two-stage procedure.
\begin{theorem}[{\bf variance-dependent rate}]\label{thm variance rate}
Let $\LLL=\inf_{\Hy}\mathbb{P}_{S'} \ell(h;z)$ be attained via empirical risk minimization on the auxiliary data set $S'$.  Assume that the meaningful surrogate function $\psi(r;\delta)$ is ``sub-root," i.e. $\frac{\psi(r;\delta)}{\sqrt{r}}$ is non-increasing over $r\in[0,4B^2]$ {for all fixed $\delta$. Then for any  $\delta\in(0,\frac{1}{2})$,} by performing the moment-penalized estimator in  Strategy \ref{strategy estimator}, with probability at least $1-2\delta$,
\begin{align*}
    \mathcal{E}(\hh_{\MP})\leq 2\psi \left(c_1\V;\frac{\delta}{C_n}\right)\lor \frac{c_1 r^*}{8B},
\end{align*}
where $r^*$ is the fixed point of $B\psi(r;\frac{\delta}{C_n})$ and $c_1$ is an absolute constant.
\end{theorem}
\paragraph{Remarks.} 1) In high-probability bounds, $C_{n}$ will only appear in the form $\log ({C_{n}}/{\delta}))$, which is of a negligible $O(\log\log n)$ order, so  $C_n$ can effectively be viewed as constant for all practical purposes.

\noindent 2) The ``sub-root" assumption in Theorem \ref{thm variance rate} is only used to to bound the first-stage estimation error (see \eqref{eq: first stage error}). This assumption is not needed for the result \eqref{eq: difficult variance} on the second-stage moment-penalized estimator.

\noindent 3) Replacing $\V$ by an empirical centered second moment, we can prove a fully data-dependent generalization error bound that is computable from data without the knowledge of $\V$. 
We leave the full discussion to Appendix \ref{subsec estimate variance rate}.
\paragraph{Comparison with existing results.} The best variance-dependent rate attained  by existing estimators is of the order \cite{foster2019orthogonal} $$\sqrt{\frac{\V  r^*}{B^2}}\lor\frac{r^*}{B},$$  which is strictly worse than the rate proved in Theorem \ref{thm variance rate}. The  reasoning is similar to Statement \ref{state improvement} and the explanation after Theorem \ref{thm slow}: when $\V\leq O( r^*)$ the two results are essentially identical, but our estimator can perform much better when $\V\geq \Omega (r^*)$. Because $\psi$ is sub-root and $r^*$ is the fixed point, we can compare the orders of the rates 
\begin{align*}
    \psi(\V;\delta)\overset{\text{sub-root}}{\leq} \sqrt{\frac{\V}{r^*}}\psi(r^*;\delta)\overset{\text{fixed point}}{=}O\left( \sqrt{\frac{\V r^*}{B^2}}\right).
\end{align*}
 Since variance-dependent rates are generally used in applications that require robustness or exhibit large worst-case boundedness parameter,  $\V\geq r^*$ is the more critical regime where one wants to ensure the estimation performance will not degrade.

\paragraph{Discussion.} Per our ``uniform localized convergence" principle, the most obvious difficulty in proving Theorem \ref{thm variance rate} is in establishing \eqref{eq: difficult variance}: the empirical second moment is sample-dependent, whereas standard tools in empirical process theory (e.g., Talagrand's concentration inequality, see Lemma \ref{lemma bartlett}) require the localized subsets to be independent of the samples. The core techniques in our proof essentially overcome this difficulty by concentrating data-dependent localized subsets to data-independent ones. This idea may be of independent interest; We defer details to  Appendix \ref{subsec appendix them slow localized complexity penalization}.

 The tightness of our variance-dependent rates depend on tightness of the computable surrogate function $\psi$. When covering numbers of the excess loss class are given, a direct choice is Dudley's integral bound (see Lemma \ref{lemma dudley Rademacher}), which is known to be rate-optimal for many important classes.

Previous approaches usually take a simper regularization term \cite{maurer2009empirical, foster2019orthogonal} that is
proportional to the square root of the empirical second moment (or empirical variance). 
That type of penalization is ``too aggressive" for rich classes from our viewpoint. 
\cite{namkoong2017variance} propose a regularization term that preserves convexity of empirical risk.  However, based on an equivalence proved in their paper, they have similar limitations to the approaches that penalizes the square root of the empirical variance. 
Outside the study of variance-dependent rates,  integral-based and local-Rademacher-complexity-based penalization is also used in model selection \cite{lugosi2002pattern}, but the setting and the goal of model selection are very different from the problem we study here.

\section{Illustrative examples in the ``slow rate" regime}\label{sec application}

\subsection{Discussion and illustrative examples}\label{subsec slow rate regime examples}
Recall that our  loss-dependent rates and variance-dependent (moment-penalized) rates are
 \begin{align}
 \mathcal{E}(\hh_{\ERM})\leq O\left(\psi(B\pazocal{L}^*;\delta)\lor\frac{r^*}{B}\right) \quad [\text{Theorem \ref{thm slow}}],\label{eq: our loss dependent}\\ \mathcal{E}(\hh_{\MP})\leq O\left(\psi(\V;\delta)\lor \frac{r^*}{B}\right) \quad  [\text{Theorem \ref{thm variance rate}}],\label{eq: our variance dependent}
 \end{align}respectively.
In contrast to our results \eqref{eq: our loss dependent} \eqref{eq: our variance dependent}, the best known  loss-dependent rates \cite{bartlett2005local} and variance-dependent rates \cite{foster2019orthogonal} are
\begin{align}
\mathcal{E}(\hh_{\ERM})\leq O\left(\sqrt{\frac{\pazocal{L}^*r^*}{B}}\lor\frac{r^*}{B}\right)\quad [\text{existing result \cite{bartlett2005local}}],\label{eq: previous loss dependent}\\
\mathcal{E}(\hh_{\text{previous}})\leq
O\left(\sqrt{\frac{\V r^*}{B^2}}\lor\frac{r^*}{B}\right)\quad [\text{existing result \cite{foster2019orthogonal}}],\label{eq: previous variance dependent}
\end{align}
respectively (we use $\hh_{\text{previous}}$ to denote the previous best known moment-penalized estimator proposed in \cite{foster2019orthogonal}). To illustrate the noticeable gaps between our new results and previous known ones, we compare the two different variance-dependent rates, \eqref{eq: our variance dependent} and \eqref{eq: previous variance dependent} on two important families of ``rich" classes: non-parametric classes of polynomial growth and VC classes. The  implications of this comparison will similarly apply to loss-dependent rates. 

Before presenting the advantages of the new problem-dependent rates, we would like to discuss how to compute them. In Theorem \ref{thm slow} and Theorem \ref{thm variance rate}, the class of concentrated functions, $\F$, is the excess loss class $\ell\circ \Hy-\ell\circ h^*$ in \eqref{eq: excess loss class}. As we have mentioned in earlier sections, a general solution for the $\psi$ function is to use Dudley's integral bound (Lemma \ref{lemma dudley Rademacher}). Knowledge of  the metric entropy of the excess loss class, 
 \begin{align*}
     \log\pazocal{N}(\eps,\ell\circ \Hy-\ell\circ h^*, L_2(\Pn )),
 \end{align*}
 can be used to calculate Dudley's integral bound and construct the surrogate function $\psi$ needed in our theorems. Note that there is no difference between the metric entropy of the excess loss class and metric entropy of the loss class itself: from the definition of covering number, one has
  \begin{align*}
     \pazocal{N}(\eps,\ell\circ \Hy-\ell\circ h^*, L_2(\Pn ))=\pazocal{N}(\eps,\ell\circ \Hy, L_2(\Pn )).
 \end{align*}
 We comment that almost all existing  {\it theoretical} literature that {discusses} general function classes and losses \cite{bartlett2005local, maurer2009empirical, swaminathan2015counterfactual, foster2019orthogonal} imposes metric entropy conditions on the loss class/excess loss class rather than the hypothesis class $\Hy$, and we follows that line as well to allow for a seamless comparison of the results. 
As a complement,  we will discuss how to obtain such metric entropy conditions for practical applications in Section \ref{subsec application areas slow rate}.

\subsubsection{Non-parametric classes of polynomial growth}
\begin{example}[{\bf non-parametric classes of polynomial growth}]\label{example non para}Consider a loss class $\ell\circ\Hy$ with the metric entropy  condition
\begin{align}\label{eq: metric entropy growth}
    \log\pazocal{N}(\eps, \ell\circ\Hy, L_2(\Pn))\leq O\left( \eps^{-2\rho}\right),
\end{align} where $\rho\in(0,1)$ is a constant.  Using Dudley's integral bound  to find  $\psi$ and solving $r\leq O\left( B\psi(r;\delta)\right)$, it is not hard to verify that
\begin{align*}
\psi(r;\delta)\leq O\left( \sqrt{\frac{r^{1-\rho}}{n}}\right), \quad r^*\leq O\left (\frac{B^\frac{2}{1+\rho}}{n^{\frac{1}{1+\rho}}}\right).
\end{align*}
\end{example}

As a result, our variance-dependent rate  \eqref{eq: our variance dependent} is of the order
 \begin{align}\label{eq: our poly}
 \mathcal{E}(\hat{h}_{\MP})\leq O\left( {\V}^{\frac{1-\rho}{2}}  {n}^{-\frac{1}{2}}\lor \frac{r^*}{B}\right),
 \end{align} which is  $O\left({\V}^{\frac{1-\rho}{2}}  {n}^{-\frac{1}{2}}\right)$  when $\V \geq \Omega( r^*)$. In contrast, the previous best-known result \eqref{eq: previous variance dependent} is of the order
 \begin{align}\label{eq: their poly}
  \mathcal{E}(\hat{h}_{\textup{previous}})\leq   O\left( \sqrt{\V}B^{-\frac{\rho}{1+\rho}}n^{-\frac{1}{2+2\rho}}\lor \frac{r^*}{B}\right),
 \end{align}
 which is $O\left(\sqrt{\V}B^{-\frac{\rho}{1+\rho}}n^{-\frac{1}{2+2\rho}}\right)$ when $\V \geq \Omega\left( r^*\right)$. 
 Therefore, for arbitrary choice of $n, \V, B$, the ``sub-optimality gap"   is 
 \begin{align}\label{eq: gap}
  \textup{ratio between \eqref{eq: their poly} and  \eqref{eq: our poly}}:=\frac{\sqrt{\V}B^{-\frac{\rho}{1+\rho}}n^{-\frac{1}{2+2\rho}}\lor \frac{r^*}{B}}{{\V}^{\frac{1-\rho}{2}}  {n}^{-\frac{1}{2}}\lor \frac{r^*}{B}}= 1\lor  (\V (\frac{n}{B^2})^{\frac{1}{1+\rho}})^{\frac{\rho}{2}},
 \end{align}
 which can be arbitrary large and grows polynomially with $n$. 
 
  We consider two stylized regimes as follows (we use the notation $\approx$ when the left hand side and the right hand side are of the same order).
  \begin{itemize}
  \item The more ``traditional" regime: $B\approx 1$, $\V\approx n^{-a}$ where $a>0$ is a fixed constant. This regime captures the  traditional supervised learning problems where  $B$ is not large, but one wants to use the relatively small order of $\V$ to achieve ``faster" rates.
  
  \item The ``high-risk" regime: $B\approx n^b$ where $b>0$ is a fixed constant, and  $\V\ll B^2$ (i.e., $\V$ is much smaller than order $n^{2b}$). This regime captures modern ``high-risk" learning problems such as counterfactual risk minimization, policy learning, and supervised learning with limited number of samples. In those settings, the worst-case boundedness parameter is considered to scale with $n$ so that one wants to avoid (or reduce) the dependence on $B$.
  \end{itemize}
In both of the two regimes,  generalization errors via naive (non-localized) uniform convergence arguments will be worse than our approach by orders polynomial in $n$, so we only need to compare with previous variance-dependent rates.
 
\paragraph{\bf The ``traditional" regime.} The ``sub-optimality gap" \eqref{eq: gap} is  $1\lor (\V n^{\frac{1}{1+\rho}})^{\frac{\rho}{2}}$. It is quite clear that when $\V\approx n^{-a}$ where $0<a<\frac{1}{1+\rho}$, our variance-dependent rate improves over all previous generalization error rates by orders polynomial in $n$.
 
\paragraph{\bf The ``high-risk" regime.} We focus on the simple case $B^{\frac{2}{1+\rho}}\leq  \V\ll 4B^2$ to gain some insight, where our result exhibits an improvement of order $O(n^{\frac{\rho}{2}(\frac{1}{1+\rho})})$ relative to the previous result. Clearly the larger $\rho$, the more improvement we provide. By letting $\rho\rightarrow 1$ our improvement can be as large as $O(n^{\frac{1}{4}})$. 
 

\subsubsection{VC-type classes}
 Our next example considers VC-type classes. Although this classical example has been extensively studied in learning theory, our results provide strict improvements over antecedents. 
 \begin{example}[{\bf VC-type classes}]\label{example vc}One general definition of VC-type classes (which is not necessarily binary) uses the metric entropy condition. Consider a loss class $\ell\circ \Hy$ that satisfies
\begin{align*}
    \log\pazocal{N}(\eps, \ell\circ\Hy, L_2(\Pn))\leq O\left( d\log\frac{1}{\eps}\right),
\end{align*}where $d$ is th so-called the Vapnik–Chervonenkis (VC) dimension \cite{van2000asymptotic}. 
 Using Dudley's integral bound  to find the surrogate $\psi$ and solving $r\leq O( B\psi(r;\delta))$, it can be proven \cite{koltchinskii2000rademacher} that 
\begin{align*}
    \psi(r;\delta)\leq O\left( \sqrt{\frac{d r}{n}\log\frac{{8}B^2}{r}}\lor \frac{Bd}{n}\log\frac{8B^2}{r}\right), \quad  r^*\leq O\left( \frac{B^2d\log n}{n}\right).
\end{align*} 
\end{example}

Recently,  \cite{foster2019orthogonal} proposed a moment-penalized estimator whose generalization error is of the rate
\begin{align*}
  \mathcal{E}(\hh_{\textup{previous}})\leq O\left( \sqrt{\frac{d\V \log n}{n}}+ \frac{Bd\log n}{n}\right),
\end{align*}
in the worst case without invoking other assumptions. This result has a $O(\log n)$ gap compared with the  $\Omega(\sqrt{\frac{d\V }{n}})$ lower bound \cite{devroye1995lower}, which holds for arbitrary sample size. There is much recent interest focused on the question of when the sub-optimal $\log n$ factor can be removed \cite{athey2017efficient, foster2019orthogonal}.

By applying Theorem \ref{thm variance rate}, our refined moment-penalized estimator gives a generalization error bound of  tighter rate
\begin{align}\label{eq: our vc}
  \mathcal{E}(\hh_{\MP})\leq  O\left(\sqrt{\frac{d\V\log\frac{{8}B^2}{\V}}{n}}\lor\frac{Bd\log n}{n}\right).
\end{align} This closes the $O(\log n)$ gap in the regime $\V\geq \Omega(\frac{B^2}{(\log n)^{\alpha}})$, where $\alpha>0$ is an arbitrary positive constant. Though this is not the central regime, it is  the first positive result that closes the notorious $O(\log n)$ gap without invoking any additional assumptions on the loss/hypothesis class  (e.g., the
rather complex “capacity function" assumption introduced in \cite{foster2019orthogonal}). We anticipate additional improvements are possible under further assumptions on the hypothesis class and the loss function.

\subsection{Problem areas to which ``localization"  theory is applicable}\label{subsec application areas slow rate}

In practical applications it is more standard to consider metric entropy conditions of the hypothesis class $\Hy$.
 In view of this, we introduce two important settings where the metric entropy on the loss/excess loss class can be obtained from  metric entropy conditions on the hypothesis class $\Hy$. {Thus, the improvements illustrated in Section \ref{subsec slow rate regime examples} can be directly transferred to these application areas.}
 
\paragraph{\bf Supervised learning {with  Lipchitz continuous cost}.} In supervised learning, the data $z$ is a feature-label pair $(x,y)$,  and the loss  $\ell(h;z)$ is of the form  $$\ell(h;z)=\ell_{\sv}(h(x),y),$$ where   $\ell_{\sv}$  is a fixed cost function that is $L_{\sv}-$Lipchitz continuous with respect to its first argument, namely, Lipchitz with parameter $L_{\sv}$. 
For hypothesis classes characterized by metric entropy conditions,  properties are preserved because
\begin{align*}
   \log \pazocal{N}(\eps, \ell\circ\Hy, L_2(\Pn))\leq \log \pazocal{N}(\frac{\eps}{L_{\sv}}, \Hy, L_2(\Pn)).
\end{align*}
Note that $L_{\sv}$ only depends on the cost function and is usually of  constant order. Our theory naturally applies to supervised learning problems where the  cost function is Lipchitz continuous and not strongly-convex (for example, the $\ell_1$ cost, the hinge cost, the ramp cost, etc.).
\paragraph{\bf Counterfactual risk minimization.} Denote $x\in\pazocal{X}$ the feature and $t\in\pazocal{T}$ the treatment (e.g. $\pazocal{T}=\{0,1\}$ in binary treatment experimental design), and $c(x,t)$ the unknown cost function. A hypothesis (policy) $h$ is a map from  $\pazocal{X}\times \pazocal{T}$ to $[0,1]$ such that $\sum_{t\in\pazocal{T}}h(x,t)=1$. Thus, a hypothesis (policy) essentially maps features to a distribution over treatments. We consider the standard formulation of  ``learning with logged bandit feedback," dubbed ``counterfactual risk minimization" \cite{swaminathan2015counterfactual}: a batch of samples $\{(x_i,t_i,c_i)\}_{i=1}^n$ are obtained by applying a known policy $h_0$, so that $t_i$ is sampled from  $h_0(x_i,\cdot)$ and one can only observe the cost $c_i$ associated with $t_i$. We write $z=(x,t,c)$ and let 
\begin{align}\label{eq: counterfactual risk minimization}
\ell(h;z_i)=\frac{c_i}{h_0(x_i,t_i)}h(x_i,t_i),
\end{align} be the ``constructed loss" using importance sampling. It is straightforward to show that the population risk $\P\ell(h;z)$ is equal to the expected cost of policy $h$, so determining good policies requires one to minimize the generalization error $\mathcal{E}(\hh)$. It is usually convenient to obtain metric entropy condition of the  loss/excess loss class by using the linearity structure of \eqref{eq: counterfactual risk minimization}. In particular, from the Cauchy-Schwartz inequality we can prove that
\begin{align}\label{eq: crm metric entropy}
   \log \pazocal{N}(\eps, \ell\circ\Hy, L_2(\Pn))\leq  \log \pazocal{N}(\frac{\eps}{\gamma_n}, \Hy, L_{4}(\Pn)),
\end{align}
where $\gamma_n:=\sqrt[4]{\Pn\left[ (\frac{c(x,t)}{h_0(x,t)})^4\right]}$ only depends on the functions $c$, $h_0$ in the given problem, and the samples rather than the worst-case parameters. A systematical challenge in counterfactual risk minimization is that the worst-case boundedness parameter, $\sup_{h,z}|\ell(h;z)|$, is typically very large, since the inverse probability term $\frac{1}{h_0(x_i,t_i)}$ in \eqref{eq: counterfactual risk minimization} is typically large in the worst case.

\section{Problem-dependent rates in the parametric ``fast rate" regime}\label{sec parametric}
 
\subsection{Background}
When assuming suitable curvature or margin conditions,  the direct dependence of the generalization error on $n$ is typically faster than $O(n^{-\frac{1}{2}})$. We call this  the ``fast rate" regime.
A well-known example is, when the hypothesis class is parametrized and the loss is strongly convex with respect to the parameter, in which case the direct dependence of the generalization error on $n$ is typically the ``parametric rate" $O(n^{-1})$. In the ``fast rate" regime, existing problem-dependent rates are mostly studied in  supervised learning with structured convex cost; see  Section \ref{subsec history} for a historical review of existing localization approaches and problem-dependent rates.
Through our proposed ``uniform localized convergence" procedure, we can recover results in \cite{mendelson2014learning, mendelson2018learning} for supervised learning problems with structured convex cost, and our approach is able to systematically weaken some restrictions. (We defer the discussion to Section \ref{sec supervised learning strongly convex} {as the contributions there mostly lie in unification and some technical improvements.)}

In this and the next two sections we study more modern applications in the fast rate regime, focusing on  computationally efficient estimators and algorithms, where the derivation of sharp problem-dependent rates remains an open question. A secondary objective is to  {provide unification to vector-based uniform convergence analysis}.
The main focal points are  the following.
 
{
\paragraph{\bf Non-convexity, stationary points, and iterative algorithms.} Classical generalization error analysis relies on the property of global empirical risk minimizers. However, many important machine learning problems are non-convex. For those problems, guarantees on global empirical risk minimizes are not sufficient, and therefore one typically targets  guarantees on stationary points and the iterates produced by optimization algorithms. 
This motivates us to study uniform convergence of gradients and sample-based iterative algorithms. Existing generalization error bounds in this area are typically not localized, and connections to traditional localization frameworks is not fully understood.

\paragraph{\bf General formulation of stochastic optimization.} Existing problem-dependent rates mostly focus on supervised learning settings, with specialized assumptions on the problem structure. Hence, it is important to characterize   problem-dependent generalization error bounds for more general stochastic optimization problems, in which the classical asymptotic results do not depend on the parametric dimension $d$ and global parameters. Existing methods, however, typically give rise to dimension-dependent factors and ``large" global parameters.

\ 

}

Organization of this section is as follows. In Section \ref{subsec theoretical foundation fast}, we will strengthen the ``uniform convergence of gradients" idea by developing  a theory for  ``uniform localized convergence of gradients." In Section \ref{subsec problem dependent fast} we will provide problem-dependent rates for approximate stationary points of empirical risk and iterates of the gradient descent algorithm.

 \subsection{Theoretical foundations}\label{subsec theoretical foundation fast}
 
 Recently, the idea of ``uniform convergence of gradients" \cite{mei2018landscape, foster2018uniform, davis2018graphical} has been applied successfully to many non-convex learning and stochastic optimization problems. These works do not consider problem-dependent rates, and their results typically rely on various global parameters, like global Lipchitz constants and the radius of the parametric set.  In this subsection we strengthen these ideas by developing  a theory of  
``uniform localized convergence of gradients." This theory will be proven to be  more powerful in deriving problem-dependent rates. Before moving to state a key assumption, we introduce some additional notation.
 
 \paragraph{Notation for the parametric ``fast rate" regime.} We  write the loss function as $\ell(\theta;z)$ where $\theta\in \R^d$ is the parameter representation of the hypothesis $h$. Consider a compact set $\Theta\subseteq \R^d$ and let
 $\theta^*$ be the best parameter within $\Theta$, which satisfies
$\theta^*\in\argmin_{\Theta}\P\ell(\theta;z)$. Denote $\|\cdot\|$ to be the  $L_2$ norm in $\R^d$, noting that most of our results can be generalized to matrix learning problems by considering the Frobenius norm.  We let  $\pazocal{B}^d(\theta_0,\rho):=\{\theta\in\R^d: \|\theta-\theta_0\|\leq \rho\}$ denote a ball with center $\theta_0\in\R^d$ and radius $\rho$. 
    We assume that there are two radii $\Delta_m, \Delta_M$ such that  $\pazocal{B}^d(\theta^*,\Delta_m)\subseteq \Theta\subseteq \pazocal{B}^d(\theta^*,\Delta_M)$.  We would like to provide guarantee for the generalization error
\begin{align*}
    \mathcal{E}(\hth):=\P\ell(\hth;z)-\P\ell(\theta^*;z).
\end{align*} 

We state a key  assumption of our framework.  

\begin{assumption}[{\bf statistical noise of smooth population risk}]\label{asm hessian noise} 
 For all $\theta_1,\theta_2\in\Theta$, $\frac{\nabla  \ell(\theta_1;z)-\nabla \ell(\theta_2;z) }{\|\theta_1-\theta_2\|}$ is a $\beta-$sub-exponential random vector. Formally there exist $\beta>0$ such that 
for any unit vector $u\in\pazocal{B}(0,1)$ and  $\theta_1,\theta_2\in \Theta$, 
\begin{align*}
    \E\left\{\exp\left(\frac{|u^T(\nabla\ell(\theta_1;z)-\nabla \ell(\theta_2;z))|}{\beta\|\theta_1-\theta_2\|}\right)\right\}\leq 2.
\end{align*}
\end{assumption}

 From Jensen's inequality and convexity of the exponential function, a simple consequence of Assumption \ref{asm hessian noise} is that the population risk is $\beta-$smooth: for any $\theta_1,\theta_2\in\Theta$, 
\begin{align}\label{eq: smoothness due to noise assumption}
    \|\P\nabla \ell(\theta_1;z)-\P\nabla \ell(\theta_2;z)\|\leq \beta \|\theta_1-\theta_2\|.
\end{align} 
Smoothness of the population risk  is a standard assumption in the optimization literature. Compared with the smoothness condition \eqref{eq: smoothness due to noise assumption}, Assumption \ref{asm hessian noise} is  a stronger distributional assumption:  it requires the random vectors $\left\{\frac{\nabla \ell(\theta_1;z)-\nabla \ell(\theta_2;z)}{\|\theta_1-\theta_2\|}\right\}$ to be a $\beta-$sub-exponential for all $\theta_1,\theta_2\in\Theta$, while the smoothness condition \eqref{eq: smoothness due to noise assumption}  only concerns the expectation of these random vectors. Certain distributional assumptions are necessary to analyze the generalization performance of unbounded losses (e.g., see the ``Hessian statistical noise" assumption in  \cite{mei2018landscape},  and the ``moment-equivalence" assumptions in \cite{mendelson2017aggregation, mendelson2018robust}).
Assumption \ref{asm hessian noise} imposed here is applicable to many smooth machine learning models studied in previous literature, and its verification is often no  harder than verification of  smoothness conditions.

\begin{lemma}[{\bf Hessian statistical noise implies Assumption \ref{asm hessian noise}}]\label{lemma hessian}
Assumption \ref{asm hessian noise} is satisfied if for all $\theta\in\Theta$, the random Hessian matrix $\nabla^2\ell(\theta;z)$ is a $\beta-$sub-exponential matrix. Formally, Assumption \ref{asm hessian noise} is satisfied when there exist $\beta>0$ such that 
for any unit vectors $u_1,u_2\in\pazocal{B}^d(0,1)$ and any $\theta\in \Theta$, 
\begin{align}\label{eq: hessian 2}
    \E\left\{\exp\left(\frac{1}{\beta}|u_1^T\nabla^2\ell(\theta;z)u_2|\right)\right\}\leq 2.
\end{align}
\end{lemma}
Therefore, one only needs to compute the Hessian matrices and verify they are sub-exponential  over $\Theta$. For instance,  many statistical estimation problems  have $\nabla^2\ell(\theta;z)$ proportional to $zz^T$ (or $xx^T$ when the problem is a supervised learning problem and $z=(x,y)$ is the feature-label pair). By  assuming the data $z$ (or $x$) is a $\tau-$sub-Gaussian vector\footnote{a random vector $z$ is called $\tau-$sub-Gaussian if for all unit vectors $u\in\pazocal{B}^d(0,1)$, $u^Tz$ is a $\tau-$sub-Gaussian variable. See Definition \ref{def orlicz subgaussian subexponential} for details.}, $zz^T$ (or $xx^T$) becomes a $\tau^2-$sub-exponential matrix. If the remaining quantities in $\nabla^2\ell(\theta;z)$ can be uniformly bounded by some constant $C_0$, then Assumption \ref{asm hessian noise} holds with $C_0\tau^2$. 

We carefully choose a function class $\G=\left\{g_{(\theta,v)}: \theta\in\Theta,  v\in\pazocal{B}^d(0,\max\{\Delta_M,\frac{1}{n}\})\right\}$ to apply concentration, where each element is a function $g_{(\theta,v)}:\Z\rightarrow\R$ defined by
\begin{align*}
g_{(\theta, v)}(z)=(\nabla\ell(\theta;z)-\nabla\ell(\theta^*;z))^T v.
\end{align*}
By applying Proposition \ref{prop peeling} with the ``concentrated functions" $g_{(\theta,v)}$ and the ``measurement functional" $T$ defined by $T((\theta,v))=\|\theta-\theta^*\|^2+\|v\|^2$,
  we obtain the key arguemnt of ``uniform localized convergence" for  gradients. 
\begin{proposition}[{\bf uniform localized convergence of gradients}]\label{prop localized gradient 1}
 Under Assumption \ref{asm hessian noise},  $ \forall\delta\in(0,1)$,   with probability at least $1-\delta$, for all $ \theta\in \Theta$,
\begin{align}\label{eq: localized uniform covergence of gradient}
&\left\|(\P-\Pn) (\nabla \ell(\theta;z)-\nabla \ell(\theta^*;z))\right\|\nonumber\\
&\leq 
     c \beta\max\left\{\|\theta-\theta^*\|,\frac{1}{n}\right\}\left( \sqrt{\frac{d+\log\frac{4\log_2 (2n\Delta_M+2)}{\delta}}{n}}+\frac{d+\log\frac{4\log_2 (2n\Delta_M+2)}{\delta}}{n}\right),
\end{align}
where $c$ is an absolute constant.
\end{proposition}

 Distinct from previous results on ``uniform convergence of gradients," which give the same upper bound on $\|(\P-\Pn) \nabla \ell(\theta;z)\|$ for all $\theta\in\Theta$,  the right hand side of \eqref{eq: localized uniform covergence of gradient} in Proposition \ref{prop localized gradient 1} scales linearly with $\|\theta-\theta^*\|$ for all $\theta$ such that $\|\theta-\theta^*\|\geq \frac{1}{n}$. Therefore, Proposition \ref{prop localized gradient 1} provides a refined, ``localized" upper bound on the concentration of gradients. This property makes Proposition \ref{prop localized gradient 1} the key in deriving problem-dependent rates.

\subsection{Main results}\label{subsec problem dependent fast}
In order to obtain tight problem-dependent rates, we require a very mild assumption on the noise at the optimal point $\theta^*$.
\begin{assumption}[{\bf noise at the optimal point}]\label{asm bernstein condition para}
The gradient at $\theta^*$ satisfies the Bernstein condition: there exists $G_*>0$ such that for all $2\leq k\leq n$,
\begin{align}
    \E[\|\nabla \ell(\theta^*;z)\|^k]\leq \frac{1}{2}k!\E[\|\nabla\ell(\theta^*;z)\|^2] G_{*}^{k-2}.
\end{align}
\end{assumption}
We note that this assumption is very mild because $G_{*}$ only depends on gradients at $\theta^*$, and $G_{*}$ will only appear in the $O(n^{-2})$ terms in our theorems. Our approach also allows many other noise conditions at $\theta^*$ (e.g., those for heavy-tailed noise). At a high level, the order of our generalization error bounds only depends on the concentration of $\Pn\nabla\ell(\theta^*;z)$ relative to $\P \nabla\ell(\theta^*;z)$, which barely depends on noise at the single point $\theta^*$ and can be analyzed under various types of noise conditions. We introduce Assumption \ref{asm bernstein condition para} here because it leads to the asymptotically optimal problem-dependent parameter $\P[\|\nabla \ell(\theta^*;z)\|^2]$ and simplifies comparison with previous literature. 

Now we turn to establish problem-dependent rates under a curvature condition. While our  methodology is widely-applicable without restriction to particular curvature conditions, we will focus on the Polyak-Lojasiewicz (PL) condition, which  is known to be one of the weakest conditions that guarantee linear convergence of optimization algorithms \cite{karimi2016linear} as well as fast-rate generalization error \cite{foster2018uniform}.
\begin{assumption}[{\bf Polyak-Lojasiewicz  condition}]\label{asm pl}
There exist $\mu>0$ such that for all $\theta\in \Theta$,
\begin{align*}
    \P\ell(\theta;z)-\P\ell(\theta^*;z)\leq \frac{\|\P\nabla \ell(\theta;z)\|^2}{2\mu}.
\end{align*}
\end{assumption}
The PL condition is weaker than many others recently introduced in the areas of matrix recovery, deep learning, and learning dynamical systems, such as  ``one-point convexity" \cite{li2017convergence, kleinberg2018alternative},  ``star convexity" \cite{zhou2019sgd}, and ``quasar convexity" \cite{hinder2020near, hardt2018gradient}, not to mention the classical ``strong convexity." It is also referred to as ``gradient dominance condition" in previous literature \cite{foster2018uniform}. Under suitable assumptions on their inputs, many popular non-convex models have been shown to satisfy the PL condition (sometimes locally rather than globally). An incomplete list of these models includes: 
matrix factorization \cite{liu2016quadratic}, neural networks with one hidden layer \cite{li2017convergence}, ResNets with linear activations \cite{hardt2016identity}, binary linear classification \cite{mei2018landscape}, robust regression \cite{mei2018landscape}, phase retrieval \cite{sun2018geometric}, blind deconvolution \cite{li2019rapid}, linear dynamical systems \cite{hardt2018gradient}, mixture of two Gaussians \cite{balakrishnan2017statistical}, to name a few.

While the PL condition is known to be one of the weakest conditions that can be used to establish linear convergence to the global minimum (see \cite{karimi2016linear} for its relationship with other common curvature conditions), 
 the generalization aspects of such structural non-convex learning problems have not been fully understood. In particular, existing generalization error bounds often contain  global Lipchitz parameters that can be large for unbounded smooth losses. 

Our next theorem provides problem-dependent bounds for approximate stationary points of the empirical risk, under the PL condition of the population risk. The theorem implies that  {\it optimization} procedures that find
stationary points of the empirical risk are also {\it learning} algorithms for the population risk.
\begin{theorem}[{\bf generalization error of the approximate stationary point}]\label{thm pl}

 Under Assumptions \ref{asm hessian noise}, \ref{asm bernstein condition para} and \ref{asm pl}, $\forall \delta\in(0,1)$, with probability at least $1-\delta$, we have the following results: 

(a) there exist approximate stationary points of the empirical risk, $\widehat{\theta}\in\Theta$ such that 
\begin{align}\label{eq: stationary point}
\|\Pn \nabla\ell(\theta;z)\|\leq \sqrt{\frac{2\P[\|\nabla\ell(\theta^*;z)\|^2]\log\frac{4}{\delta}}{n}}+\frac{G_{*}\log\frac{4}{\delta}}{ n}.
\end{align} 

(b) for all $\hth$ that satisfy \eqref{eq: stationary point}, when
$
    n\geq \frac{c\beta^2}{\mu^2}\left(d+\log\frac{4\log (2n\Delta_M+1)}{\delta}\right)
$,
where $c$ is an absolute constant, we have
\begin{align*}
    \mathcal{E}(\hth)\leq  \frac{64\P[\|\nabla \ell(\theta^*;z)\|^2]\log\frac{4}{\delta}}{\mu n}+\frac{32 G_{*}^2\log^2\frac{4}{\delta}+4\mu^2}{\mu n^2}.
\end{align*}
\end{theorem}
Ignoring higher order terms and absolute constants, Theorem \ref{thm pl} implies a problem-dependent bound
\begin{align}\label{eq: parametric fast rate 1}
    \mathcal{E}(\hth)\leq O\left( \frac{\P[\|\nabla\ell(\theta^*;z)\|^2]\log\frac{1}{\delta}}{\mu n}\right),
\end{align}
which scales tightly with the problem-dependent parameter $\P[\|\nabla\ell(\theta^*;z)\|^2]$. The proof of Theorem \ref{thm pl algorithm} can be found in Appendix \ref{subsec proof fast algorithm}.
Optimality and implications of this problem-dependent rate will be discussed shortly after we present an additional theorem.

Many optimization algorithms, including  gradient descent \cite{karimi2016linear}, stochastic gradient descent \cite{ghadimi2013stochastic}, non-convex SVRG \cite{reddi2016stochastic} and  SCSG \cite{lei2017non} can efficiently find (approximate) stationary points of the empirical risk. However, convergence of these optimization algorithms is mostly studied  under assumptions on the  empirical risk.  The next theorem demonstrates that under assumptions on  the population risk, gradient descent provably achieve ``small" generalization error. These type of results are challenging to prove because properties of the population risk may not transfer to the empirical risk. 

Consider the gradient descent algorithm with fixed step size $\alpha$ and initialization $\theta^0$, generating  iterates according to
\begin{align}\label{eq: gradient descent scheme}
    \theta^{t+1}=\theta^t-\alpha\Pn\nabla\ell(\theta^t;z), \quad t=0,1,\dots
\end{align}

\begin{theorem}[{\bf generalization error of gradient descent}]\label{thm pl algorithm}
Assume Assumptions \ref{asm hessian noise}, \ref{asm bernstein condition para}, \ref{asm pl}. Then for an initialization $\theta^0\in\pazocal{\B}^d(\theta^*,\sqrt{\frac{\mu}{\beta}}\Delta_m)$ and step size $\frac{1}{\beta}$,  the iterates of the gradient descent algorithm \eqref{eq: gradient descent scheme} satisfy for any fixed $\delta\in(0,1)$,   with probability at least $1-\delta$,   for all $t=0,1,\dots$,
\begin{align}\label{eq: result theorem algorithm}
  \mathcal{E}(\theta^t)\leq \underbrace{\frac{16\P[\|\nabla\ell(\theta^*;z)\|^2]\log\frac{4}{\delta}}{\mu n}+\frac{8G_{*}^2\log^2\frac{4}{\delta}+\mu^2}{\mu n^2}}_{\textup{statistical error}}+(1-\frac{\mu}{2\beta})^t\mathcal{E}(\theta^0),
\end{align}
provided that the sample size $n$ is large enough such that  that the ``statistical error" term in \eqref{eq: result theorem algorithm} is smaller than $\frac{\mu}{2}\Delta_m^2$ and $n\geq \frac{c\beta^2}{\mu^2}\left(d+\log\frac{8\log_2(2n\Delta_M+2)}{\delta}\right)$, where $c$ is an absolute constant. 
\end{theorem}

Theorem \ref{thm pl algorithm} is the first broad scope result on the generalization error of gradient descent under the PL condition. It implies that after a logarithmic number of iterations, gradient descent achieves the problem-dependent rate \eqref{eq: parametric fast rate 1}. Note that  the algorithm only requires the initialization condition in the theorem, rather than any knowledge of $\Theta$ and the problem-dependent parameters. The proof of Theorem \ref{thm pl algorithm} can be found in Appendix \ref{subsec proof fast algorithm}, and the main idea is applicable to many other optimization algorithms as well. For example, in Section \ref{sec em} we provide a similar analysis to the first-order Expectation-Maximization algorithm.

\paragraph{Optimality of the problem-dependent rates in Theorem \ref{thm pl} and Theorem \ref{thm pl algorithm}.}
 It is well-known in the asymptotic statistics literature \cite{van2000asymptotic} that when $n$ tends to infinity, under mild local conditions,
 \begin{align}\label{eq: asymptotic rate}
      \sqrt{n}(\hat{\theta}_{\ERM}-\theta^*)\overset{d}{\rightarrow}N({\bf 0},H^{-1}QH^{-1}),
 \end{align} where $H=\mathbb{P}\nabla^2 \ell(\theta^*;z)$, $Q=\mathbb{P}[\nabla \ell(\theta^*;z)\nabla \ell(\theta^*;z)^T]$, and $\overset{d}{\rightarrow}$ means convergence in distribution. 
 The asymptotic rate \eqref{eq: asymptotic rate} is often information theoretically optimal under suitable conditions \cite{van1989asymptotic} (e.g., it matches the H{\'a}jek-Le Cam asymptotic minimax lower bound \cite{hajek1972local, le1972limits} when the loss $\ell(\theta;z)$ is a log likelihood function). The generalization error bounds in Theorem \ref{thm pl} and Theorem \ref{thm pl algorithm}, which are of the order
 \begin{align}\label{eq: parametric fast rate}
     \mathcal{E}(\hth)\leq O\left( \frac{\P[\|\nabla\ell(\theta^*;z)\|^2]\log\frac{1}{\delta}}{\mu n}\right),
 \end{align}
 are natural finite-sample versions of the ``ideal" asymptotic benchmark \eqref{eq: asymptotic rate}.

 In both Theorem \ref{thm pl} and Theorem \ref{thm pl algorithm}, the  sample complexity required to make the generalization error smaller than a fixed $\eps>0$ is 
\begin{align}\label{eq: overall sample complexity}
    n\geq  \Omega\left(\frac{\beta^2d}{\mu^2} + \frac{\P[\|\nabla\ell(\theta^*;z)\|^2]\log\frac{1}{\delta}}{\mu\eps}\right).
\end{align} Here we only consider the ``interesting" case of $\eps\in(0, \frac{\mu \Delta_m^2}{2})$ in Theorem \ref{thm pl algorithm}; otherwise the initialization point $\theta^0$ will satisfy $\mathcal{E}(\theta^0)\leq\eps$. Clearly, the sample complexity threshold $n\geq \Omega({\beta^2d}/{\mu^2})$ scales with the dimension $d$. This threshold is sharp up to absolute constants---there exists simple linear regression constructions where we require $\Omega({\beta^2d}/{\mu^2})$ samples \cite{litvak2005smallest} to make the empirical Hessian  positive definite. 
As a result, our overall sample complexity \eqref{eq: overall sample complexity} is essentially the sharpest result one can expect under the aforementioned assumptions.

\section{Applications to non-convex learning and stochastic optimization}\label{sec application fast rate regime}

In this section we will compare our problem-dependent rates with previous results from two areas: non-convex learning and stochastic optimization. 
Another topic that nicely illustrates the advantages of our approach,  Expectation-Maximization algorithms for missing data problems, is deferred to Section \ref{sec em}.

\subsection{Non-convex learning under curvature conditions}\label{sec nonconvex}
{In this subsection we discuss generalization error bounds for non-convex losses that satisfy the Polyak-Lojasiewicz (PL) condition. The PL condition is one of the weakest curvature conditions that have been rigorously and extensively studied in the areas of matrix recovery, deep learning, learning dynamical systems and learning mixture models. See our prior discussion under Assumption \ref{asm pl} and the reference thereof for representative models that satisfy this condition, and its relationship with other curvature conditions. The topic has attracted much recent attention because there is some empirical evidence suggesting that modern deep neural networks might satisfy this condition in large neighborhoods of global minimizers \cite{kleinberg2018alternative, zhou2019sgd}.
 
 For structured non-convex learning problems, a benchmark approach to prove generalization error bounds is    ``uniform convergence of gradients." \cite{mei2018landscape} presents the  ``uniform convergence of gradients" principle and proves dimension-dependent generalization error bounds to several representative problems; \cite{foster2018uniform} extends this idea to obtain norm-based generalization error bounds. We will compare our problem-dependent rates with these results.}
\paragraph{\bf Comparison with the results in Mei et al. \cite{mei2018landscape}.}
The main regularity assumptions imposed in \cite{mei2018landscape} include: 1) an assumption on the statistical noise of the Hessian matrices, whose content is similar to  our Assumption \ref{asm hessian noise}; and 2) an assumption that the random gradients $\nabla \ell(\theta;z)$ are $G-$sub-Gaussian for all $\theta\in\Theta$, which is not used in our framework (in contrast, we only impose a mild assumption on the gradient noise at  $\theta^*$). They also assume the Hessian is Lipchitz continuous which we view as redundant.

The theoretical foundation in \cite{mei2018landscape} is the following result on the (global) uniform convergence of gradients: with probability at least $1-\delta$,
\begin{align}\label{eq: uniform convergence gradient}
    \sup_{\theta\in\Theta}\|(\P-\Pn)\nabla \ell(\theta;z)\| \leq O\left(G\sqrt{\frac{ d\log n\log \frac{1}{\delta}}{n}}\right).
\end{align}
The sub-Gaussian parameter $G$ is  larger than the global Lipchitz constant, and can be quite large  in practical applications.
 From \eqref{eq: uniform convergence gradient}, when the population risk satisfies the PL condition, the generalization error for a stationary point $\hth$ of the empirical risk can be bounded as follows:
 \begin{align}\label{eq: mei18}
     \mathcal{E}(\hth)\leq O\left(\frac{G^2d\log n\log\frac{1}{\delta}}{\mu n}\right).
 \end{align}
\cite{mei2018landscape}  also provides {guarantees} for iterates of the gradient descent algorithm, but the analysis is specialized to the three applications considered in the paper. {It is worth mentioning that \cite{mei2018landscape} also studies the high-dimensional setting and provides a series of important results; we will not compare with those.}

Our approach improves both the result \eqref{eq: mei18} as well as the methodology \eqref{eq: uniform convergence gradient} as follows. 
\begin{itemize}
\item Our Theorem \ref{thm pl}  and Theorem \ref{thm pl algorithm} provide generalization error bounds for approximate stationary points and iterates of the gradient descent algorithm, which are of the order
\begin{align*}
    \mathcal{E}(\hth)\leq O\left(\frac{\P[\|\nabla\ell(\theta^*;z)\|^2]\log\frac{1}{\delta}}{\mu n}\right)
\end{align*}
Focusing on the the dominating $O(n^{-1})$ term, our new result replaces $O( G^2 d\log n)$ in the numerator with the typically much smaller localized quantity $\P\|\nabla\ell(\theta^*;z)\|^2$. In fact,  from the definition of sub-Gaussian vectors, one can prove (see, e.g. \cite{vershynin2018high}) that  
\begin{align*}
     \P[\|\nabla\ell(\theta^*;z)\|^2]\ll\sup_{\Theta}\P\|\nabla\ell(\theta;z)\|^2\leq 8G^2 d,
 \end{align*} 
 so our bounds are always more favorable than \eqref{eq: mei18}. In passing, we also  remove a superfluous $\log n$ factor by using generic chaining rather than simple discretization.

\item  Our Proposition \ref{prop localized gradient 1} on the {\it localized} uniform convergence of  gradients,
 \begin{align*}
     \|(\P-\Pn) (\nabla \ell(\theta;z)-\nabla \ell(\theta^*;z))\|\leq
     O\left( \beta\|\theta-\theta^*\| \sqrt{\frac{d+\log\frac{2\log (2n\Delta_M)}{\delta}}{n}}\right),
 \end{align*}
strengthens  \eqref{eq: uniform convergence gradient} to a localized result, under fewer assumptions.
 \end{itemize}
 
 We illustrate our improvements  on a particular non-convex learning example.
 \begin{example}[{\bf non-convex regression with non-linear activation}]\label{example one point convexity}
Consider the model 
    \begin{align}\label{eq: regression nonlinear}
        \ell(\theta;z)= (\eta(\theta^Tx)-y)^2,
    \end{align}
    where $\eta(\cdot)$ is a non-linear {\it activation function}, and there exists $\theta^*\in\Theta$ such that $\E[y]=\eta(x^T\theta^*)$. 
    This model has been empirically shown to be superior relative to convex  formulations in several applications \cite{chapelle2009tighter, laska2012regime, nguyen2013algorithms}, and is  representative of the class of one-hidden-layer neural network models.   Popular choices of $\eta$  include sigmoid link $\eta(t)=(1+e^{-t})^{-1}$ and probit link $\eta(t)=\Phi(t)$ where $\Phi$ is the Gaussian cumulative distribution function. Under  mild regularity conditions,  the population risk $\P\ell(\theta;z)$  satisfies the PL condition and the statistical noise conditions.
  \end{example}    
    
    \begin{assumption}[{\bf regularity conditions for Example \ref{example one point convexity}}]\label{asm regression nonlinear}
   (a)  $\|x\|_{\infty}$ is uniformly bounded by $\tau$, the feasible parameter set $\Theta$ is given by $\{\theta\in \R^d:\|\theta\|\leq \frac{\Delta_M}{2}\}$, and $B$ is the worst-case boundedness parameter of $(\eta(\theta^Tx)-y)^2$ (which can scales with $n$). (b)  there exists $C_{\eta}>0, c_{\eta}>0$ such that $$\sup_{|t|\leq \Delta_M\tau\sqrt{d}}\max\{\eta'(t), \eta''(t)\}\leq C_{\eta}, \quad \inf_{|t|\leq \Delta_M\tau\sqrt{d}}\eta'(t)\geq c_{\eta}.$$ (c) The feature vector $x$ spans all directions in $\R^d$, that is, $\E[xx^T]\succeq \gamma \tau^2 {I}_{d\times d}$ for some $0<\gamma<1$.
    \end{assumption}
    
   Under Assumption \ref{asm regression nonlinear}, all of our proposed assumptions in Theorem \ref{thm pl} and Theorem \ref{thm pl algorithm} are satisfied. In particular: Assumption \ref{asm hessian noise} holds with $\beta=2C_{\eta}(C_{\eta}+\sqrt{B})\tau^2$; Assumption \ref{asm bernstein condition para} holds with $G_{*}=2C_{\eta}\tau\sqrt{Bd}$; and Assumption \ref{asm pl} holds with $\mu=\frac{2c_{\eta}^3\tau^2\gamma}{C_{\eta}}$ (see Appendix \ref{subsec proof application fast rate} for details). Let $\hth$ be the approximate stationary point in Theorem \ref{thm pl}, or the output of the gradient descent algorithm in Theorem \ref{thm pl algorithm} (after running sufficiently many iterations), we have the following corollary.
   
\begin{corollary}[{\bf generalization error bound for Example \ref{example one point convexity}}] \label{coro regression nonlinear}
 Under Assumption \ref{asm regression nonlinear}, 
 \begin{align}\label{eq: our rate regression nonlinear}
  \mathcal{E}(\hth)\leq O\left(  \frac{ d\LL (\tau C_{\eta})^2\log\frac{1}{\delta}}{\mu n}\right),
 \end{align} where $\LL:=\P[(y-\eta({\theta^*}^Tx))^2]$ is the population risk at the optimal parameter. $\theta^*$.
\end{corollary}

Since the sub-Gaussian parameter of the random gradient satisfies $G\leq O(\tau C_{\eta}\sqrt{B})$ under Assumption \ref{asm regression nonlinear}, the result \eqref{eq: mei18} in Mei et al. \cite{mei2018landscape} implies a generalization error bound of the order 
\begin{align}\label{eq: mei18 example}
 \mathcal{E}(\hth)\leq O\left( \frac{  dB (\tau C_{\eta})^2\log n\log\frac{1}{\delta}}{\mu n}\right) \quad [\text{existing result \cite{mei2018landscape}}],
 \end{align} 
 where $B=\sup_{\theta,x,y} (\eta(\theta^Tx)-y)^2$ is the worst-case boundedness parameter.
 Let us now compare our result \eqref{eq: our rate regression nonlinear} with the the existing result \eqref{eq: mei18 example}: 1) our result \eqref{eq: our rate regression nonlinear} improves the worst-case boundedness parameter $B$, replacing it with the much smaller optimal risk $\LL$; and 2) it removes the superfluous logarithmic factor $\log n$.

 \paragraph{\bf Comparison with the norm-based generalization error bound in Foster et al. \cite{foster2018uniform}.}
Let us now compare our problem-dependent rates with the norm-based bounds in   \cite{foster2018uniform} ({it is worth mentioning that they also provide novel results in the infinite-dimensional and high-dimensional settings}). Under the formulation in Example \ref{example one point convexity} and Assumption \ref{asm regression nonlinear}, the generalization error bound proved in \cite{foster2018uniform}   is of the order
\begin{align}\label{eq: foster18 example}
    \mathcal{E}(\hth)\leq O\left( \frac{d^2 B (\tau C_{\eta})^4+dB(\tau C_\eta)^2\log\frac{1}{\delta}}{\mu n}\right) \quad [\text{{converted from} \cite{foster2018uniform}}],
\end{align}
  for an approximate stationary point $\hth$. To be specific, their original result assumes $\|x\|$ to be uniformly bounded by $D$ and the generalization error scales with $D^4$. Under the standard assumption $\|x\|_{\infty}\leq \tau$ (or $x$ being a $\tau-$sub-Gaussian random vector), $D^4$ is of order $\tau^4d^2$ so their result does not achieve optimal dependence on $d$ (in the original statements in \cite{foster2018uniform} there is a potential misunderstanding of the dependence  on $d$). Besides improving the worst-case boundedness parameter $B$ to the optimal risk $\LL$, our result \eqref{eq: our rate regression nonlinear} further improves \eqref{eq: foster18 example} by  order  $d(\tau C_{\eta})^2$. 

Lastly, we comment that there is no formal guarantee on how to find $\hth$ by an optimization algorithm in \cite{foster2018uniform}. They merely establish the generalization error bound for approximate stationary points, but  analysis of an optimization algorithm is more challenging because properties of the population risk may not carry over to the empirical risk.

\subsection{Stochastic optimization}\label{subsec stochastic optimization}
The  parametric learning setting we discussed is sometimes referred to as ``stochastic optimization" \cite{shalev2009stochastic, shalev2010learnability}. Beyond supervised learning, stochastic optimization also covers operations research and system control problems, where the dimension $d$ may no longer be pertinent in the generalization error bound for sufficiently large $n$ (i.e., the bound should be dimension-independent in the asymptotic regime). Therefore, it is  preferable to prove norm-based generalization error bounds, which do not explicitly scale with $d$ after a certain sample complexity threshold. 

 We compare our results with previous ones in the area of stochastic optimization. Those results typically assume the population risk to be strongly convex, i.e.,
 there exists $\mu>0$ such that $\forall \theta_1, \theta_2\in\Theta$,
\begin{align*}
   \P\ell(\theta_1;z)-\P\ell(\theta_2;z)- \left(\P\nabla\ell(\theta_2;z)\right)^T(\theta_1-\theta_2) \geq \frac{\mu}{2}\|{\theta_1}-{\theta_2}\|^2.
\end{align*} While this assumption is much more restrictive than our Assumption \ref{asm pl}, we note that our problem-dependent rate and sample complexity results are novel even in this well-studied strongly convex setting.

Recall that our problem-dependent generalization error bounds in Theorem \ref{thm pl} and Theorem \ref{thm pl algorithm} are of the order
 \begin{align}\label{eq: problem dependent stochastic optimization}
     \mathcal{E}(\hth)\leq O\left( \frac{\P[\|\nabla\ell(\theta^*;z)\|^2]\log\frac{1}{\delta}}{\mu n}\right),
 \end{align}
provided $n\geq \Omega({\beta^2d}/{\mu^2})$; and the sample complexity (to achieve $\eps$ generalization error) is
\begin{align}\label{eq: our sample complexity}
    n\geq  \Omega\left(\frac{\beta^2d}{\mu^2}+ \frac{\P[\|\nabla\ell(\theta^*;z)\|^2]\log\frac{1}{\delta}}{\mu\eps}\right).
\end{align}
 Our results are natural finite-sample extensions of the classical ``asymptotic normality" result \eqref{eq: asymptotic rate}, and hence are the sharpest results one can expect under aforementioned assumptions (see the discussion at the end of Section \ref{sec parametric}).

\paragraph{\bf Comparison with the classical result from Shapiro et al.  \cite{shapiro2009lectures}.}
Perhaps the most representative result on the generalization error of empirical risk minimization (also referred to as ``sample average approximation") in the stochastic optimization literature, is Corollary 5.20 from the monograph \cite{shapiro2009lectures}. When the random gradient $\nabla \ell(\theta;z)$ is $G-$sub-Gaussian for all $\theta \in\Theta$, that result implies  \begin{align}\label{eq: shapiro}
   \mathcal{E}(\hth_{\ERM})\leq O\left({\frac{G^2 {d}\log n\log\frac{1}{\delta}}{\mu n}}\right).
\end{align}
One advantage of \eqref{eq: shapiro} is that it does not require the population risk to be smooth. However,  the explicit dependence on $d$ and the global sub-Gaussian parameter $G$ in \eqref{eq: shapiro} make it less favorable for some operations research applications and M-estimation problems, where the asymptotic complexity 
\eqref{eq: asymptotic rate} does not depend on $d$. 
It is easy to show that our problem-dependent generalization error bound \eqref{eq: problem dependent stochastic optimization}
 strictly improves on this classical result. Specifically,  under the sub-Gaussian distributional assumptions on gradients, one can prove that 
 $$ \P[\|\nabla\ell(\theta^*;z)\|^2]\ll\sup_{\theta\in\Theta}\P[\|\nabla \ell(\theta;z)\|^2]\leq 8d G^2.$$ Plugging this into \eqref{eq: problem dependent stochastic optimization} and \eqref{eq: shapiro}, we observe  that our bound  improves on \eqref{eq: shapiro} by removing dependence on the worst-case $L_2$ norm of the gradient over the entire parameter space $\Theta$.

\paragraph{\bf Comparison with results obtained from the ``online to batch conversion" \cite{kakade2009generalization}.} 
By assuming the population risk is strongly convex and satisfies the following ``uniform Lipchitz continuous" condition,
 \begin{align*}
    |\ell(\theta_1;z)-\ell(\theta_2;z)|\leq L_{\text{unif}}\|\theta_1-\theta_2\|, \quad \forall z\in \pazocal{Z},\quad \forall \theta_1,\theta_2\in \Theta,
\end{align*} \cite{kakade2009generalization} proves an ``online to batch conversion" that relates the regret of an algorithm (on past data) to the generalization performance (on future data). As a result, the output $\hth_{SGD}$ of certain stochastic gradient methods (also referred to as ``stochastic approximation" in the stochastic optimization literature) can be proved to satisfy
\begin{align}\label{eq: SGD}
    \mathcal{E}(\hth_{\text{SGD}})\leq O\left(\frac{G_{\unif}^2\log\frac{1}{\delta}}{n}\right).
\end{align}
Note that  \eqref{eq: SGD} does not require any sample size threshold. In contrast, our problem-dependent generalization error bound \eqref{eq: problem dependent stochastic optimization} provides an improved rate, but only  as long as the sample size condition $n\geq \Omega({\beta^2d}/{\mu^2})$ is satisfied, because in this case   $$\P[\|\nabla\ell(\theta^*;z)\|^2]\ll \sup_{\theta\in\Theta,z\in\pazocal{Z}}\|\nabla\ell(\theta;z)\|^2= L_{\unif}^2.$$
 Plugging this into \eqref{eq: problem dependent stochastic optimization} and \eqref{eq: SGD}, this claimed   improvement can be  immediately verified.

\paragraph{\bf Comparison with problem-dependent bounds in Zhang et al. \cite{zhang2017empirical}.} 
By imposing both strong convexity and a uniform smoothness condition,   \cite{zhang2017empirical}  systematically provide  dimension-independent generalization error bounds for empirical risk minimization.  However, there are several limitations in their approach:
1) their sample complexity threshold to achieve dimension-independent generalization error is sub-optimal for many popular problems; 
and 2) many of their assumptions are restrictive and not required (as our analysis shows).

The main source of the limitations in \cite{zhang2017empirical} is the assumption that $\ell(\theta;z)$ admits a uniform smooth parameter $\beta_{\text{unif}}$, i.e., 
 \begin{align}\label{eq: unif smooth}
    \|\nabla\ell(\theta_1,z)-\nabla\ell(\theta_2,z)\|\leq \beta_{\text{unif}}\|\theta_1-\theta_2\| , \quad \forall z\in\Z, \quad \forall \theta_1,\theta_2\in \Theta.
 \end{align} This quantity serves as the main complexity proxy. With additional assumptions that $\ell(\theta,z)$ is non-negative and convex for all $z$,  [\citealp{zhang2017empirical}, Theorem 3]  proves that when
 \begin{align*}
     n\geq \Omega\left(\frac{\beta_{\textup{unif}}d\log n}{\mu}\right),
 \end{align*} empirical risk minimization achieves the problem-dependent bound
 \begin{align*}
     \mathcal{E}(\hth_{\ERM})\leq O\left(\frac{\beta_{\unif}\LL\log\frac{1}{\delta}}{\mu n}\right).
 \end{align*}
 However, as $\beta_{\text{unif}}$ is effectively the largest value of the operator norm of the Hessian---$\sup_{\theta;z}\|\nabla^2\ell(\theta;z)\|_{\text{op}}$, it scales with dimension $d$  in most statistical estimation problems. As a result,  the sample complexity threshold $\Omega({\beta_{\textup{unif}}d\log n}/{\mu})$ becomes sub-optimal for most statistical regression problems. For example, consider a simple linear regression set up:
\begin{align}\label{eq: linear regression}
  \ell(\theta;(x,y))=\|y-\theta^Tx\|^2, \quad   y=x^T\theta^*+\upsilon, \quad \upsilon\sim N(0,1), \quad x\sim N(0, \tau{{I}}_{d\times d}).
\end{align}
 In this example, we have $\mu=1$ and
$\beta_{\unif}=\Omega(\tau^2d )$
, so the sample complexity needed to achieve $\eps$ accuracy in \cite{zhang2017empirical} is 
  \begin{align}\label{eq: stochastic optimization zhang}
     n\geq  {\Omega}\left(\tau^2d^2\log n +\frac{d\LL\tau^2}{\eps}\right) \quad  [\text{sample complexity  \cite{zhang2017empirical}}],
  \end{align} 
In contrast, our sample complexity \eqref{eq: our sample complexity} is 
  \begin{align*}
     n\geq  {\Omega}\left(\tau^2 d +\frac{d\LL\tau^2}{\eps}\right) \quad [\text{sample complexity \eqref{eq: our sample complexity}}],
  \end{align*} 
  in this example.
Therefore,  the ${\Omega}(\tau^2d^2\log n )$ term in \eqref{eq: stochastic optimization zhang}  is sub-optimal primarily because the analysis in \cite{zhang2017empirical} relies on the uniform smoothness parameter $\beta_{\unif}$.
Moreover,  their assumptions that $\ell(\theta;z)$ is non-negative and convex for all $z$  may rule out interesting stochastic optimization applications.
These are not required by our framework and results.

Recently, the approach in \cite{zhang2017empirical} was also extended to analyze a variant of the stochastic gradient descent algorithm \cite{zhang2019stochastic}, but the results hold only in expectation rather than with high probability, and they have similar limitations to the results in \cite{zhang2017empirical}. This approach has also been extended to non-convex stochastic optimization problems \cite{liu2018fast}, where the generalization error bounds are of the from \eqref{eq: SGD}---they contain the uniform Lipchitz parameter $L_{\text{unif}}$ and are not problem-dependent.

 \section{Learning with missing data and Expectation-Maximization algorithms}
\label{sec em}
In this section we introduce a broad class of applicable non-convex and semi-supervised learning problems, in the area of ``learning with missing data." We again apply our proposed ``uniform localized convergence" framework and prove a variant of Theorem \ref{thm pl algorithm}, which gives the sharpest local convergence rate for first-order Expectation-Maximization (EM) algorithms  in many widely studied problems. Our analysis improves the  framework introduced recently in Balakrishnan et al. \cite{balakrishnan2017statistical}.

\subsection{Background}

Convex maximum likelihood estimation problems will generally become non-convex when there is missing or unobserved data.  Assume the data $(z,w)$ is generated from an unknown distribution specified by the true parameter $\theta^*\in\R^d$, where $z\in\pazocal{Z}$ corresponds to the observable data, and $w \in \pazocal{W}$ corresponds to the unobservable data (also referred to as the ``latent variable"). For every $\theta\in\R^d$, let $f_{\theta}(z,w)$ be the likelihood of observing $z$ conditioned on  $w$, if the underlying distribution is specified by $\theta$. (Throughout this section we will assume the existence of density functions for simplicity.) Consider the loss function
\begin{align}\label{eq: em model}
    \ell(\theta;z)=-\log\left[\int_{\pazocal{W}}f_{\theta}(z,w)d{w}\right].
\end{align}
  Our goal is to estimate the true parameter $\theta^*$, which minimizes the population risk $$\P\ell(\theta;z)=\int_{\pazocal{Z}}\ell(\theta;z)dz$$ over all $\theta\in\R^d$. (Equivalently, $\theta^*$ maximizes the population log-likelihood function.) The main challenge is that $\P\ell(\theta;z)$ is typically non-convex, despite the fact that the conditional log-likelihood function $\log f_{\theta}(z,w)$ would usually be convex with respect to $\theta$, if both $z$ and $w$ were observable. 

The following $\ell_{\theta'}(\theta;z)$ function provides a convex upper bound on $ \ell(\theta;z)$, and can be interpreted as the conditional expectation of the loss,  as if  $\theta'$ is the true parameter $\theta^*$:
\begin{align*}
    \ell_{\theta'}(\theta;z)=-\int_{\pazocal{W}} k_{\theta'}(w|z)\log f_{\theta}(z,w)dw,
\end{align*}
where $k_{\theta'}(w|z)$ is the  conditional density of $w$ given $z$. Denote 
$
    \nabla_{\theta} \ell_{\theta'}(\theta;z)$
 the gradient of $\ell_{\theta'}(\theta;z)$ when fixing $\theta'$. From the definition of $k_{\theta'}(w|z)$, it is easy to verify that the vector-value of $
    \nabla_{\theta} \ell_{\theta'}(\theta;z)$ at $\theta'$ is exactly the gradient of $\ell(\theta';z)$: that is, for all $\theta,\theta'\in \Theta$,
\begin{align}\label{eq: em identity}
    \nabla_{\theta} \ell_{\theta'}(\theta;z)|_{\theta=\theta'}=\nabla\ell(\theta';z).
\end{align}
In view of the identity \eqref{eq: em identity}, it is known \cite{balakrishnan2017statistical}  that  gradient descent on the empirical risk $\Pn \ell(\theta;z)$ is equivalent to the first-order Expectation-Maximization algorithm: at the $t-$th iteration, the ``expectation" step calculates the sample average $\Pn \ell_{\theta^t}(\theta;z)$, and the ``maximization" step executes the first-order update
\begin{align}\label{eq: foem update}
    \theta^{t+1}=\theta^t-\alpha \nabla \Pn\nabla \ell_{\theta^t}(\theta^t;z)=\theta^t-\alpha \nabla \Pn\nabla \ell(\theta^t;z),
\end{align}
where $\alpha>0$ is the step size. First-order EM is known to be more computationally efficient than standard EM, and more amendable for analysis \cite{balakrishnan2017statistical}.

Examples of learning with missing data problems for which the above observations apply include the followings.
 \begin{example}[{\bf Mixture of two  Gaussians}]\label{example mixture of Gaussian}
In this problem, the missing variable $w\in\{-1,1\}$ is an indicator of the underlying mixture component, which has $\frac{1}{2}$ probability to be $1$ and the other $\frac{1}{2}$ probability to be $-1$. Conditioned on $w$, the observable variable $z$ is generated as follows.
\begin{align*}
  (z|w=1)\sim  N(\theta^*,\sigma^2I_{d\times d}), \quad (z|w=-1)\sim  N(-\theta^*,\sigma^2I_{d\times d}).
\end{align*} For this problem, we have
\begin{align*}
    \ell_{\theta'}(\theta;z)=\frac{w_{\theta'}(z_i)}{2}\|z_i-\theta\|^2+\frac{(1-w_{\theta'}(z_i))}{2}\|z_i+\theta\|^2,
\end{align*}
where $w_{\theta'}(z)=e^{-\frac{\|\theta'-z\|^2}{2\sigma^2}}[e^{-\frac{\|\theta'-z\|^2}{2\sigma^2}}+e^{-\frac{\|\theta'+z\|^2}{2\sigma^2}}]^{-1}$.
 \end{example}
 
 \begin{example}[{\bf Mixture of two component linear regression}]\label{example mixture regression}
In this problem, $x\sim N(0,I_{d\times d})$ is a random feature vector, and $w\in\{-1,1\}$ is a missing indicator variable that has $\frac{1}{2}$ probability to be $1$ and  $\frac{1}{2}$ probability to be $-1$. Conditioned on $w$ and $x$, the label variable $y$ is generated as follows.
\begin{align*}
  (y|w=1,x)\sim  N(x^T\theta^*,\sigma^2), \quad (y|w=-1,x)\sim  N(-x^T\theta^*,\sigma^2).
\end{align*} In this problem, the observable variable $z$ is the feature-label pair $(x,y)$, and we have
\begin{align*}
   \ell_{\theta'}(\theta;z)=\frac{w_{\theta'}(x,y)}{2}(y-x^T\theta)^2+\frac{1-w_{\theta'}(x,y)}{2}(y+x^T\theta)^2,
\end{align*}
where $w_{\theta'}(x,y)=e^{-\frac{(x^T\theta'-y)^2}{2\sigma^2}}[e^{-\frac{(x^T\theta'-y)^2}{2\sigma^2}}+e^{-\frac{(x^T\theta'+y)^2}{2\sigma^2}}]^{-1}$.
 \end{example}
 
 \subsection{Problem-dependent rates for first-order EM}

Motivated by the breakthrough work Balakrishnan et al. \cite{balakrishnan2017statistical}, we assume that the feasible parameter space $\Theta$ contains the true parameter $\theta^*$, and satisfies the two assumptions.
\begin{assumption}[{\bf strong convexity of $\bm{\P\ell_{\theta^*}(\theta;z)}$}]\label{asm em strongly convex}
There exists $\mu_1>0$ such that $\forall \theta_1,\theta_2\in\Theta$
\begin{align*}
    \P\ell_{\theta^*}(\theta;z)-\P\ell_{\theta^*}(\theta^*;z)\leq \frac{\|\P\nabla\ell_{\theta^*}(\theta;z)\|^2}{2\mu_1}.
    \end{align*}
\end{assumption}
Recall that $\P\ell_{\theta^*}(\theta;z)$ is the underlying ``true" log likelihood with respect to parameter $\theta$, which is unknown due to lack of information on $\theta^*$. It is standard to assume that $\P\ell_{\theta^*}(\theta;z)$ is a strongly convex when there is no missing data \cite{mclachlan2007algorithm, balakrishnan2017statistical}.
\begin{assumption}
[{\bf gradient smoothness}] \label{asm em gradient smoothness} There exists $0<\mu_2<\mu_1$ such that  $\forall \theta\in \Theta$
\begin{align*}
    \|\P\nabla\ell_{\theta}(\theta;z)-\P\nabla\ell_{\theta^*}(\theta;z)\|\leq \mu_2\|\theta-\theta^*\|.
\end{align*}
\end{assumption}
  Assumption \eqref{asm em gradient smoothness} is also assumed in  \cite{balakrishnan2017statistical}. While this assumption does not typically hold over all $\theta\in\R^d$,   it is often satisfied with small enough $\mu_2$ over a local region around the true parameter $\theta^*$ \cite{balakrishnan2017statistical, wang2014high}.  Under the above two assumptions, according to the identity \eqref{eq: em identity}, the gradient of the population risk,
\begin{align*}
    \P\nabla\ell(\theta;z)=\P\nabla\ell_\theta(\theta;z),
\end{align*} can be viewed as a perturbation  of $\nabla \ell_{\theta^*}(\theta;z)$---the gradient of the strongly convex function $\P\ell_{\theta^*}(\theta;z)$.  Therefore,  Assumption \ref{asm em strongly convex} and Assumption \ref{asm em gradient smoothness} play a similar  role to that of the PL condition that we have analyzed in Section \ref{subsec problem dependent fast}. The following theorem can be viewed as a modification of our previous Theorem \ref{thm pl algorithm}, where the proof is tailored to these new assumptions but the key ideas remain mostly similar.

\begin{theorem}[{\bf generalization error of first-order EM}]\label{thm fo em}
 Assume Assumptions \ref{asm hessian noise}, \ref{asm bernstein condition para}, \ref{asm em strongly convex}, \ref{asm em gradient smoothness}, and assume  access to an initialization  $\theta^0\in\pazocal{B}^d(\theta^*,\Delta_m)$. For any fixed $\delta\in(0,1)$, iterates of the first-order EM algorithm $\{\theta^t\}$ generated by  \eqref{eq: foem update} with the fixed step size $\frac{2}{\beta+\mu_1}$ 
  satisfy with probability at least $1-\delta$ and  all $t=0,1,\dots$,
\begin{align}\label{eq: em estimation error}
\mathcal{E}(\theta^t)&\leq \frac{16\beta}{\mu_1^2}\left(\sqrt{\frac{2\P[\|\nabla\ell(\theta^*;z)\|^2]\log\frac{4}{\delta}}{n}}+\frac{G_{*}\log\frac{4}{\delta}+\mu_1}{n}\right)^2 +\left(1-\frac{2\mu_1-\mu_2}{2(\beta+\mu_1)}\right)^{2t}\beta\|\theta^0-\theta^*\|^2, \quad \text{and}\nonumber\\
  \|\theta^t-\theta^*\| 
    &\leq \frac{4}{\mu_1}\left(\sqrt{\frac{2\P[\|\nabla\ell(\theta^*;z)\|^2]\log\frac{4}{\delta}}{n}}+\frac{G_{*}\log\frac{4}{\delta}+\mu_1}{n}\right)+\left(1-\frac{2\mu_1-\mu_2}{2(\beta+\mu_1)}\right)^t\|\theta^0-\theta^*\|, 
\end{align}
provided the sample size condition
$
         n\geq \max\left\{ \frac{c\beta^2}{\mu_1^2}\left(d+\log\frac{8\log_2(2n\Delta_M+2)}{\delta}\right), \frac{128\P[\|\nabla\ell(\theta^*;z)\|^2]\log\frac{4}{\delta}}{\mu_1\Delta_M}, \frac{8G_*\log\frac{4}{\delta}+8\mu_1}{\mu_1\Delta_M} \right\}
$ holds, 
 where $c$ is an absolute constant.
\end{theorem}

We comment that it is usually straightforward to verify Assumption \ref{asm hessian noise} for a specific missing data applications (no harder than verifying it on the completely observable case). From Lemma \ref{lemma hessian}, we only need to show that the hessian matrices $\{\frac{\partial^2}{\partial \theta^2}[-\log f_{\theta}(z,w)]\}_{\theta\in\Theta}$ are sub-exponential for all fixed $w$ and $\theta$. That is,
\begin{align}\label{eq: hessian em}
    \E\left\{\exp\left(\frac{1}{\beta}\left|u_1^T\left(\frac{\partial^2[\log f_{\theta}(z,w)]}{\partial \theta^2}\right)u_2\right|\right)\right\}\leq 2,
\end{align}for any unit vectors $u_1,u_2$, any $z$ and any $\theta\in \Theta$. Typically, condition \eqref{eq: hessian em} simply requires that the {\it observable} data $z$ be a  sub-Gaussian vector, regardless of the (fixed) values the unobservable data $w$ take.

As a result, Theorem \ref{thm fo em}  applies to a broad class of ``learning with missing data" problems, including  Example \ref{example mixture of Gaussian} and Example \ref{example mixture regression}. In order to validate Assumption \ref{asm em gradient smoothness} on these two examples, a common strategy \cite{balakrishnan2017statistical} is to assume the signal-to-noise ratio (SNR) to be lower bounded as
\begin{align}\label{eq: snr condition}
    \frac{\|\theta^*\|}{\sigma}\geq \eta,
\end{align}
 for some absolute constant $\eta>0$. The following corollary holds  under identical assumptions on $\eta$ as in \cite{balakrishnan2017statistical}.

\begin{corollary}[{\bf Theorem \ref{thm fo em} applied to Example \ref{example mixture of Gaussian} and Example \ref{example mixture regression}}]\label{coro application example em} In both Example \ref{example mixture of Gaussian} and Example \ref{example mixture regression}, 
 after sufficiently many iterations, the first-order EM algorithm  with step size $1$ satisfies the generalization error bound
\begin{align*}
    \mathcal{E}(\theta^t)\leq O\left(\frac{\sigma^2d\log\frac{1}{\delta}}{n}\right).
\end{align*}
Specifically,
\begin{itemize}
\item For the Gaussian mixture model (Example \ref{example mixture of Gaussian}), assuming the signal-to-noise condition \eqref{eq: snr condition} holds, and the initialization point satisfies $\theta^0\in\pazocal{\B}^d(\theta^*,\frac{\|\theta^*\|}{4})$, 
then the result of Theorem \ref{thm fo em} holds with $\beta=1$,  $G_{*}=\sigma\sqrt{d}$,  $\mu_1=1$ and $\mu_2=c_1(1+\frac{1}{\eta^2}+\eta^2)e^{-c_2\eta^2}$, where $c_1,c_2$ are absolute constants.

\item For the mixture of linear regression model (Example \ref{example mixture regression}), assuming the signal-to-noise condition \eqref{eq: snr condition}, and the initialization point satisfies  $\theta^0\in\pazocal{B}^d(\theta^*; \frac{\|\theta^*\|}{32})$, then the result of Theorem \ref{thm fo em} holds with $\beta=1$,  $G_{*}=\sigma\sqrt{d}$,  $\mu_1=1$ and $\mu_2=\frac{1}{4}$.
\end{itemize}
\end{corollary}

Proofs can be found in Appendix \ref{subsec proof application fast rate em}. Notably, our results do not depends on $\|\theta^*\|$, and hence refine those in \cite{balakrishnan2017statistical}.

\subsection{Discussion and improvements over previous results}
In this subsection, we will first compare our general theory with the methodology in Balakrishnan et al. \cite{balakrishnan2017statistical}. Then, we will discuss the  improvement over  \cite{balakrishnan2017statistical} as illustrated in Example \ref{example mixture of Gaussian} and Example \ref{example mixture regression}. Lastly, we will provide some intuition pertaining to this improvement. 

\paragraph{Improvements in the methodology.}
We now restate the theoretical result from Balakrishnan et al. \cite{balakrishnan2017statistical} on the estimation error of first-order EM. Assume with probability at least $1-\delta$, 
\begin{align}\label{eq: unif fo em}
     \sup_{\Theta}\|(\P-\Pn)\nabla \ell_{\theta}(\theta;z)\|\leq \eps_{\unif}(n,\delta).
\end{align}When the sample size $n$ is large enough, \cite{balakrishnan2017statistical} proves that the first-order EM iterates $\{\theta^t\}_{t=0}^{\infty}$ satisfy 
\begin{align}\label{eq: fo em}
  \|\theta^t-\theta^*\|\leq O\left( \frac{\eps_{\unif}(n,\delta)}{\mu_1-\mu_2}\right)+\left(1-\frac{2\mu_1-\mu_2}{\beta+\mu_1}\right)^{t}\|\theta^0-\theta^*\|.
\end{align}
Compared with \eqref{eq: em estimation error} in Theorem \ref{thm fo em}, the approach in \cite{balakrishnan2017statistical} has two main limitations.
\begin{itemize}
\item The result \eqref{eq: fo em}  contains a loose, global uniform convergence terms $\eps_{\unif}(n,\delta)$ defined via \eqref{eq: unif fo em}. In contrast, our Theorem \ref{thm pl algorithm} suggests that the statistical error only depends on the concentration of $\Pn \nabla\ell(\theta^*;z)$ relative to $\P\nabla\ell(\theta^*;z)$ at the single point $\theta^*$. The precise improvement will be illustrated on Example \ref{example mixture of Gaussian} and Example \ref{example mixture regression} shortly.

\item \cite{balakrishnan2017statistical} does not discuss how to calculates the complex uniform convergence term $\eps^{\unif}(n,\delta)$ for general models. In fact, \cite{balakrishnan2017statistical} only calculate this term for  Example \ref{example mixture of Gaussian}. For the rest of the applications they consider, they turn to analyze sample-splitting heuristics. Although these  heuristics are easier to analyze, they are less common in practice. In contrast, our Theorem \ref{thm fo em} applies to general models without leaving the uniform convergence term unspecified.
\end{itemize}

\paragraph{Improvements on the examples.}
For the mixture of two Gaussians (Example \ref{example mixture of Gaussian}), \cite{balakrishnan2017statistical} proves that after sufficiently many iterations,   the first-order EM algorithm satisfies the generalization error bound
\begin{align}\label{eq: balak gaussian}
\mathcal{E}(\theta^t)\leq O\left(  \frac{\|\theta^*\|^2(1+\frac{\|\theta^*\|^2}{\sigma^2})d\log\frac{1}{\delta}}{n}\right) \quad [\text{GMM result \cite{balakrishnan2017statistical}}],
\end{align}
and for the mixture of regressions (Example \ref{example mixture regression}), \cite{balakrishnan2017statistical} proves that after sufficient iterations, the first-order EM algorithm satisfies the generalization error bound
\begin{align}\label{eq: balak regression}
\mathcal{E}(\theta^t)\leq O\left(\frac{(\sigma^2+\|\theta^*\|^2)d\log\frac{1}{\delta}}{n}\right) \quad [\text{regression result \cite{balakrishnan2017statistical}}].
\end{align}
In contrast, our problem-dependent generalization error bounds given by  Corollary \ref{coro application example em} are of the order
\begin{align*}
 \mathcal{E}(\theta^t)\leq O\left( \frac{\sigma^2d\log\frac{1}{\delta}}{n}\right) \quad [\text{Corollary \ref{coro application example em}}],
\end{align*}
which exhibits an improvement over the previous results \eqref{eq: balak gaussian} \eqref{eq: balak regression} from \cite{balakrishnan2017statistical}, under  identical assumptions on the signal-noise ratio ($\|\theta^*\|/\sigma\geq \eta$, where $\eta$ is a sufficiently large absolute constant specified in \cite{balakrishnan2017statistical}). In particular, in the high signal-to-noise ratio regime, $\|\theta^*\|^2\gg \sigma^2$ so our improvements are significant.

Tight characterization of the statistical error is traditionally considered challenging in the area of mixture models. Only recently, \cite{kwon2018global} provided a refined analysis of the mixture of regression problem (Example \ref{example mixture regression}), and proved that the achievable generalization error is indeed of the order $\big({\sigma^2d\log\frac{1}{\delta}}/{n}\big)$. However, the analysis in \cite{kwon2018global} is fairly involved and customized to the specifics of the mixture of regression setting, and it is not clear how to extend the analysis to more general problems. Our theory can be applied to quite general settings, and moreover simplifies existing approaches.

From our theory, the optimal $O\big({\sigma^2d\log\frac{1}{\delta}}/{n}\big)$ characterization is very natural. Theorem \ref{thm fo em} indicates the crucial fact that statistical error of the first-order EM algorithm  only relies on $\P[\|\nabla\ell(\theta^*;z)\|^2]$, a quantity that depends only on the optimal parameter $\theta^*$. 

We now use  Example \ref{example mixture of Gaussian} to illustrate the simplicity of our analysis. Define the function $g:\R\rightarrow \R^+$ as 
\begin{align*}
    g(u)=\frac{2e^{-\frac{\|2\theta^*-u\|^2}{2\sigma^2}}}{e^{-\frac{\|u\|^2}{2\sigma^2}}+e^{-\frac{\|2\theta^*-u\|^2}{2\sigma^2}}},
\end{align*}
where $u$ is a random vector drawn from $N(0, \sigma^2 I_{d\times d})$. In the high SNR regime, $g(u)$ is anticipated to be very close to zero with high probability, due to the fact that
\begin{align*}
    \frac{\|u\|^2}{2\sigma^2}\gg \frac{\|2\theta^*-u\|^2}{2\sigma^2}.
\end{align*} In the Gaussian mixture model, whether $w=1$ or $w=-1$, it is straightforward to show that when conditioned on $w$, the random vector $(\nabla \ell(\theta^*;z)|w)$ has the same distribution as $u(1-g(u))+\theta^*g(u)$. As a result, we have
\begin{align*}
    \P[\|\nabla \ell(\theta^*;z)\|^2]=\E_{u}[\|u\cdot(1-g(u))+\theta^*\cdot g(u)\|^2].
\end{align*}
As $g(u)$ is very close to $0$ with high probability, $ \P[\|\nabla \ell(\theta^*;z)\|^2]$ should only scale with $\E_{u}[\|u\|^2]=\sigma^2 d$ rather than $\|\theta^*\|$. This intuition also applies to other examples like the mixture of linear regression model.

\section{Fast rates in supervised learning with structured convex cost}\label{sec supervised learning strongly convex}

The main purpose of this section is to recover the problem-dependent rates in \cite{mendelson2018learning, mendelson2014learning} for (possibly non-parametric and heavy-tailed) supervised learning problems with structured convex cost functions. While  \cite{mendelson2018learning, mendelson2014learning} propose an approach they call ``learning without concentration," our approach emphasizes the use of surrogate functions that are not ``sub-root," {and relates one-sided uniform inequalities to two-sided concentration of  ``truncated" functions. Besides providing a unification, there are some technical improvements as well. For example,  our approach does not require the ``star-hull" of the hypothesis class that may increase complexity, and there are concrete examples showing that the improvement may be meaningful for non-convex  classes. See  Section \ref{subsec comparison sup} for contributions of our method, and detailed comparison with existing approaches.}

\subsection{Background}

\paragraph{Problem formulation and assumptions.}
 Let the data $z$ be a feature-label pair $(x,y)$ where  $x\in\pazocal{X}$ and $y\in\pazocal{Y}\subseteq\R$. We assume every hypothesis $h$ in the hypothesis class $\Hy$ is a mapping from $\pazocal{X}$ to $\R$.  In supervised learning, the loss function is of the form $\ell(h;(x,y))=\ell_{\sv}(h(x),y)$ where the deterministic bivariate function $\ell_{\sv}:\R\times\R\rightarrow \R$ is called the {\it cost function}. 
  We assume that the cost function is differentiable, globally convex with respect to its first argument, and the population risk is smooth.
 \begin{assumption}[{\bf  differentiability,  convexity and smoothness}]\label{asm convex smooth}
The partial derivative of $\ell_{\textup{su}}$ with respect to its first argument, denoted  $\partial_1\ell_{\textup{su}}$, exists and is continuous everywhere, and
 $\ell_{\sv}$ is a convex function with respect to its first argument, i.e., $\forall u_1,u_2,y\in\R$,
 \begin{align*}
      \ell_{\textup{su}}(u_1,y)-\ell_{\textup{su}}(u_2,y)-\partial_1\ell_{\textup{sv}}(u_2,y)(u_1-u_2)\geq 0.
 \end{align*}
 {In addition, the population risk  is  smooth, i.e., there exists a constant $\beta_{\sv}>0$ such that $\forall h_1, h_2\in\Hy$, 
\begin{align*}
    \P\ell_{\textup{sv}}(h_1(x),y)-\P\ell_{\textup{sv}}(h_2(x),y)-\P[\partial_1\ell_{\textup{sv}}[(h_2(x),y)(h_1(x)-h_2(x))]\leq\frac{\beta_{\sv}}{2} \P[(h_1(x)-h_2(x))^2].
\end{align*}}
 \end{assumption}
\noindent Given a cost function that is globally convex and locally  strongly convex, we define  $\{\alpha(v)\}_{v\geq 0}$ as follows.
\begin{definition}[{\bf strong convexity parameter}]\label{def sup strongly convex}
 For a fixed $v>0$, let $\alpha(v)$ be the largest constant such that for all $y\in \pazocal{Y}$, $\ell_{\sv}(u+y,y)$ is $\alpha(v)-$strongly convex  with respect to $u$ when $u\in [-v,v]$. That is,
\begin{align*}
\ell_{\sv}(u_1+y,y)-\ell_{\sv}(u_2+y,y)-\partial_1\ell_{\sv}(u_2+y,y)(u_1-u_2)\geq \frac{\alpha(v)}{2}(u_1-u_2)^2, \quad \forall u_1, u_2\in [-v,v], \forall y\in\pazocal{Y}.
\end{align*}
Clearly $\{\alpha(v)\}_{v\geq 0}$ is non-increasing with respect to $v$, and we denote $\alpha(0)=\limsup_{v\rightarrow0}\alpha(v)$.
\end{definition}
\noindent When $\ell_{\sv}$ is second-order continuously differentiable, we have the simple relation
\begin{align*}
    \alpha(v)= \sup_{|u|\leq v, y\in \pazocal{Y}} \partial^2_{1,1}\ell_{\sv}(u+y,y), \quad, \forall v\geq 0,
\end{align*}
where  $\partial^2_{1,1}\ell_{\sv}$ is the second order partial derivative of $\ell_{\sv}$ with respect to its first argument. Moreover, to accommodate popular choices of robust costs,  Definition \ref{def sup strongly convex} also allows $\partial_1\ell_{\sv}$ to be non-differentiable at certain points in its domain. We list  three widely used  cost functions, their strong convexity parameters $\{\alpha(v)\}_{v\geq0}$, and the smoothness parameters $\beta_{\sv}$ of the corresponding population risks.
\begin{itemize}
\item Square cost: {consider the regression setting $
    \E[y|x]=h_{\text{true}}(x)$,
where $h_{\text{true}}$ is the function we want to estimate (not necessarily in $\Hy$). It is natural to consider the square cost function}
\begin{align*}\ell_{\sv}(h(x),y)=\frac{1}{2}(h(x)-y)^2.\end{align*} Here $\ell_{\sv}(u+y,y)=u^2$. Thus $\alpha(v)=\frac{1}{2}, \forall v\geq0$. {The smoothness parameter of the population risk is $\beta_{\sv}=\frac{1}{2}$.} In this example, one does not need to localize the  strong convexity parameter $\alpha(v)$ as it is a constant.

\item Huber cost: {consider the regression setting $
    \E[y|x]=h_{\text{true}}(x)$,
where $h_{\text{true}}$ is the function we want to estimate (not necessarily in $\Hy$). When the conditional distribution of $y$ is ``heavy tailed," one often considers the Huber cost function as follows.} For $\gamma>0$, let
\begin{align} \ell_{\sv,\gamma}(h(x),y)=\left\{
\begin{aligned}\label{eq: definition huber}
&\frac{1}{2}(h(x)-y)^2 & \text{ for } |h(x)-y|\leq \gamma,\\
&\gamma |h(x)-y|-\frac{\gamma^2}{2} & \text{ for } |h(x)-y|> \gamma.
\end{aligned}
\right.
\end{align}
Here $\alpha(v)=\frac{1}{2}$ whenever $v\leq \gamma$ but $\alpha(v)=0$ for all $v> \gamma $. {The smoothness parameter of the population risk is $\beta_{\sv}=\frac{1}{2}$. Localization analysis of  $\alpha(v)$ is required for this loss, and the key is to avoid its inverse diverging to infinity.}

\item Logistic cost:  {consider the standard logistic regression setting, where $y\in\{-1,1\}$ and one models the ``log odd ratio" as \begin{align}\label{eq: logistic regression}
    \log\left(\text{Prob}(y=1|x)/\text{Prob}(y=-1|x)\right)=h_{\text{true}}(x).
\end{align} Here $h_{\text{true}}$ is the discriminant function to be estimated (not necessarily in $\Hy$). The maximum likelihood estimation problem corresponds to using the cost function} \begin{align*}
\ell_{\sv}(h(x),y)=\log\Big(1+\exp(-yh(x))\Big).
\end{align*} Here $\partial_{1,1}^2\ell_{\sv}(u+y,y)=\frac{\exp(1+uy)}{(1+\exp(1+uy))^2}$, so we have $\alpha(v)=\frac{\exp(v+1)}{(\exp(v+1)+1)^2}$, $\forall v\geq 0$, {and the smoothness parameter of the population risk is $\beta_{\sv}=\frac{1}{4}$.}  The issue is that $\frac{1}{\alpha(v)}$, a complexity constant that will appear in the generalization error bound, grows exponentially with $v$  \cite{hazan2014logistic, marteau2019beyond}. {This issue strongly motivate us to  localize the parameter $v$ within $\alpha(v)$ to avoid large exponential constants.}

\end{itemize}

 The following assumption is usually invoked in the most representative literature on this topic  \cite{mendelson2014learning, mendelson2018learning}.
\begin{assumption}[{\bf optimality condition}]\label{asm regularity supervised} Recall that $h^*\in\P\ell_{\sv}(h(x),y)$ is the population risk minimizer. Assume for all $h\in \Hy$, 
\begin{align*}
    \P[\partial_1\ell_{\sv}(h^*(x),y)(h(x)-h^*(x))]\geq 0. 
\end{align*}
\end{assumption}
\noindent We summarize the two primary settings where Assumption \ref{asm regularity supervised} holds true.
\begin{itemize}
 \item Well-specified models: for certain problems, as long as the model is well-specified, then  $\partial_1\ell_{\sv}(h^*(x),y)$ is independent of $x$ and $\E\partial_1\ell_{\sv}(h^*(x),y)=0$. Thus Assumption \ref{asm regularity supervised} will hold. {Examples include 1) the settings studied in \cite{mendelson2018learning} where $\ell_{\sv}$ is a univariate function of $(h(x)-y)$ and $\partial_1\ell_{\sv}(h^*(x),y)$ is odd with respect to $y$, such as applications that use the square cost or the Huber cost; and 2) generalized linear models where the conditional distribution of $y$ belongs to the exponential family, such as the the logistic regression problem \eqref{eq: logistic regression}.}

\item $\Hy$ is a convex class of functions: in this case, we verify Assumption \ref{asm regularity supervised} as follows. If there exists some $h_1\in \Hy$ such that Assumption \ref{asm regularity supervised} is not true, then by considering $h_{\lambda}=\lambda h_1+(1-\lambda)h^*\in\Hy$ with $\lambda$ sufficiently close to 
$0$, we find $\P\ell_{\sv}(h_{\lambda}(x),y)<\P\ell_{\sv}(h^*(x),y)$, in contradiction, as $h^*$ is the population risk minimizer.
\end{itemize}

{
We call the random variable $\partial_1\ell_{\sv}(h^*(x),y)$ the ``noise multiplier" as it often characterizes the ``effective noise" of the learning problem when using a particular cost function.
We define another random variable
$
    \xi:=h^*(x)-y.
$
In some applications,  $\xi$ is closely related to the ``noise multiplier" (e.g., they are equivalent when one uses the square cost). And the notation $\xi$ is useful in other applications as well, because one always seeks to localize the parameter $v$ in $\alpha(v)$ to the order of $\|\xi\|_{L_2}$.
We denote $\Delta=\sup_{h\in\Hy}\|h(x)-y\|_{L_2}$ and ${\Delta_{\infty}}=\sup_{h,x,y}|h(x)-y|$ as the worst-case $L_2$ distance and $L_{\infty}$ distance between $h(x)$ and $y$. respectively. It is clear that we typically have $\|\xi\|_{L_2}\ll\Delta\ll\Delta_{\infty}$ in  practical applications.

Our analysis requires a very weak distributional assumption:  
\begin{assumption}[{\bf ``small ball" property}]\label{asm moment condition}
There exist constants $\kappa>0$ and $c_\kappa\in(0,1)$ such that for all $h\in\Hy$,
\begin{align*}
  \textup{Prob}\left(|h(x)-h^*(x)|\geq \kappa \|h-h^*\|_{L_2}\right)\geq c_\kappa .
\end{align*}
\end{assumption}
\noindent Assumption \ref{asm moment condition} is often referred to as ``minimal" in the literature, and there are many examples in which it can be verified for $\kappa$ and $c_\kappa$ that are absolute constants \cite{mendelson2014learning, mendelson2018learning, lecue2014sparse, koltchinskii2015bounding,   rudelson2015small, lecue2018regularization}.
The scope of Assumption \ref{asm moment condition} subsumes and is much broader than the ``sub-Gaussian" setting. For example,  it is naturally satisfied when the class $\{h-h^*:h\in\Hy\}$ satisfies any sort of moment equivalence (see, e.g., [\citealp{mendelson2014learning}, Lemma 4.1]).
}

\paragraph{Main challenges.} 
Let us first examine limitations of the results obtained using the traditional ``local Rademacher complexity" analysis (Statement \ref{state current blueprint}), which includes the results from  \cite{bartlett2005local, wainwright2019high, foster2019orthogonal} in the fast-rate regime.  Assuming the cost function to be $L_{\sv}-$Lipchitz continuous with respect to its first argument and setting  $f(z)=\ell_{\sv}(h(x),y)-\ell_{\sv}(h^*(x),y)$, $T(f)=\P[f^2]$, and $B_e={L_{\sv}^2}/{\alpha(\Delta_{\infty}})$, following  Statement \ref{state current blueprint}, one can prove that the empirical risk minimizer $\hh$ satisfies 
 \begin{align}\label{eq: sup traditional}
     \mathcal{E}(\hh)\leq O\left(\frac{r^*}{B_e}\right),
 \end{align}
where  $r^*$ is the fixed point of  $B_e\psi$, and $\psi$ is a sub-root surrogate function that governs $\sup_{\P[f^2]\leq r}{(\P-\Pn)f}$. Denote by $r_{1}^*$  the fixed point of $\psi$. From the sub-root property of $\psi$ we know that $r^*\geq B_e^2 r_1^*$, so the generalization error bound \eqref{eq: sup traditional} is at least of order
\begin{align}\label{eq: at least order traditional sup}
  \frac{r^*}{B_e}\geq B_e r_1^*= \frac{L_{\sv}^2}{\alpha(\Delta_{\infty})} r_1^*.
\end{align}The main message here is that the traditional result \eqref{eq: sup traditional} is often loose  and not problem-dependent. As indicated by Mendelson in a series of papers \cite{mendelson2014learning, mendelson2018learning}, the traditional result \eqref{eq: sup traditional} has the following limitations.
\begin{itemize}
    \item   The global Lipchitz constant $L_{\sv}$ is not problem-dependent and potentially unbounded.  {$L_{\sv}$ is effectively the worst-case value  $\sup_{h,x,y}|\partial_1\ell(h(x),y)|$. For the square cost, this is  $\Delta_{\infty}=\sup_{h,x,y}|h(x)-y|$ and is unbounded when either the hypothesis class or  noise are unbounded.
     It would be beneficial to have a bound that only scales  with  a measure related to the ``noise multiplier" $\partial_1\ell(h^*(x),y)$, because we usually have  $|\partial_1\ell(h^*(x),y)|\ll L_{\sv}$ in practical applications.}
    
    \item The global strong convexity parameter $\alpha(\Delta_{\infty})$ is often very small for the logistic cost and the Huber cost, so its inverse is often large (and potentially unbounded).  The challenge here is to  sharpen this to the inverse of a localized strongly convex parameter $\alpha(O(\|\xi\|_{L_2}))$. Since we usually have $\sigma\ll\Delta_{\infty}$, the inverse of
    $\alpha(O(\|\xi\|_{L_2}))$ can be much smaller than the inverse of $\alpha(\Delta_{\infty})$.
\end{itemize}
  \paragraph{The ``small ball method" and beyond.}  {The breakthrough papers \cite{mendelson2014learning, mendelson2018learning}  propose the ``small ball method" (also referred to as ``learning without concentration") to provide problem-dependent rates that overcome the limitations mentioned above. Their proofs builds on structural results of $0-1$ valued indicator functions under the small-ball condition, whose connection to the traditional localization analysis may not be completely obvious. {Moving the focal point from indicator functions to  ``truncated" functions, we  provide the following perspectives.
  
    1) A simple interpretation to the ``small-ball" condition is that, suitably ``truncated"  quadratic forms are of the same scale as the original quadratic forms. Under the ``small-ball" condition,  one can trivially show that uniformly over all $h\in\Hy$,
    \begin{align*}
        \P[\min\{(h(x)-h^*(x))^2, \kappa^2\|h-h^*\|_{L_2}^2\}]\geq  \text{Prob}\left(|h(x)-h^*(x)|\geq \kappa\|h-h^*\|_{L_2}\right)\kappa^2\|h-h^*\|_{L_2}^2\\
        \geq c_\kappa \kappa^2 \P[(h(x)-h^*(x))^2].
    \end{align*}
    This suggests that one only needs to concentrate simple ``truncated" functions to derive  generalization error bounds.
    
   2) One-sided uniform inequalities are contained in the ``uniform localized convergence" framework and are often  derived from  concentration of truncated functions. Many one-sided uniform inequalities can be equivalently written as ``uniform localized convergence" arguments. Consider the uniform ``lower isomorphic bound" (which plays a  central role in the ``small-ball" method):   for some constant $c>0$, with high probability,  uniformly over all $h\in\Hy$, 
    \begin{align*}
        \Pn[(h(x)-h^*(x))^2]\geq c\P[(h(x)-h^*(x)^2].
    \end{align*}
    The above argument is equivalent with the following ``uniform localized convergence" argument:
    \begin{align*}
        (\P-\Pn)[(h(x)-h^*(x))^2]\leq (1-c)T(h),\quad \forall h\in\Hy
    \end{align*}
    where the measurement functional $T(h)$ is set to be $\|h-h^*\|^2_{L_2}$.
    A more flexible perspective may directly view the truncated quadratic forms as the concentrated functions, making traditional two-sided uniform convergence tools  applicable in a straightforward manner.

     Motivated by the above observations, an interesting question is to  recover} the results in \cite{mendelson2014learning, mendelson2018learning} through a ``one-shot" concentration framework,  explicitly figuring out which component of the excess loss contributes to which part of the surrogate function. In what follows, we will present such an analysis. While our  error bounds roughly follow the same form as the results in \cite{mendelson2014learning, mendelson2018learning}, we obtain several technical improvements; see Section \ref{subsec comparison sup} for the novel implications and methodological contributions of our approach.
   }

\subsection{Main results  and illustrative examples}\label{subsec main results sup}

{We assume some regularity conditions that  hold for non-pathological choices of surrogate functions.}

\begin{assumption}[{\bf regularity conditions on surrogate functions}]\label{asm regularity surrogate sup}
Assume there is a non-decreasing, non-negative and bounded function $\varphi(r)$ such that $\forall r>0$, \begin{align}\label{eq: surrogate ver}
    \mathfrak{R}\{h-h^*: h\in\Hy, \|h-h^*\|_{L_2}^2\leq r\} \leq \varphi(r); 
\end{align}
and there is a meaningful surrogate function
$\varphi_{\textup{noise}}(r,\delta)$ that is non-decreasing w.r.t. $r$, and satisfies that  $\forall \delta\in(0,1)$, with probability at least $1-\delta$,
\begin{align}\label{eq: thm3 surrogate}
    \sup_{h\in\Hy,\|h-h^*\|_{L_2}^2\leq r}\left\{ (\P-\Pn)[\partial_1\ell_{\textup{sv}}(h^*(x),y)(h-h^*)]\right\} \leq \varphi_{\textup{noise}}(r,\delta).
\end{align}
Given any fixed $\delta\in(0,1)$ and $r_0\in(0,4\Delta^2)$, denote $C_{r_0}= 2+\left(\frac{16}{c_\kappa}+2\right)\log\frac{4\Delta^2}{r_0}$. Assume there is a positive integer $\bar{N}_{\delta, r_0}$ such that for all $n\geq \bar{N}_{\delta, r_0}$, 
\begin{align}\label{eq: weak sample size sup}
    \varphi_{\textup{noise}}\left(8\Delta^2; \frac{\delta}{C_{r_0}}\right)\leq \frac{\alpha(4\|\xi\|_{L_2}/\sqrt{c_\kappa})\|\xi\|_{L_2}^2}{2} \quad  \text{and}\quad
       \varphi\left(8\Delta^2\right)\leq \frac{\sqrt{2c_\kappa}\|\xi\|^2_{L_2}}{16\Delta}.
\end{align}
\end{assumption}
We note that the requirements do not place meaningful  restrictions on the choice of surrogate function. The main requirement, condition \eqref{eq: weak sample size sup}, {asks for uniform errors} over $\Hy$ to be smaller than some fixed values that are independent of $n$. For non-pathological choices of surrogate functions, this will always be satisfied as long as the sample size $n$ is larger than some positive integer $\bar{N}_{\delta, r_0}$. The boundedness requirement for $\varphi$ (and $\varphi_{\text{noise}}$) can always be met by setting $\varphi(r)=\varphi(4\Delta^2)$ (and $\varphi_{\text{noise}}(r;\delta)=\varphi_{\text{noise}}(4\Delta^2;\delta)$) for all $r\geq 4\Delta^2$, because $\|h-h^*\|_{L_2}\leq 2\Delta$ for all $h\in\Hy$.

\begin{theorem}[{\bf supervised learning with structured convex cost}]\label{thm sup} Let Assumptions \ref{asm convex smooth} \ref{asm regularity supervised}, \ref{asm moment condition}, \ref{asm regularity surrogate sup} hold and $\alpha\left(4\|\xi\|_{L_2}/\sqrt{c_\kappa}\right)>0$. Let $r_{\text{ver}}^*$ be the fixed point of the function
\begin{align}\label{eq: thm 3 ver}
 \frac{4}{c_\kappa \kappa^2\cdot \alpha\left(4\|\xi\|_{L_2}/\sqrt{c_\kappa}\right)}\varphi_{\text{noise}}\left(2r;\frac{\delta}{C_{r_0}}\right).
\end{align}
Given any fixed $\delta\in(0,1)$ and $r_0\in(0,4\Delta^2)$, let $r_{\text{noise}}^*$ be the fixed point of the function
\begin{align}\label{eq: thm 3 noise}
     \frac{8}{c_\kappa\kappa}\sqrt{2r}\varphi(2r).
\end{align}
Then with probability at least $1-\delta$, the empirical risk minimizer $\hh$ satisfies
\begin{align*}
    \|\hh-h^*\|^2_{L_2(\P)}\leq \max\left\{ r^*_{\textup{noise}},\ r^*_{\textup{ver}},\ r_0\right\} \quad \text{and} \\
   \quad \mathcal{E}(\hh)\leq\frac{\beta_{\sv}}{2}\max\left\{ r^*_{\textup{noise}},\ r^*_{\textup{ver}},\ r_0\right\},
\end{align*}
provided that $ n>\max\left\{\bar{N}_{\delta,r_0},\frac{72}{c_\kappa^2}\log\frac{C_{r_0}}{\delta}\right\}$.
\end{theorem}

 \paragraph{Remarks.}
 1) The term $r_0$ is negligible since it can be arbitrarily small. One can simply set $r_0={1}/{n^4}$, which will be much smaller than $r^*_{\text{noise}}$ for typical applications. 
 In high-probability bounds, $C_{r_0}$ will only appear in the form $\log ({C_{r_0}}/{\delta}))$, which is of a negligible $O(\log\log n)$ order. {In the subsequent discussion, we will hide parameters that only depend on $\kappa$ and $c_\kappa$ in the big $O$ notation, as they are often absolute constants in practical applications. }
 
 \noindent 2) {The two fixed points $r^*_{\text{noise}}$ and $r^*_{\text{ver}}$ correspond to the two sources of complexities: the uniform errors characterized by the two surrogate functions in \eqref{eq: thm 3 ver} and \eqref{eq: thm 3 noise}. 
 Recall that a fundamental limitation of the traditional ``local Rademacher complexity analysis" is that it requires a ``sub-root" surrogate function that can not differentiate  the two sources of complexity. In contrast, the surrogate function in \eqref{eq: thm 3 ver} (which we write as $O(\sqrt{r}\varphi(r))$ for simplicity) is obviously a  ``super-root" function, thus our analysis overcomes that limitation and provides more precise upper bounds. The key point is that $O(\sqrt{r}\varphi(r))$ is a  benign ``super-root" surrogate function, in the sense that its fixed point $r^*_{\text{ver}}$ is ``very small" when the sample size is large enough;  in other words, when the problem is learnable. For example, for a $d-$dimensional linear classes, where $\varphi=O(\sqrt{{dr}/{n}})$,  $r^*_{\text{ver}}$ will be the fixed point of $O(dr/n)$. Thus $r^*_{\text{ver}}$ will be $0$ as long as the sample size $n$ is larger than $O(d)$. Therefore, the typical generalization error derived by Theorem \ref{thm sup} is of order
 \begin{align*}
     \mathcal{E}(\hh)\leq \frac{\beta_{\sv}}{2} r^*_{\text{noise}}, 
 \end{align*}
 where $r^*_{\text{noise}}$ is the fixed point of the function in \eqref{eq: thm 3 noise}. Clearly, $r^*_{\textup{noise}}$ only depends on the noise multiplier at $h^*$ and the local strong convexity parameter, and it is independent of the worst-case parameters of the cost function.}

 At a high level, the subscripts ``ver" and ``noise" have the the meaning of ``version space" and ``noise multiplier," respectively. Intuitively, $r_{\text{ver}}$ is the estimation error of the noise-free realizable problem, which reflects the complexity of version space---the random subset of $\Hy$ that consists of all $h$ that agree with $h^*$ on $\{x_i\}_{i=1}^n$.  On the other hand, $r_{\textup{noise}}$ is the estimation error induced by the interaction of $\Hy$ and noise multiplier $\partial_1\ell_{\textup{sv}}(h^*(x),y)$. We refer to \cite{mendelson2014learning} for a more detailed discussion on the source of these two fixed points.

{
Now we present some representative applications of Theorem \ref{thm sup}.

\begin{example}[{\bf localization of unfavorable  parameters}]
In practical applications, one often wants to avoid the global Lipchitz constant and the inverse of the global strong convexity parameter. For example, in  regression with square cost, the global Lipchitz constant is equal to $\Delta_{\infty}$ and is often unbounded, so it is desirable  to convert it to $\|\xi\|_{L_2}$; and in logistic regression, the inverse of global strong convexity parameter is an exponential constant $e^{O(\Delta_{\infty})}$,  which we hope to  convert  to  $e^{O(\|\xi\|_{L_2})}$. 
These goals are achieved in Theorem \ref{thm sup}:
since the right hand side of \eqref{eq: thm 3 ver} contains an extra $\sqrt{r}$ factor,  $r^*_{\text{ver}}$ is typically  much smaller than $r^*_{\textup{noise}}$ for sufficiently large $n$ (see remark 2 after Theorem \ref{thm sup}). Therefore, the generalization error bound will be determined by the fixed point $r^*_{\textup{noise}}$, which only depends on the noise multiplier at $h^*$ and the local strong convexity parameter.

\end{example}
}

\begin{example}[{\bf  regression with heavy-tailed noise}]
 { We consider the problem of predicting $y$ using $h(x)$, and allow the ``noise" $\xi=h^*(x)-y$ to be heavy-tailed.} To illustrate the main message of this example, we consider the $d-$dimensional linear class with sub-Gaussian features. That is,  $h(x)=\theta^Tx$ where $\theta\in\R^d$, and the random feature $x\in\R^d$ is sub-Gaussian. In this setting, the Huber cost is preferred to the square cost.
\begin{itemize}

\item For the Huber cost and truncation parameter  $\gamma=O(\|\xi\|_{L_2})$ in the definition \eqref{eq: definition huber}, Theorem \ref{thm sup} implies that the parameter $v$ will be localized to the region where the strong convexity parameter  $\alpha(v)$ is non-zero. As a result, the strong convexity parameter in the generalization error bound will be $\frac{1}{2}$ rather than the problematic value $0$ (since the generalization error scales with the inverse of $\alpha(v)$, the value $0$ will make the bound vacuous).  For the $d-$dimensional linear class,  $r^*_{\text{ver}}$ will be $0$ as long as $n\geq O(d)$. Since $\partial_1\ell_{\sv}(h^*(x),y)$ will be uniformly bounded by $O(\sigma)$, we obtain 
\begin{align*}
    r_{\textup{noise}}^*\leq O\left (\frac{\|\xi\|_{L_2}^2 (d+\log\frac{1}{\delta})}{n}\right),
\end{align*}
which recovers the problem-dependent rate in \cite{mendelson2018learning}.

\item For the square cost, the fixed point $r_{\textup{noise}}$ will often cause the generalization error to be sub-optimal. For the $d-$dimensional linear class, $r_{\textup{noise}}$ will have a {polynomial dependence on $1/\delta$} as explained in \cite{mendelson2018learning}.  The reason is that in the definition of $\varphi_{\textup{noise}}(r,\delta)$ in \ref{eq: thm3 surrogate},  the noise multiplier $\partial_1\ell_{\sv}(h^*(x),y)$ is equal to $\xi$ for the square cost. For ``heavy-tailed" $\xi$, this will cause the rate $r_{\textup{noise}}$ to be sub-optimal.
\end{itemize}
\end{example}

We note that the condition that $\hh$ is the empirical risk minimizer is not essential to the proof of Theorem \ref{thm sup}. Similar to the prior work \cite{lecue2018regularization}, we can extend the result to more general learning rules that are based on regularization (e.g., LASSO \cite{tibshirani1996regression}, SLOPE \cite{bogdan2015slope}, etc.) as follows.

\begin{corollary}[{\bf extension to general regularized learning rules}]\label{coro sup}
Let Assumptions \ref{asm convex smooth} \ref{asm regularity supervised}, \ref{asm moment condition} hold. Let $\hh$ be the solution of
\begin{align}\label{eq: definition regularized risk minimizer}
    \min_{\Hy} \Pn \ell_{\sv}(h(x),y)+ \Psi (h),
\end{align}
where $\Psi(h)$ is a non-negative regularization term. Let ${\Hy_0}$ be a subset of $\Hy$ that is independent of the samples. If  inequality \eqref{eq: surrogate ver} is modified to 
\begin{align*}
     \mathfrak{R}\{h-h^*: h\in\Hy_0, \|h-h^*\|_{L_2}^2\leq r\} \leq \varphi(r),
\end{align*}
and  inequality  \eqref{eq: thm3 surrogate} is modified to
\begin{align*}
    \sup_{h\in\Hy_0,\|h-h^*\|_{L_2}^2\leq r}\bigg\{ (\P-\Pn)[\partial_1\ell_{\textup{su}}(h^*(x),y)(h-h^*)]\bigg\} +\Psi (h^*) \leq {\varphi}_{\textup{noise}}(r;\delta),
\end{align*}
then under Assumption \ref{asm regularity surrogate sup}, conditioned on the event $\{\hh\in\Hy_0\}$, the conclusion of Theorem \ref{thm sup} remains true.
\end{corollary}

{As illustrated in the following example, Corollary \ref{coro sup} is able to recover several  important results in the high-dimensional statistics literature.}

\begin{example}[{\bf high-dimensional estimation and LASSO}]\label{eq: example high dimensional}
Consider the linear regression set-up $\E[y|x]=x^T\theta^*$ where $\theta\in\Theta\subseteq \R^d$, $d\gg n$ and $\|\theta^*\|_0\leq s\ll d$.  Consider the LASSO estimator $\hth$, which is the solution of the $\ell_1-$norm regularized risk minimization problem, where the regularization term is $\Phi(h)=\lambda\|\theta\|_1$ and $\lambda>0$ is the regularization parameter, i.e.,
\begin{align*}
 \hth \in \argmin_{\Theta} \Pn \ell_{\sv}(\theta^Tx, y)+\lambda\|\theta\|_1.
\end{align*}
 Assume $\ell_{\sv}$ is the square cost and $\xi$ is $\sigma-$sub-Gaussian, or $\ell_{\sv}$ is the Huber cost with truncation parameter $\gamma=O(\sigma)$. Assume the feature $x\in\R^d$ is sub-Gaussian. Following standard analysis (see, e.g., [\citealp{negahban2012unified}, Lemma 1]),  by setting $\lambda$ to be of order $\sqrt{\sigma^2\log(d/\delta)/n}$, the Lasso estimator $\hth$ will lie in a sparse cone $\Theta_S$ (with high probability), where 
it can be proven \cite{loh2013regularized} that $\varphi(r)=O(\sqrt{{ rs\log d}/{n}})$
      and $\varphi_{\text{noise}}(r;\delta)=O(\sqrt{{r\sigma^2  s \log ({d}/{\delta})}/{n}})$  (ignoring dependence on the parameters $C$ and $p$ described in Assumption \ref{asm moment condition}).
 Applying Corollary \ref{coro sup} with $\Hy_0=\{x\mapsto\theta^Tx: \theta\in\Theta_S\}$ and $n\geq \Omega(s\log d)$, we have $r^*_{\text{ver}}=0$ and
\begin{align*}
    r^*_{\textup{noise}}\leq O\left(\frac{\sigma^2 s\log\frac{d}{\delta} }{n}\right).
\end{align*}
\end{example}

\subsection{Contributions relative to previous approaches}\label{subsec comparison sup}
 So far we have recovered the main  results in the prior works \cite{mendelson2014learning, mendelson2018learning}, which are valid for unbounded regression problems and thus improve the traditional ``local Rademacher complexity" analysis. Now we would like to illustrate how Theorem \ref{thm sup} improves the results in \cite{mendelson2014learning, mendelson2018learning} by removing a ``star-shape" requirement. That is, we do not need to assume the hypothesis class is star-shaped/convex, or consider the star-hull of it which may increase  complexity. 

To be specific, \cite{mendelson2014learning, mendelson2018learning} assumes that $\Hy$ is a convex class (and thus star-shaped). When $\Hy$ is not star-shaped, the  results in \cite{mendelson2014learning, mendelson2018learning} are still valid by taking the star-hull of $\F$ and considering the local Rademacher/Gaussian complexity of the star-hull. The increase in complexity is quite moderate for traditional hypothesis classes (e.g., those chracterized by covering number conditions [\citealp{mendelson2002improving}, Lemma 4.6]). However, taking the star-hull may significantly increase the local Rademacher complexity of modern non-convex and overparameterized classes. Here we show that, even for very simple function classes (e.g., linear classes with  non-convex support), our approach improves on what can be achieved using the star-hulls. 

Note that the improvement brought by our approach is systematic and may carry over to more complicated learning procedures as well.  A more comprehensive comparison with existing localization approaches will be presented after the following example.

\begin{example}[\bf overparameterized linear class with growing sparsity]\label{example improvement star}
 Consider the linear regression model 
 \begin{align*}
     y\sim N(x^T\theta^*, \sigma^2), \quad x\sim N(0, I_{d\times d}),
 \end{align*} where $\theta^*\in\Theta\subseteq\R^d$ and $d\gg n$ (i.e., the model is overoarameterized). Assume the feasible parameter set $\theta$ satisfies that for all $\theta\in\Theta$,
 \begin{align}\label{eq: sparsity condition}
     \|\theta-\theta^*\|_{0}\leq \lfloor \|\theta-\theta^*\|_{2}^2\rfloor.
 \end{align}In other words, the sparsity of $\theta$ increases the more $\theta$ deviates from $\theta^*$. The maximum likelihood estimation problem corresponds to minimize the empirical average of the square cost with respect to  $\Hy=\{x\mapsto x^T\theta: \theta\in\Theta\}$. 
For this problem, the surrogate function $\varphi_{\text{noise}}$ need to satisfy (with probability at $1-\delta$)
\begin{align}\label{eq: requirement overparameterized example}
    \sup_{\theta\in\Theta, \|\theta-\theta^*\|_{2}^2\leq r }(\P-\Pn)[\xi\cdot x^T(\theta-\theta^*)]\leq \varphi_{\text{noise}}(r;\delta),
\end{align}
 where the left hand side of \eqref{eq: requirement overparameterized example} is the localized Gaussian complexity of $\Hy$. Thanks to the sparsity condition \eqref{eq: sparsity condition}, it can be tightly controlled by
\begin{align}\label{eq: surrogate overparameterized}
    \varphi_{\text{noise}}(r;\delta)=O\left(\sqrt{\frac{\sigma^2(\|\theta^*\|_{0}+r)r\log\frac{d}{\delta}}{n}}\right)=\underbrace{O\left(\sqrt{\frac{\sigma^2\|\theta^*\|_{0}r\log\frac{d}{\delta}}{n}}\right)}_{\text{problem-dependent component}}+\underbrace{O\left(\sqrt{\frac{\sigma^2\log\frac{d}{\delta}}{n}}\cdot r \right)}_{\text{benign ``super-root" component}}.
\end{align}
Here, the benign ``super-root" component" in $\varphi_{\text{noise}}(r;\delta)$ does not affect the order of its fixed point $r_{\text{noise}}^*$: when $n\geq \Omega(4\sigma^2\log\frac{d}{\delta})$, the ``super-root" component" in \eqref{eq: surrogate overparameterized} will be less than $\frac{1}{2}r$ so that  $r_{\text{noise}}^*$ is of  order $\sigma^2\|\theta^*\|_0\log\frac{d}{\delta}/n$. In other words, only the problem-dependent component in $\varphi_{\text{noise}}(r;\delta)$ matters.

In contrast, if one takes the star-hull (e.g., expanding $\Theta$ to $\text{star}(\Theta)=\{\theta^*+\lambda(\theta-\theta^*): \theta\in\Theta, \lambda\in[0,1]\}$, then it is straightforward to verify that $\varphi_{\text{noise}}$ has to be a ``sub-root" function. A sub-root surrogate function that governs \eqref{eq: surrogate overparameterized} will be at least of order
\begin{align*}
   \overline{\varphi_{\text{noise}}}(r;\delta)= O\left(\sqrt{\frac{\sigma^2(\|\theta^*\|_0+\Delta)r\log\frac{d}{\delta}}{n}}\right),
\end{align*}
whose fixed point unavoidably scales with the worst-case $L_2$ distance $\Delta$. Here we do not consider computational issues, and the key message is that if the complexity (e.g., the ``effective dimension") of an overparameterized non-convex class grows very rapidly with respect to its localization scale, then some ``fast growing components" may still be benign and they may not necessarily increase the complexity. It is an open question whether such phenomena manifests in more practical applications.
\end{example}

\paragraph{Comparison with the ``small ball method."}
In a series of pioneering works, Mendelson \cite{mendelson2014learning, mendelson2018learning, mendelson2017aggregation, mendelson2017optimal} proposes the ``small ball method"  as an alternative approach to the traditional ``concentration-contraction" framework. Under the  ``small ball" condition,  that approach establishes one-sided uniform inequalities through structural results on binary valued indicator functions. Motivated by these works, we seek to refine the traditional concentration framework. Our approach brings added flexibility to concentration by emphasizing the use of surrogate functions that are not ``sub-root," and  relates one-sided uniform inequalities to two-sided concentration of simple ``truncated" functions. Following are the main contributions relative to the ``small ball method."

First, our approach does not require the hypothesis class to be star-shaped/convex (or to consider the star-hull of the hypothesis class).  
This improvement is particularly relevant for non-convex hypothesis classes whose complexity can grow rapidly when ``away" from the optimal hypothesis. In Example \ref{example improvement star} (and its discussion) we show that the improvement may be meaningful for  some non-convex, overparametrized classes; and the phenomenon of ``benign fast growing" components in overparameterized models may be of independent interest.

To the best of our knowledge, the ``small ball method" cannot overcome the star-shape requirement in a straightforward manner,  without additional uniform convergence arguments. The ``small ball method" is able to prove one-sided inequalities that hold uniformly over a fixed sphere $\{h\in\Hy:\|h-h^*\|_{L_2}^2=r\}$,  and by assuming  the class $\Hy$ to be star-shaped around $h^*$, it circumvents the need to have a uniform bound that holds simultaneously for all possible values of $r$. However, without the star-shape assumption and additional uniform convergence arguments, it is not clear how to uniformly extend the bound to all $r$ using peeling.
In our analysis, we introduce some new tricks to address this issue. In particular, we use ``adaptive truncation levels" and concentration over ``rings." Combining these  with the ``uniform localized convergence" procedure, we completely circumvent the need for star-hulls (see ``Part \RNum{2}" in Appendix \ref{subsec sup} for details).

The discussion here is orthogonal to lifting the  star-shape/convexity assumptions   using  aggregation  \cite{liang2015learning}, whose primary goal is to remove Assumption \ref{asm regularity supervised} (recall that this assumption implicitly asks the hypothesis class to be convex/star-shaped when the model is mis-specified). When using aggregation and improper learning procedures, it is natural to consider the complexity of the enlarged class. Still, we suspect that taking the star-hull may be unnecessary if the enlarged class need not to be star-shaped \cite{mendelson2017aggregation, mendelson2017optimal}, and our analysis may be useful there as well. We note in passing that aggregation procedures are often  computationally demanding.

  Lastly, the formulation of supervised costs  is  slightly broader here compared with  \cite{mendelson2018learning}. In that paper, the loss is assumed to be a univariate function of $(h(x)-y)$, so  costs involving the term $yh(x)$ (e.g., the canonical logistic cost and the costs in some other generalized linear models) are not permitted (\cite{mendelson2018learning} instead analyzes a modified version of the logistic cost). 
 
 \paragraph{Comparison with offset Rademacher complexity.} Under the square cost and assuming the so-called ``lower isometry bound" as an a priori condition (see [\citealp{liang2015learning}, Definition 5]), offset Rademahcer complexity \cite{liang2015learning} is also able to  provide problem-dependent rates. However, establishing such a ``lower isometry bound" is typically challenging, so this approach may still need to rely on the ``small ball method" (or our analysis) for unbounded regression problems. Moreover, this tool is tailored to the setting of supervised learning with  square cost, and it is unclear how to extend the analysis to more general losses.

 \paragraph{Comparison with the ``restricted strong convexity" framework in high-dimensional statistics.}  
   In the  high-dimensional statistics literature, the ``restricted strong convexity" framework \cite{negahban2012unified, wainwright2019high} provides analytical tools to prove problem-dependent rates, but only when such condition is assumed as an a priori (see [\citealp{negahban2012unified}, Definition 2]). To achieve this,    \cite{raskutti2012minimax, negahban2012unified, loh2013regularized}  develop a  truncation-based analysis that can establish ``restricted strong convexity" for sparse kernel regression and sparse generalized linear models. Those works also indicate that one-sided uniform inequalities can be established by two-sided concentration of the ``truncated" functions. There are several differences between their analysis and ours. First, those proofs rely on linearity/star-shape of the hypothesis class and thus only need to prove the ``restricted strong convexity" on a fixed sphere  (similar to what we have discussed in comparison with the ``small-ball method"). In contrast, our framework does not put any geometric restriction on the hypothesis class, by passing this through the use of  ``adaptive truncation levels" and  concentration over ``rings," tools that may be of independent interest from a technical perspective. Second, when seeking problem-dependent generalization error bounds, the proposed $L_2-L_4$ moment equivalence condition \cite{negahban2012unified, loh2013regularized} is stronger than the ``small ball" condition used in our analysis. Third, the analysis does not fully localize the strong convexity parameter, and does not cover interesting supervised costs that may have zero curvature, e.g., the Huber cost.

\section{Concluding remarks}\label{sec concluding remarks}

This paper provides contributions both in the ``uniform localized convergence" approach it develops, as well as the applications thereof to various problems areas.  Below we highlight some key implications.

{
From a methodological viewpoint, our approach resolves some fundamental limitations of the existing
uniform convergence and localization analysis methods, such as the traditional ``local Rademacher complexity”
analysis and the ``uniform convergence of gradients.” 
At a high-level, it provides some general guidelines to derive generalization error bounds that are sharper than the worst-case uniform error. In particular, the following observations are of particular interest: 1) problem-dependent rates can often be explained by uniform inequalities whose right hand side is a function of the ``free" variable $T(f)$; 2) the choice of surrogate function and concentrated function are flexible, and our proposed framework brings some level of unification to localized complexities, vector-based uniform convergence results and one-sided uniform inequalities; and 3)  ``uniform localized convergence" arguments are also suitable to study regularization and  iterative algorithms. These observations lead to a unified perspective on problem-dependent rates in various problem settings studied in the paper. 
}

{ Many problem-dependent generalization error bounds proved in the paper may be of independent interest. Their study also informs the design of optimal procedures. }
For example:  in the ``slow rate" regime, we propose the first (moment-penalized) estimator that achieves optimal variance-dependent rates for general ``rich" classes; and in the parametric ``fast rate" regime, we show that efficient algorithms like gradient descent  and the first-order Expectation-Maximization algorithm can achieve optimal problem-dependent rates in several representative problems from non-convex learning, stochastic optimization, and learning with missing data. 

There are several future directions for this line of research.  Applications to causal inference and machine learning problems with unobservable components may be very promising, as the focus there is typically on  avoiding the worst-case parameter-dependence (see also discussion in Section \ref{sec application} and Section \ref{sec em}). 
 Another interesting topic is applying the ``uniform localized convergence" principle to study distributional robustness, since there are profound connections between the latter and variation-based regularization \cite{namkoong2017variance,  blanchet2019robust}.  
 {Lastly, extension of our framework to overparameterized models is interesting from both the theoretical and practical viewpoints.  Our results in the slow rate regime may be directly applicable to the study of overparameterized neural networks, in particular, if one has sharp upper bounds on the local Rademacher complexity. While there has been much recent interest in proving norm-based upper bounds on the {\it global} Rademacher complexity for neural network models \cite{bartlett2017spectrally, golowich2018size}, proving meaningful upper bounds on the {\it local} Rademahcer complexity remains largely open. It is worthy mentioning that the recent negative results in \cite{nagarajan2019uniform, bartlett2020failures} neither apply to our general framework nor the notion of problem-dependency we suggest (distribution-dependent quantities that depend on the best hypothesis). It is possible that combining our framework with more suitable concentrated functions and localized subsets (e.g., generalizing the data-dependent subset considered in \cite{zhou2020uniform})  may shed light also on the study of some overparameterized models.}

\bibliography{references}

\appendix

\section{Proofs for Section \ref{sec theoretical foundations} and Section \ref{sec slow rate regime}}\label{sec appendix proof}

 In all the proofs we consider a fixed sample size $n$. In order to distinguish ``probability of events" and ``expectation with respect to $\P$," we will use the notation $\text{Prob}(\mathcal{A})$ to denote the probability of the event $\mathcal{A}$.

\subsection{Variants of Proposition \ref{prop peeling}}\label{subsec appendix proposition}

We prove a more general version of of Proposition \ref{prop peeling}.  The differences are that 1) here we use a more general ``peeling scale" $\lambda$ which can be any value larger than $1$, while in Proposition \ref{prop peeling} we simply set $\lambda$ to be $2$; and 2)
we only ask $\psi(r;\delta)$ to be a high-probability surrogate function of the uniform error over the ``ring" $\{f\in\F: r/\lambda\leq T(f)\leq r\}$ rather than the ``bigger" localized area $\{f\in\F: 0\leq T(f)\leq r\}$. 
\begin{proposition}[\bf a more general ``uniform localized convergence" argument]\label{prop peeling stronger}
For a function class  $\G=\{g_f:f\in\F\}$ and   functional $T:\F\rightarrow [0,R]$, assume there is a function  $\psi(r;\delta)$ (possibly depending on the samples), which is non-decreasing with respect to $r$ and satisfies that $\forall \delta\in(0,1), \forall r\in[0,R]$, with probability at least $1-\delta$,
\begin{align*}
\sup_{f\in\F:\frac{r}{\lambda}\leq T(f)\leq r}(\P-\Pn)g_f\leq \psi(r;\delta).
\end{align*}
 Then, given any $\delta\in(0,1)$, $r_0\in(0,R]$ and $\lambda>1$, with probability at least $1-\delta$, for all $f\in \F$, either $T(f)\leq {r_0}$ or
\begin{align*}
    (\P-\Pn) g_f\leq  \psi\left(\lambda T(f);\delta\left({\log_{\lambda}\frac{\lambda R}{r_0}}\right)^{-1}\right).
\end{align*}
\end{proposition}

\paragraph{Proof of Proposition \ref{prop peeling stronger}:}

given any $r_0\in (0,R]$, take $r_k=\lambda^k r_0$, $k=1,\cdots, \lceil\log_\lambda\frac{R}{r_0}\rceil$. Note that $\lceil\log_\lambda\frac{R}{r_0}\rceil\leq \log_\lambda\frac{\lambda R}{r_0}$. 

We use a union bound to establish that $\sup_{\frac{r}{\lambda} T(f)\leq r}(\P-\Pn)g_f\leq \psi(r;\delta)$ holds for all these $r_{k}$ simultaneously: $\forall \delta\in(0,1)$, with probability at least $1-\delta$, 
\begin{align*}
\sup_{r_{k-1} \leq T(f)\leq r_k}(\P-\Pn)g_f\leq \psi\left(r_k;\frac{\delta}{\log_2\frac{2R}{r_0}}\right), \quad  k=1,\cdots, \left\lceil\log_2\frac{R}{r_0}\right\rceil.
\end{align*}

For any fixed $f\in\F$, if $T(f)\leq r_0$ is false, then let $k$ be the non-negative integer such that $\lambda^k r_0< T(f)\leq \lambda^{k+1}r_0$, and we further know that $r_{k+1}=\lambda^{k+1}r_0\leq \lambda T(f)$.
Therefore, with probability at least $1-\delta$,
\begin{align*}
   (\P-\Pn) g_f
    &\leq  \sup_{\tilde{f}\in\F: r_k\leq T(\tilde{f})\leq r_{k+1}}(\P-\P_n)g_{\tilde{f}}\\
    &\leq  \psi\left(r_{k+1};\frac{\delta}{\log_\lambda\frac{\lambda R}{r_0}}\right)\\
    &\leq   \psi\left(\lambda T(f);\frac{\delta}{\log_\lambda\frac{\lambda R}{r_0}}\right).
\end{align*}

Therefore, with probability at least $1-\delta$, $\forall f\in\F$, either $T(f)\leq r_0$ or
\begin{align*}
    (\P-\Pn) g_f
    \leq  \psi\left(\lambda T(f);\frac{\delta}{\log_\lambda \frac{\lambda R}{r_0}}\right).
\end{align*}
This completes the proof of Proposition \ref{prop peeling stronger}.
\hfill$\square$
\paragraph{}

Clearly, Proposition \ref{prop peeling} can be viewed as a corollary of Proposition \ref{prop peeling stronger}. We now present an implication of Proposition \ref{prop peeling}, which may be more convenient to use for some problems.

\begin{proposition}[\bf a variant of the ``uniform localized convergence" argument]\label{prop variant peeling}
For a function class  $\G=\{g_f:f\in\F\}$ and   functional $T:\F\rightarrow [0,R]$, assume there is a function  $\psi(r;\delta)$ (possibly depending on the samples), which is non-decreasing with respect to $r$ and satisfies that $\forall \delta\in(0,1)$, $\forall r\in[0,R]$, with probability at least $1-\delta$,
\begin{align*}
\sup_{f\in\F: T(f)\leq r}(\P-\Pn)g_f\leq \psi(r;\delta).
\end{align*}
 Then, given any $\delta\in(0,1)$ and $r_0\in (0,R]$, with probability at least $1-\delta$,  for all $f\in \F$,
\begin{align*}
    (\P-\Pn) g_f\leq  \psi\left(2T(f)\lor r_0;\frac{\delta}{C_{r_0}}\right),
\end{align*}
where $C_{r_0}=2\log_2\frac{2R}{r_0}$.
\end{proposition}

\paragraph{Proof of Proposition \ref{prop variant peeling}:} From Proposition \ref{prop peeling} we know that with probability at least $1-\frac{\delta}{2}$, for all $f\in\F$, either $T(f)\leq r_0$ or 
\begin{align}\label{eq: event variant peeling}
    (\P-\Pn)g_f\leq \psi\left(2T(f);\frac{\delta}{2}\left(\log_2\frac{2R}{r_0}\right)^{-1}\right)=\psi\left(2T(f); \frac{\delta}{C_{r_0}}\right).
\end{align}
We denote the event 
\begin{align*}
    \mathcal{A}_1=\left\{\text{there exists $f\in\F$ such that $T(f)\geq r_0$ and $(\P-\Pn)g_f>\psi\left(2T(f);\frac{\delta}{C_{r_0}}\right)$}\right\}.
\end{align*}
Then from \eqref{eq: event variant peeling}, we have
\begin{align}\label{eq: event 1 variant peeling}
    \text{Prob}(\mathcal{A}_1)\leq \frac{\delta}{2}.
\end{align}

We denote the event 
\begin{align*}
    \mathcal{A}_2=\left\{\text{there exists $f\in\F$ such that $T(f)> r_0$ and $(\P-\Pn)g_f>\psi\left(r_0;\frac{\delta}{C_{r_0}}\right)$}\right\}.
\end{align*}
Then from the surrogate property of $\psi$ and the fact $C_{r_0}\geq 2$, we have
\begin{align}\label{eq: event 2 variant peeling}
    \text{Prob}(\mathcal{A}_2)\leq \frac{\delta}{C_{r_0}}\leq \frac{\delta}{2}.
\end{align}
Combining \eqref{eq: event 1 variant peeling} and \eqref{eq: event 2 variant peeling} by an union bound, we have
\begin{align*}
    \text{Prob}(\mathcal{A}_1\cup\mathcal{A}_2)\leq \text{Prob}(\mathcal{A}_1)+\text{Prob}(\mathcal{A}_2)\leq \delta.
\end{align*}
From the above argument, it is straightforward to prove that with probability at least $1-\delta$, for all $f\in\F$, 
\begin{align*}
    (\P-\Pn)g_f\leq \psi\left(2T(f)\lor r_0;\frac{\delta}{C_{r_0}}\right).
\end{align*}
This completes the proof of Proposition \ref{prop variant peeling}.
\subsection{Proof of Theorem \ref{thm slow}}\label{subsec appendix thm slow}

Let $\F$ be the excess loss class in \eqref{eq: excess loss class}, and define its member $f$ by $f(z)=\ell(h;z)-\ell(h^*;z), \forall z\in\Z$. Clearly, $\F$ is uniformly bounded in  $[-2B,2B]$. Let $T(f)=\P[f^2]$. Define $\fh$ by $\fh(z)=\ell(\hh_{\ERM};z)-\ell(h^*;z), \forall z\in\Z$.

For a fixed $r_0\in(0,4B^2)$, Denote $C_{r_0}={2\log_2\frac{8B^2}{r_0}}$. 
From now to the end of this proof, we will prove the generalization error bound on the event 
\begin{align}\label{eq: event loss rate}
\mathcal{A}=\left\{\text{for all $f\in\F$, $(\P-\Pn)f\leq \psi\left(2T(f)\lor r_0 ;\frac{\delta}{C_{r_0}}\right)$}\right\}.
\end{align}From Proposition \ref{prop variant peeling} we know that
\begin{align*}
     \text{Prob}(\mathcal{A})\geq 1-\delta.
\end{align*}
This means that proving the generalization error bound on the event $\mathcal{A}$ suffices to prove the theorem.

Denote $g(z)=\ell(h;z)-\inf_{\Hy}\ell(h;z)$ and $\gh(z)=\ell(\hh_{\ERM};z)-\inf_{\Hy}\ell(h;z)$. Let $T(g)=\P[g^2]$. We have $$f(z)= g(z)-(\ell(h^*;z)-\inf_{\Hy}\ell(h;z)),\quad  \forall z,$$ which implies that
\begin{align*}
    \P[f^2]\leq 2\P[g^2]+2\P[(\ell(h^*;z)-\inf_{\Hy}\ell(h;z))^2]\\
    \leq 2\P[g^2]+4B\LL\leq 4\P[g^2]\lor8B\LL.
\end{align*}
Therefore, we have
\begin{align}\label{eq: relate f g}
    T(\fh)\leq 4T(\gh)\lor8B\LL.
\end{align}
From the property of ERM, we have $\Pn\fh\leq 0$, which implies that
\begin{align}\label{eq: conclusion before locailzation}
    \mathcal{E}(\hh_{\ERM})\leq (\P-\Pn) \fh\leq \psi\left(2T(\fh)\lor r_0;\frac{\delta}{C_{r_0}}\right).
\end{align}

From \eqref{eq: relate f g} and \eqref{eq: conclusion before locailzation} we have
\begin{align}\label{eq: inequality use g}
   \P\gh -\LL= \mathcal{E}({\hh_{\ERM}})\leq \psi\left(8T(\gh)\lor 16B\LL\lor r_0;\frac{\delta}{C_{r_0}}\right).
\end{align}

Since $\gh(z)\in[0,2B]$ for all $z$, we have $T(\gh)\leq 2B\P\gh$. From this fact and \eqref{eq: inequality use g} we obtain
\begin{align*}
    T(\gh)&\leq 2B\P\gh\nonumber\\&\leq  2B\left(\LL+\psi\left(8T(\gh)\lor 16B\LL\lor r_0;\frac{\delta}{C_{r_0}}\right)\right)\nonumber\\
    &=2B\LL+2B\psi\left(8T(\gh)\lor 16B\LL\lor r_0;\frac{\delta}{C_{r_0}}\right).
    \end{align*}
Whether $B\LL\leq 2B\psi\left(8T(\gh)\lor 16B\LL\lor r_0;\frac{\delta}{C_{r_0}}\right)$ or $B\LL> 2B\psi\left(8T(\gh)\lor 16B\LL\lor r_0;\frac{\delta}{C_{r_0}}\right)$, the above inequality always implies that    
    \begin{align}\label{eq: thm erm fixed point inequality}
   T(\gh) &\leq 3B\LL\lor 6B\psi\left(8T(\gh)\lor 16B\LL\lor r_0;\frac{\delta}{C_{r_0}}\right)\nonumber\\
    &\leq 3B\LL\lor6B\psi\left(8T(\gh);\frac{\delta}{C_{r_0}}\right)\lor 6B\psi\left(16B\LL\lor r_0;\frac{\delta}{C_{r_0}}\right).
\end{align}
Let $r^*$ be the fixed point of $6B\psi\left(8r;\frac{\delta}{C_n}\right)$. From the definition of  fixed points whether $2B\LL\lor \frac{r_0}{8}\leq r^*$ or $2B\LL\lor \frac{r_0}{8}>r^*$, we always have $$6B\psi\left(16B\LL\lor r_0;\frac{\delta}{C_{r_0}}\right)\leq r^*\lor 2B\LL\lor \frac{r_0}{8}.$$ Combining the above inequality with  \eqref{eq: thm erm fixed point inequality}, we have
\begin{align*}
    T(\gh)\leq  3B\LL\lor 6B\psi\left(8T(\gh);\frac{\delta}{C_{r_0}}\right)\lor r^*\lor \frac{r_0}{8}.
\end{align*} From the above inequality and again the definition of fixed points, it is straightforward to prove that
\begin{align*}
    T(\gh)\leq 3B\LL\lor r^*\lor\frac{r_0}{8}.
\end{align*}
Combining the above inequality with \eqref{eq: relate f g}, we have
\begin{align*}
    T(\fh)\leq 12B\LL\lor 4r^*\lor \frac{r_0}{2}.
\end{align*}

 From the above inequality and \eqref{eq: conclusion before locailzation} we have
\begin{align}\label{eq: use later corollary}
    \mathcal{E}(\hh_{\ERM})\leq (\P-\Pn)\fh\leq \psi\bigg(24B\LL\lor 8 r^*\lor r_0 ;\frac{\delta}{C_{r_0}}\bigg),
    \end{align}
    which implies that
    \begin{align*}
       \mathcal{E}(\hh_{\ERM})  \leq \psi\left(24B\LL;\frac{\delta}{C_{r_0}}\right)\lor \psi\left(8 r^*\lor r_0;\frac{\delta}{C_{r_0}}\right).
    \end{align*}
Recall that $r^*$ is the fixed point of $6B\psi(8r;\frac{\delta}{C_{r_0}})$. Since $r^*\lor \frac{r_0}{8}\geq r^*$, from the definition of fixed points we have
\begin{align*}
     6B\psi(8r^*\lor 2r_0;\frac{\delta}{C_{r_0}})\leq r^*\lor \frac{r_0}{8}.
\end{align*}
So we finally obtain
\begin{align*}
    \mathcal{E}(\hh_{\ERM})\leq \psi\left(24B\LL;\frac{\delta}{C_{r_0}}\right)\lor \frac{r^*}{6B}\lor \frac{r_0}{48B}.
\end{align*}
Recall that the generalization error bound holds true on the event $\mathcal{A}$ defined in \eqref{eq: event loss rate}, whose measure is at least $1-\delta$. This completes the proof.
\hfill$\square$

\subsection{Estimating the loss-dependent rate from data}\label{subsubsec evaluate loss rate}
In the remarks following Theorem \ref{thm slow}, we comment that fully data-dependent loss-dependent bounds can be derived using the empirical ``effective loss," $\Pn[\ell(\hh_{\ERM};z)-\inf_{\Hy}\ell(h;z)]$  to estimate the unknown parameter $\LL$. Here we present the full details and some discussion of this approach.
\begin{theorem}[{\bf estimate of the loss-dependent rate from data}]\label{thm estimate loss rate}
Recall the term  $\LL$ is $\P[\ell(h^*;z)-\inf_{\Hy}\ell(h^*;z)]$ and denote $\widehat{\LL}=\Pn[\ell(\hh_{\ERM};z)-\inf_{\Hy}\ell(h;z)]$. Under the conditions of Theorem \ref{thm slow}, setting $C_n=2\log_2 n+6$, then for any fixed $\delta\in(0,\frac{1}{2})$, with probability at least $1-2\delta$, we have
\begin{align}\label{eq: estimate loss dependent rate bound}
    \mathcal{E}(\hh_{\ERM})\leq \psi\left(c B\widehat{\LL};\frac{\delta}{C_n}\right)\lor \frac{c r^*}{B}\lor\frac{cB\log\frac{2}{\delta}}{n}
\end{align}
and
\begin{align}\label{eq: estimate loss dependent rate relation}
    \LL\leq c_1\left(\widehat{\LL}\lor \frac{r^*}{B}\lor\frac{B\log\frac{2}{\delta}}{n}\right)\leq c_2\left(\LL\lor \frac{r^*}{B}\lor\frac{B\log\frac{2}{\delta}}{n}\right),
\end{align}
where $c,c_1,c_2$ are absolute constants.
\end{theorem}
\paragraph{Remarks.} 
1) The  $B\log\frac{2}{\delta}/n$ terms  \eqref{eq: estimate loss dependent rate bound} and \eqref{eq: estimate loss dependent rate relation} are negligible, because $r^*$ is at least  of order $B^2\log\frac{1}{\delta}/n$ for most practical applications. This order is unavoidable in traditional ``local Rademacher complexity" analysis and two-sided concentration inequalities.

\noindent 2) The generalization error bound \eqref{eq: estimate loss dependent rate bound} shows that without knowledge of $\LL$, one can estimate the order of our loss-dependent rate by using $\widehat{\LL}=\Pn[\ell(\hh_{\ERM};z)-\inf_{\Hy}\ell(h;z)]$ as a proxy. Despite replacing $\LL$ by $\widehat{\LL}$, other quantities in the bound remain unchanged in order. 

\noindent 3) The inequality \eqref{eq: estimate loss dependent rate relation} shows that the estimation of $\LL$ is tight.

\paragraph{Proof of Theorem \ref{thm estimate loss rate}:}
from the definitions, we know that $\LL=\P[\ell(h^*;z)-\inf_{\Hy}\ell(h^*;z)]$, $\widehat{\LL}=\Pn[\ell(\hh_{\ERM};z)-\inf_{\Hy}\ell(h;z)]$ and $\P\ell(h^*;z)\leq\P\ell(\hh_{\ERM};z)$. As a result, we have
\begin{align}\label{eq: bound for estimating optimal effective loss 1}
    \LL-\widehat{\LL}= \P\ell(h^*;z)-\Pn \ell(\hh_{\ERM};z)-(\P-\Pn)[\inf_{\Hy}\ell(h;z)] \nonumber
   \\ \leq (\P-\Pn)\ell(\hh_{\ERM};z)-(\P-\Pn)[\inf_{\Hy}\ell(h;z)] \nonumber\\
   =(\P-\Pn)\hat{f}+(\P-\Pn)[\ell(h^*;z)-\inf_{\Hy}\ell(h;z)],
\end{align}
where $\fh$ is defined by $\fh(z)=\ell(\hh_{\ERM};z)-\ell(h^*;z), \forall z\in\Z$.

We take $r_0=\frac{B^2}{n}$ in Theorem \ref{thm slow}, and denote $C_n:=C_{r_0}=2\log_2 n+6$. From \eqref{eq: use later corollary} in the proof of Theorem \ref{thm slow}, on the event $\mathcal{A}$ defined in \eqref{eq: event loss rate} (whose measure is at least $1-\delta$),  
\begin{align}\label{eq: use theorem loss rate}
   \mathcal{E}(\hh_{\ERM})\leq (\P-\Pn)\hat{f}\leq \psi( 24B\LL\lor8r^*\lor \frac{B^2}{n};\frac{\delta}{C_{n}}),
\end{align}
where $\fh$ is defined by $\fh(z)=\ell(\hh_{\ERM};z)-\ell(h^*;z), \forall z\in\Z$.

Since $3B\LL\lor r^*\lor\frac{B^2}{4n}\geq r^*$, from the  definition of  fixed points we have
\begin{align}\label{eq: bound for estimating optimal effective loss 2}
    (\P-\Pn)\hat{f}\leq \psi\left( 8\left(3B\LL\lor r^*\lor\frac{B^2}{8n}\right);\frac{\delta}{C_{n}}\right)
    \nonumber\\
    \leq \frac{3B\LL\lor r^*\lor\frac{B^2}{8n}}{6B}\leq\frac{\LL}{2}+ \frac{r^*}{6B}+ \frac{B}{48n}.
\end{align}
This result holds together with the result of Theorem \ref{thm slow} on the event $\mathcal{A}$.

The random variable $\ell(h^*;z)-\inf_{\Hy}\ell(h;z)$ is uniformly bounded by $[0,2B]$. From Bernstein's inequality and the fact $\var[\ell(h^*;z)-\inf_{\Hy}\ell(h;z)]\leq 2B\LL$, 
with probability at least $1-\delta$, 
\begin{align}\label{eq: bound for estimating optimal effective loss 3}
\left|(\P-\Pn)[\ell(h^*;z)-\inf_{\Hy}\ell(h;z)]\right|\leq \sqrt{\frac{4B\LL\log\frac{2}{\delta}}{n}}+\frac{2B\log\frac{2}{\delta}}{n}\leq \frac{\LL}{4}+\frac{3B\log\frac{2}{\delta}}{n}.
\end{align}
Consider the event 
$$\mathcal{A}_3=\mathcal{A}\cup\{\text{inequality \eqref{eq: bound for estimating optimal effective loss 3} holds true}\},$$ whose measure is at least $1-2\delta$. On the event  $\mathcal{A}_3$, from
 inequalities \eqref{eq: bound for estimating optimal effective loss 1} \eqref{eq: bound for estimating optimal effective loss 2} \eqref{eq: bound for estimating optimal effective loss 3}, it is straightforward to show that
\begin{align*}
    \LL-\widehat{\LL}\leq \frac{3}{4}\LL+\frac{r^*}{6B}+\frac{4B\log\frac{2}{\delta}}{n},
\end{align*}
which implies 
\begin{align}\label{eq: estimate loss 1}
   \LL\leq  4\widehat{\LL}+\frac{2 r^*}{3B}+\frac{16B\log\frac{2}{\delta}}{n}.
\end{align}
From this result and \eqref{eq: use theorem loss rate},
it is straightforward to show that
\begin{align*}
    \mathcal{E}(\hh_{\ERM})\leq 
     \psi\left(c B\widehat{\LL};\frac{\delta}{C_n}\right)\lor \frac{cr^*}{n}\lor \frac{cB\log\frac{2}{\delta}}{n},
\end{align*}
where $c$ is an absolute constant.

We also have
\begin{align*}
    \widehat{\LL}-\LL=\Pn \ell(\hh_{\ERM})-\P\ell(h^*;z)-(\Pn-\P)[\inf_{\Hy}\ell(h;z)]\\
    \leq (\Pn-\P)\ell(h^*;z)-(\Pn-\P)[\inf_{\Hy}\ell(h;z)]\\
    =(\Pn-\P)[\ell(h^*;z)-\inf_{\Hy}\ell(h;z)].
\end{align*}
From this result and \eqref{eq: bound for estimating optimal effective loss 3}, on the event $\mathcal{A}_3$,
\begin{align}\label{eq: estimate loss 2}
    \widehat{\LL}\leq \frac{5}{4}\LL+\frac{3B\log\frac{2}{\delta}}{n}. 
\end{align}
Combine \eqref{eq: estimate loss 1} and \eqref{eq: estimate loss 2} we obtain
\begin{align*}
    \LL\leq c_1\left(\widehat{\LL}\lor \frac{r^*}{B}\lor\frac{B\log\frac{2}{\delta}}{n}\right)\leq c_2\left(\LL\lor \frac{r^*}{B}\lor\frac{B\log\frac{2}{\delta}}{n}\right),
\end{align*}
where $c_1$ and $c_2$ are absolute constants. This completes the proof.
\hfill$\square$

\subsection{Proof of Theorem \ref{thm variance rate}}\label{subsec appendix them slow localized complexity penalization}

The main goal of this subsection is to prove Theorem \ref{thm variance rate}. We first prove Theorem \ref{thm variance second stage} (the bound \eqref{eq: difficult variance} in the main paper),  a guarantee for the second-stage moment penalized estimator $\hh_{\MP}$. In order to prove Theorem \ref{thm variance rate}, we then  combine Theorem \ref{thm variance second stage} with a guarantee for the first-stage empirical risk minimization (ERM) estimator.

\subsubsection{Analysis for the second-stage moment-penalized estimator}
\begin{theorem}[{\bf variance-dependent rate of the second-stage estimator}]\label{thm variance second stage}
Given arbitrary preliminary estimate $\LLL\in [-B,B]$, the generalization error of the moment-penalized estimator $\hh_{\MP}$ in Strategy \ref{strategy estimator} is bounded by
\begin{align*}
    \mathcal{E}(\hh_{\MP})\leq 2\psi\left(c_0 \left[\V\lor (\LLL-\LL_0)^2\lor r^*\right];\frac{\delta}{C_n}\right),
\end{align*}
with probability at least $1-\delta$, where $c_0$ is an absolute constant and $r^*$ is  the fixed point of  $16B\psi(r;\frac{\delta}{C_n})$.
\end{theorem}

\paragraph{Proof of Theorem \ref{thm variance second stage}:}
the proof of Theorem \ref{thm variance second stage} consist of four parts.

\paragraph{Part \RNum{1}: use $\bm{\psi}$ to upper bound localized empirical processes}
Let $\F$ be the excess loss class in \eqref{eq: excess loss class}, and define its member $f$ is defined by $f(z)=\ell(h;z)-\ell(h^*;z), \forall z\in\Z$. We have the following lemma.
\begin{lemma}[{\bf bound on localized empirical processes}]\label{lemma first part variance rate}
Given a fixed $\delta_1\in(0,1)$, let $r_1^*(\delta_1)$ be the fixed point of $16B\psi(r;\delta_1)$ where $\psi$ is defined in Strategy \ref{strategy estimator}. Then with probability at least $1-\delta_1$, for all $r>0$, 
\begin{align}\label{eq: psi first part variance rate}
    \sup_{\P[f^2]\leq r}(\P-\Pn)f\leq \psi\left(r\lor r_1^*(\delta_1);\delta_1\right).
\end{align}
\end{lemma}
\paragraph{Proof of Lemma \ref{lemma first part variance rate}:}
  clearly, $\F$ is uniformly bounded in $[-2B,2B]$. 
When $\P[f^2]\leq r$, we have $\P[f^4]\leq 4B^2r$. From Lemma \ref{lemma bartlett} (the two-sided version of its second inequality), with probability at least $1-\frac{\delta_1}{2}$,
\begin{align*}
   &\  \sup_{{\P[f^2]\leq r}}\left|(\P-\Pn)f^2\right|\nonumber\\
    \leq &\  4\mathfrak{R}_n\{f^2:\P[f^2]\leq r\}+2B\sqrt{\frac{2r\log\frac{8}{\delta_1}}{n}}+\frac{18B^2\log\frac{8}{\delta_1}}{n}\nonumber\\
    \leq &\  16B\mathfrak{R}_n\{f:\P[f^2]\leq r\}+2B\sqrt{\frac{2r\log\frac{8}{\delta_1}}{n}}+\frac{18B^2\log\frac{8}{\delta_1}}{n},
\end{align*}
where the last inequality follows from the Lipchitz contraction property of Rademahcer complexity (see, e.g., [\citealp{meir2003generalization}, Theorem 7]), and the fact that for all $f_1,f_2\in\F$, $|f_1^2(z)-f_2^2(z)|\leq 4B|f_1(z)-f_2(z)|$.
We conclude that with probability at least $1-\frac{\delta_1}{2}$,
\begin{align}\label{eq: temp E4 1.1}
    \sup_{{\P[f^2]\leq r}}\left|(\P-\Pn)f^2\right|\leq \varphi_{\delta_1}(r),
\end{align}
where $\varphi_{\delta_1}(r):= 16B\mathfrak{R}_n\{f:\P[f^2]\leq r\}+2B\sqrt{\frac{2r\log\frac{8}{\delta_1}}{n}}+\frac{18B^2\log\frac{8}{\delta_1}}{n}$.

Denote $r_2^*(\delta_1)$  the fixed point of $4\varphi_{\delta_1}(r)$ (the fixed point must exist as $4\varphi_{\delta_1}(r)$ is a non-decreasing, non-negative and bounded function).
From \eqref{eq: temp E4 1.1} and the fact that $r_2^*(\delta_1)$ is the fixed point of $4\varphi_{\delta_1}(r)$,  if $r > r_2^*(\delta_1)$, then with probability at least $1-\frac{\delta_1}{2}$, 
\begin{align}
\sup_{\P[f^2]\leq r}\left|(\P-\Pn)f^2\right|\leq \frac{r}{4}.\label{eq: deviation variance 1}
\end{align}
\eqref{eq: deviation variance 1} implies that with probability at least $1-\frac{\delta_1}{2}$, for all $r > r_2^*(\delta_1)$, $\P[f^2]\leq r$ implies that
\begin{align}\label{eq: first center variance rate} \Pn[f^2]\leq \frac{5}{4}r\leq2r.
\end{align}

Again from the two-sided version of the second inequality in Lemma \ref{lemma bartlett},  we know that with probability at least $1-\frac{\delta_1}{2}$, 
\begin{align*}
    \sup_{\P[f^2]\leq r}\left|(\P-\Pn)f\right|\leq 4\mathfrak{R}_n\{f:\P[f^2]\leq r\}+\sqrt{\frac{2r\log\frac{8}{\delta_1}}{n}}+\frac{9B\log\frac{8}{\delta_1}}{n}.
\end{align*}
Combining the above inequality and \eqref{eq: first center variance rate} using a union bound,
we know that with probability at least $1-\frac{\delta_1}{2}-\frac{\delta_1}{2}=1-\delta_1$, if $r> r_2^*(\delta_1)$, then
\begin{align}\label{eq: from P to Pn}
    \sup_{\P[f^2]\leq r}(\P-\Pn)f\leq 4\mathfrak{R}_n\{f:\P[f^2]\leq r\}+\sqrt{\frac{2r\log\frac{8}{\delta_1}}{n}}+\frac{9B\log\frac{8}{\delta_1}}{n}\nonumber\\\leq 4\mathfrak{R}_n\{f:\Pn[f^2]\leq 2r\}+\sqrt{\frac{2r\log\frac{8}{\delta_1}}{n}}+\frac{9B\log\frac{8}{\delta_1}}{n}.
\end{align}

Recall that the $\psi$ function satisfies that $\forall r>0$,
\begin{align*}
    4\mathfrak{R}_n\{f:\Pn[f^2]\leq 2r\}+\sqrt{\frac{2r\log\frac{8}{\delta_1}}{n}}+\frac{9B\log\frac{8}{\delta_1}}{n}\leq \psi(r;\delta_1).
\end{align*} From this fact and \eqref{eq: from P to Pn}, we see that  with probability at least $1-\delta_1$, for all $r>0$,
\begin{align}\label{eq: nearly first part variance rate}
    \sup_{\P[f^2]\leq r}(\P-\Pn)f\leq \psi\left(r\lor r_2^*(\delta_1);\delta_1\right).
\end{align}

From \eqref{eq: nearly first part variance rate}, in order to prove the result \eqref{eq: psi first part variance rate} in Lemma \ref{lemma first part variance rate}, we only need to prove that
 \begin{align}\label{eq: compare fixed point first part variance rate}
     r_2^*(\delta_1)\leq r_1^*(\delta_1).
 \end{align}
 Assume this is not true, i.e.  $r_2^*({\delta_1})> r_1^*(\delta_1)$. Since $r_1^*(\delta_1)$ is the fixed point of $16B\psi(r;\delta_1)$, from the definition of  fixed points we have $$r_2^*({\delta_1})> 16B\psi(r_2^*({\delta_1});\delta_1).$$ From  the definitions of $\psi$ and $\varphi_{\delta_1}$, for all $r>r_1^*({\delta_1})$, $$4\varphi_{\delta_1}(r)\leq 16B\psi(r;\delta_1).$$ From the above two  inequalities and $r_2^*({\delta_1})> r_1^*(\delta_1)$, we have
\begin{align}\label{eq: variance rate fixed point contradiction 1}
    r_2^*({\delta_1})> 16B\psi(r_2^*({\delta_1});\delta_1)\geq 4\varphi_{\delta_1}(r_2^*({\delta_1})).
\end{align} From the fact that $r_2^*({\delta_1})$ is the fixed point of $4\varphi_{\delta_1}$, we have
\begin{align}\label{eq: variance rate fixed point contradiction 2}
   4\varphi_{\delta_1}(r_2^*(\delta_1)) = r_2^*(\delta_1).
\end{align}
The above two inequalities \eqref{eq: variance rate fixed point contradiction 1} and \eqref{eq: variance rate fixed point contradiction 2} result in a contradiction. So the assumption $r_2^*(\delta_1)> r_1^*(\delta_1)$ is false. Therefore $r_2^*(\delta_1)\leq r_1^*(\delta_1)$, and this completes the proof of Lemma \ref{lemma first part variance rate}.\hfill$\square$

\paragraph{\bf Part \RNum{2}: a ``uniform localized convergence" argument with data-dependent measurement.}\ 

 Based on Lemma \ref{lemma first part variance rate}, we  will modify the proof of Proposition \ref{prop peeling} to obtain a ``uniform localized convergence" argument with the data-dependent ``measurement" functional $\Pn [f^2]$.
 
\begin{lemma}[{\bf a ``uniform localized convergence" argument with the data-dependent ``measurement" functional}]\label{lemma data dependent measure function}
Given a fixed $\delta_1\in(0,1)$, let $r_1^*(\delta_1)$ be the fixed point of $16B\psi(r;\delta_1)$ where $\psi$ is defined in Strategy \ref{strategy estimator}. Then  with probability at least $1-\left(\log_2\frac{8B^2\lor 2r_1^*({\delta_1})}{r_1^*({\delta_1})}+\frac{1}{2}\right)\delta_1$, for all $f\in\F$ either $\P[f^2]\leq r_1^*({\delta_1})$, or 
\begin{align}\label{eq: temp E4 2}
    (\P -\Pn) f \leq \psi\bigg(4\Pn[f^2];\delta_1\bigg).
 \end{align}
\end{lemma}

\paragraph{Proof of Lemma \ref{lemma data dependent measure function}:}
from the definition of $\psi$ and the fact that $r_1^*(\delta_1)$ is the fixed point of $16B\psi(r;\delta_1)$, we know that $r_1^*(\delta_1)\geq \frac{144B^2\log\frac{8}{\delta_1}}{n}>0$. Take $r_0=r_1^*(\delta_1)$.

Take $R=4B^2\lor r_0$ to be a uniform upper bound for $\P f^2$, and take $r_k=2^k r_0, k=1,\cdots, \lceil\log_2\frac{R}{r_0}\rceil$. Note that $\lceil\log_2\frac{R}{r_0}\rceil\leq \log_2\frac{2R}{r_0}$. We use the union bound to establish that $\sup_{\P[f^2]\leq r}(\P-\Pn)f\leq \psi(r;\delta_1)$ holds for all $\{r_{k}\}$ simultaneously:  with probability at least $1-\log_2\frac{2R}{r_0}\delta_1$, 
\begin{align*}
\sup_{\P[f^2]\leq r_k}(\P-\Pn)f\leq \psi(r_k;{\delta_1}), \quad  k=1,\cdots, \left\lceil\log_2\frac{R}{r_0}\right\rceil.
\end{align*}

For any fixed $f\in\F$, if $\P[f^2]\leq {r_0}$ is false, let $k$ be the non-negative integer such that $2^k r_0< \P[g(h;z)^2]\leq 2^{k+1}r_0$. We further have that $r_{k+1}=2^{k+1}r_0\leq 2\P[f^2]$.
Therefore, with probability at least$1-\log_2\frac{2R}{r_0}\delta_1$,
\begin{align}\label{eq: middle sample dependent variant}
   \P f
    \leq \ &\Pn f+\sup_{\tilde{f}\in\F: \P[\tilde{f}^2]\leq r_{k+1}}(\P-\P_n)\tilde{f}\nonumber\\
    \leq \ & \Pn f+\psi(r_{k+1};\delta_1)
\end{align}
By \eqref{eq: temp E4 1.1}  we know that with probability at least $1-\frac{\delta_1}{2}$, \begin{align*}
    \sup_{\P[f^2]\leq r}\left(\P[f^2]-\Pn[f^2]\right)\leq \frac{r}{4}
\end{align*} for all $r>r_0$ (here we have used the fact $r_0=r_1^*(\delta_1)\geq r_2^*(\delta_1)$, which is the result \eqref{eq: compare fixed point first part variance rate} in the proof of Lemma \ref{lemma first part variance rate}). From the union bound, with probability at least $1-(\log_2\frac{2R}{r_0}+\frac{1}{2})\delta_1$, the condition $r_{k+1}\geq \P[f^2]> r_k$ will imply 
\begin{align*}
\P_n[f^2]\geq \P[f^2]-\frac{1}{4}r_{k+1}
\geq \frac{1}{4}r_{k+1},
\end{align*}so 
\begin{align*}
    r_{k+1}\leq 4\Pn[f^2].
\end{align*}
Combining this result with \eqref{eq: middle sample dependent variant}, we have that for all $f$ such that $T(f)>{r_0}$, with probability at least $1-\left(\log_2\frac{2R}{r_0}+\frac{1}{2}\right)\delta_1$,
\begin{align*}
   \P f
    \leq \ & \Pn f+\psi(r_{k+1};\delta_1)\\
    \leq \ & \Pn f+\psi\bigg(4\P_n[f^2];\delta_1\bigg).
\end{align*}
We conclude that with probability at least $1-\left(\log_2\frac{2R}{r_0}+\frac{1}{2}\right)\delta_1$, for all $f\in\F$, either $\P[f^2]\leq r_1^*({\delta_1})$, or 
\begin{align*}
    (\P - \Pn) f\leq \psi\bigg(4\Pn[f^2];\delta_1\bigg).
 \end{align*} This completes the proof of Lemma \ref{lemma data dependent measure function}.\hfill$\square$

\paragraph{\bf Part \RNum{3}: specify the moment-penalized estimator and its error bound.}\

We define the event $$\mathcal{A}_{1}=\left\{\text{there exists $f\in\F$ such that $\P[f^2]\geq r_0$ and $(\P-\Pn)f>\psi\left(4\Pn[f^2];\delta_1\right)$}\right\}.$$  Lemma \ref{lemma data dependent measure function} has proven that  \begin{align}\label{eq: event 1 bound variance rate}
\textup{Prob}(\mathcal{A}_{1})\leq \left(\log_2\frac{8B^2\lor2r_1^*(\delta_1)}{r_1^*(\delta_1)}+\frac{1}{2}\right)\delta_1.
\end{align}
We denote the event $$\mathcal{A}_{2}=\{\textup{there exists $f\in\F$ such that $\P[f^2]\leq r_0$ and $(\P-\Pn)f>\psi(r_0;\delta_1)$}\}.$$
 Due to the surrogate property of $\psi$, we have
\begin{align}\label{eq: event 2 bound variance rate}
    \text{Prob}\left(\mathcal{A}_{2}\right)\leq \delta_1.
\end{align}
Denote the event
\begin{align*}
\mathcal{A}=\left\{\text{for all $f\in\F$, }(\P-\Pn) f\leq \psi\bigg(4\Pn[f^2]\lor r_1^*(\delta_1);\delta_1\bigg)\right\}.
\end{align*}
From \eqref{eq: event 1 bound variance rate} and \eqref{eq: event 2 bound variance rate}, it is straightforward to prove that
\begin{align}\label{eq: before regularization variance rate}
    \text{Prob}(\mathcal{A})&\geq 1-\text{Prob}(\mathcal{A}_1)-\text{Prob}(\mathcal{A}_2)\nonumber\\&\geq 1-\left(\log_2\frac{8B^2\lor2r_1^*(\delta_1)}{r_1^*(\delta_1)}+\frac{1}{2}\right)\delta_1-\delta_1\nonumber\\&\geq  1-\left(\log_2\frac{8B^2\lor2r_1^*(\delta_1)}{r_1^*(\delta_1)}+\frac{3}{2}\right)\delta_1.
    \end{align}

 Denote $w(h;z)=\ell(h;z)-\LLL$.  Then $f(z)=w(h;z)-w(h^*;z), \forall z\in\Z$, and we have that
\begin{align*}
    4\Pn[f^2]\leq 8\Pn[w(h;z)^2]+8\Pn[w(h^*;z)^2]\\
    \leq 16\Pn[w(h;z)^2]\lor 16\Pn[w(h^*;z)^2].
\end{align*}
From the above conclusion and \eqref{eq: before regularization variance rate} we obtain that on the event $\mathcal{A}$,
\begin{align}\label{eq: desire specify regularization}
    \mathcal{E}(h)+\Pn \ell(h^*;z)\leq \Pn\ell(h;z)+\psi(4\Pn[\hat{f}^2]\lor r_1^*({\delta_1});\delta_1)
    \nonumber\\
    \leq \Pn(h;z)+\psi\bigg(16\Pn [w(h;z)^2]\lor 16\Pn [w(h^*;z)^2]\lor r_1^*({\delta_1});\delta_1\bigg)\nonumber\\
    \leq \Pn(h;z)+\psi\bigg(16\Pn [w(h;z)^2]\delta_1\bigg)+\psi\bigg(16\Pn [w(h^*;z)^2]\lor r_1^*({\delta_1});\delta_1\bigg).
\end{align}

We specify the moment-penalized estimator to be
\begin{align*}
\hh_{\MP}=\arg\min_{\Hy}\left\{\Pn \ell(h;z)+\psi\bigg(16\P_n[(\ell(h;z)-\LLL)^2];\delta_1\bigg)\right\}.
\end{align*} Then  we have
\begin{align}\label{eq: property MP}
    \Pn \ell(\hh_{\MP};z)+\psi\bigg(16\P_n[w(\hh_{\MP};z)^2];\delta_1\bigg)\leq  \Pn \ell(h^*;z)+\psi\bigg(16\P_n[w(h^*;z)^2];\delta_1\bigg)
\end{align}

Therefore, on the event $\mathcal{A}$,
\begin{align}\label{eq: second stage rate empirical moment}
    \mathcal{E}(\hh_{\MP})\leq \Pn\ell(\hh_{\MP};z)+\psi\bigg(16\Pn[w(\hh_{\MP};z)^2];\delta_1\bigg)+\psi\bigg(16\Pn [w(h^*;z)^2]\lor r_1^*({\delta_1});\delta_1\bigg)-\Pn\ell(h^*;z)
   \nonumber \\ =\argmin_{\Hy}\left\{\Pn\ell(h;z)+\psi\bigg(16\Pn[w(h;z)];\delta_1\bigg)\right\}-\Pn\ell(h^*;z)+\psi\bigg(16\Pn [w(h^*;z)^2]\lor r_1^*({\delta_1});\delta_1\bigg)\nonumber\\
   \leq \psi\bigg(16\P_n[w(h^*;z)^2];\delta_1\bigg)+\psi\bigg(16\Pn[w(h^*;z)^2]\lor r_1^*({\delta_1});\delta_1\bigg)
   \nonumber\\ \leq 2\psi\bigg(16\Pn[w(h^*;z)^2]\lor r_1^*({\delta_1});\delta_1\bigg),
\end{align}
where the first inequality is due to \eqref{eq: desire specify regularization} and the second inequality is due to \eqref{eq: property MP}.

From Bernstein's inequality at the single element $h^*$, for any fixed $\delta_2\in(0,1)$, with probability at least $1-\delta_2$,
\begin{align}\label{eq: bernstein variance rate}
    \Pn [w(h^*;z)^2]\leq \P[w(h^*;z)^2]+ 2B\sqrt{\frac{2\P[w(h^*;z)^2]\log\frac{2}{\delta_2}}{n}}+\frac{4B^2\log\frac{2}{\delta_2}}{n}\nonumber\\
    \leq 2\P[w(h^*;z)^2]+\frac{6B^2\log\frac{2}{\delta_2}}{n}.
\end{align}
From \eqref{eq: before regularization variance rate} \eqref{eq: second stage rate empirical moment} \eqref{eq: bernstein variance rate}, with probability at least $$\text{Prob}(\mathcal{A})-\delta_2\geq 1-\left(\log_2\frac{8B^2\lor2r_1^*(\delta_1)}{r_1^*(\delta_1)}+\frac{3}{2}\right)\delta_1-\delta_2,$$ we have 
\begin{align}\label{eq: part 3 variance penalization}
   \mathcal{E}(\hh_{\MP})
   &\leq 2\psi\left(16\Pn[w(h^*;z)]\lor r_1^*({\delta_1})\lor \frac{B^2}{n};\delta_1\right)\nonumber \nonumber\\
   &\leq 2\psi\left(\left(32\P[w(h^*;z)^2]+\frac{96B^2\log\frac{2}{\delta_2}}{n}\right)\lor r_1^*({\delta_1})\lor \frac{B^2}{n};\delta_1\right),
\end{align}
where the first inequality is due to  \eqref{eq: second stage rate empirical moment} and the second inequality is due to \eqref{eq: bernstein variance rate}. 

\paragraph{\bf Part \RNum{4}: final steps.}\ 

From the definition of $\psi$ and the fact that $r_1^*(\delta_1)$ is the fixed point of $16B\psi(r;\delta_1)$, we know that 
\begin{align}\label{eq: lower bound on r1 variance rate}
    r_1^*(\delta_1)\geq \frac{144B^2\log\frac{8}{\delta_1}}{n}.
\end{align} 
Denote $C_n:=2\log_2 n+5$ and take
\begin{align*}
    \delta_1=\frac{\delta}{C_n},
\end{align*}
then we have
\begin{align*}
2\log_2\frac{8B^2\lor 2r_1^*(\delta_1) }{r_1^*(\delta_1)}+3\leq \max\left\{2\log_2\frac{8n}{144\log 8}, 2+3\right\} \\
\leq \max\{2\log_2 n, 5\}\leq C_n,
\end{align*}
so 
\begin{align}\label{eq: specification delta1}
    \left(\log_2\frac{8B^2\lor 2r^*_1(\delta_1) }{r^*_1(\delta_1)}+\frac{3}{2}\right)\delta_1\leq  {\frac{\delta}{2}}.
\end{align}

 Set $r^*= r_1^*({\delta_1})$ and take $\delta_2=\frac{\delta}{2}$. From \eqref{eq: part 3 variance penalization}, we obtain that with probability at least $1-\delta$, the generalization error of $\hh_{\MP}$ is upper bounded by 
\begin{align}\label{eq: geberalization error before}
  \mathcal{E}(\hh_{\MP})\leq  2\psi\bigg(c\left[\P[w(h^*;z)^2]\lor r^*\lor \frac{B^2\log\frac{4}{\delta}}{n}\right];\frac{\delta}{C_n}\bigg),
\end{align}
where $c$ is an absolute constant.
From \eqref{eq: lower bound on r1 variance rate} we have
$r_1^*(\delta_1)\geq \frac{144B^2\log\frac{8C_n}{\delta}}{n}\geq \frac{B^2\log\frac{4}{\delta}}{n}$. Combine this fact with the inequality \eqref{eq: geberalization error before}, we obtain that
\begin{align}\label{eq: result of the first four parts}
   \mathcal{E}(\hh_{\MP})\leq 2\psi\left(c\left[\P[(\ell(h^*;z)-\LLL)^2]\lor r^*\right];\frac{\delta}{C_n}\right)\nonumber\\ \leq  2\psi\left(c_0\left[\V\lor r^*\lor (\LLL-\LL_0)^2\right];\frac{\delta}{C_n}\right).
\end{align}
where $c_0$ is an absolute constant. This completes the proof of Theorem \ref{thm variance second stage}.
\hfill$\square$

\subsubsection{Analysis of the first-stage ERM estimator}\label{subsubsec first stage ERM}

After proving Theorem \ref{thm variance second stage}, the remaining part needed to prove Theorem \ref{thm variance rate} is to bound $(\LLL-\LL_0)^2$---the error of the first-stage ERM estimator.   

\paragraph{The remaining steps in the proof of Theorem \ref{thm variance rate}:}
We will give a guarantee on the first-stage ERM estimator, and combine this guarantee with Theorem \ref{thm variance second stage} to prove Theorem \ref{thm variance rate}. Recall that $\P_{S'}$ is the empirical distribution of the ``auxiliary" data set. Denote $\hh_{\ERM}\in\argmin_{\Hy}\P_{S'}\ell(h;z)$.

From Part \RNum{1} in the proof of Theorem \ref{thm variance second stage}, $\forall \delta\in(0,\frac{1}{2})$, with probability at least $1-\delta$,  
\begin{align*}
    \sup_{\F}|(\P-\Pn)f|\leq \psi(4B^2;\delta)\leq \psi\left(4B^2;\frac{\delta}{C_n}\right).
\end{align*}
 Since $\psi$ is sub-root with respect to its first argument, we have 
\begin{align*}
    \frac{\psi(4B^2;\frac{\delta}{C_n})}{\sqrt{4B^2}}\leq \frac{\psi(r^*;\frac{\delta}{C_n})}{\sqrt{r^*}}=\frac{\sqrt{r^*}}{16B},
\end{align*}
where $r^*$ is the fixed point of $16B\psi(r;\frac{\delta}{C_n})$. So we have proved that $\psi(4B^2;\frac{\delta}{C_n})\leq \frac{\sqrt{r^*}}{8}$. Therefore, 
\begin{align*}
    \sup_{\F}\left|(\P-\Pn)f\right|\leq \frac{\sqrt{r^*}}{8}.
\end{align*}

 Because $\hh_{\ERM}\in\argmin_{\Hy}\P_{S'}\ell(h;z)$ and $\P_{S'}\ell(\hh_{\ERM};z)=\LLL$, we have
\begin{align*}
    \LLL-\LL_0=(\P_{S'}\ell(\hh_{\ERM};z)-\P_{S'}\ell(h^*;z))+(\P_{S'}\ell(h^*;z)-\P\ell(h^*;z))\\\leq \P_{S'}\ell(h^*;z)-\P\ell(h^*;z)\leq \sup_{\F}|(\P-\Pn)f|,
\end{align*}
and 
\begin{align*}
    \LLL-\LL_0=(\P_{S'}\ell(\hh_{\ERM};z))-\P\ell(\hh_{\ERM};z))+(\P\ell(\hh_{\ERM};z)-\P\ell(h^*;z))\\\geq \P_{S'}\ell(\hh_{\ERM};z))-\P\ell(\hh_{\ERM};z)
    \geq -\sup_{\F}|(\P-\Pn)f|.
\end{align*}
Hence we have
\begin{align*}
    (\LLL-\LL_0)^2\leq (\sup_{\F}|(\P-\Pn)f|)^2 \leq \frac{r^*}{64}. 
\end{align*}
Combine this result with \eqref{eq: result of the first four parts}, we have with probability $1-2\delta$, 
\begin{align*}
     \mathcal{E}(\hh_{\MP})
    &\leq 2\psi\left(c_1\left(\V\lor r^*\right);\frac{\delta}{C_n}\right)\nonumber\\
    &\leq 2\left(\psi\left(c_1\V;\frac{\delta}{C_n}\right)\lor \psi\left(c_1 r^*;\frac{\delta}{C_n}\right)\right)\nonumber\\
    &\leq 2\psi\left(c_1\V;\frac{\delta}{C_n}\right)\lor \frac{c_1 r^*}{8B},
\end{align*} where $c_1=\max\{c_0,16\}$ is an absolute constant, and the last inequality follows from the fact that $\frac{c_1 r^*}{16}>r^*$ and the definition of fixed points. This completes the proof of Theorem \ref{thm variance rate}.
\hfill$\square$

\subsection{Estimating the variance-dependent rate from data}\label{subsec estimate variance rate}
In the remark following Theorem \ref{thm variance rate}, we comment that fully data-dependent variance-dependent bounds can be derived by employing an empirical estimate  to  the unknown parameter $\V$. Here we present the full details and some discussion of this approach.
\begin{theorem}[{\bf estimate of the variance-dependent rate from data}]\label{thm estimate variance rate}
 Consider the empirical centered second moment
\begin{align*}
    \widehat{\V}:=\Pn\left[\ell(\hh_{\textup{NMP}};z)-\widehat{\LL_0})^2\right],
\end{align*}
where $\LLL\in [-B,B]$ is the preliminary estimate of $\LL$ obtained in the first-stage, $\psi$ is defined in Strategy \ref{strategy estimator}, and
\begin{align*}
    \hh_{\textup{NMP}}\in\argmin_{\Hy}\Pn\ell(h;z)-2\psi\left(16\Pn\left[(\ell(h;z)-\widehat{\LL_0})^2\right]\right).
\end{align*}
For any fixed $\delta\in(0,1)$,  by performing the moment-penalized estimator in  Strategy \ref{strategy estimator}, with probability at least $1-\frac{\delta}{2}$, 
\begin{align}\label{eq: computable variance rate}
    \mathcal{E}(\hh_{\MP})\leq 4\psi\left(16\widehat{\V};\frac{\delta}{C_n}\right)\lor\frac{r^*}{8B},
\end{align}
where $r^*$ is the fixed point of $16B\psi(r;\frac{\delta}{C_n})$.
\end{theorem}
\paragraph{Remarks.} 1) The  subscript ``NMP" within $\hh_{\textup{NMP}}$ means ``negative moment penalization." Note that $\hh_{\textup{NMP}}$ may not have good generalization performance, it is only used to compute $\widehat{\V}$ so that  we can evaluate the estimator $\hh_{\MP}$ proposed in Strategy \ref{strategy estimator}. 

\noindent 2) While the fully data-dependent generalization error bound \eqref{eq: computable variance rate} provides a way to evaluate the moment-penalized estimator in  Strategy \ref{strategy estimator} from training data, it seems  that $\widehat{\V}$ and $\V$ are not necessarily of the same order. Therefore, \eqref{eq: computable variance rate}  may not be as tight as the original variance-dependent rate in Theorem \ref{thm variance rate}. One should view \eqref{eq: computable variance rate} as a relaxation of the original variance-dependent rate in Theorem \ref{thm variance rate}. 

\noindent 3) We also comment that the ``sub-root" assumption in  Theorem \ref{thm variance rate} is not needed here as we do not discuss the precision of $\widehat{\LL_0}$. It is easy to combine Theorem \ref{thm estimate variance rate} with the guarantee on $\widehat{\LL_0}$ proved in Appendix \ref{subsubsec first stage ERM}.

\paragraph{Proof of Theorem \ref{thm estimate variance rate}:}
define $\hat{f}_{\text{NMP}}$ by $\fh_{\text{NMP}}(z)=\ell(\hh_{\text{NMP}};z)-\ell(h^*;z),\forall z\in\Z$, and $w(h;z)=\ell(h;z)-\widehat{\LL_0}$. From the
the results \eqref{eq: before regularization variance rate} \eqref{eq: second stage rate empirical moment} \eqref{eq: specification delta1} in the proof of Theorem \ref{thm variance second stage}, we have  with probability at least $1-\frac{\delta}{2}$, 
\begin{align}\label{eq: coro variance first}
    (\P-\Pn)f\leq \psi\left(4\Pn[f^2]\lor r^*;\frac{\delta}{C_n}\right),\quad \forall f\in\F
\end{align}
and 
\begin{align}\label{eq: coro variance second}
     \mathcal{E}(\hh_{\MP})\leq 2\psi\left(16\Pn[w(h^*;z)^2]\lor r^*;\frac{\delta}{C_n}\right).
\end{align}

From the definition of $\hh_{\textup{NMP}}$,
\begin{align}\label{eq: property NMP}
    \Pn\ell(\hh_{\text{NMP}};z)-2\psi\left(16\Pn[w(\hh_{\text{NMP}};z)^2];\frac{\delta}{C_n}\right)\leq \Pn \ell(h^*;z)-2\psi\left(16\Pn[w(h^*;z)^2];\frac{\delta}{C_n}\right).
\end{align}
Therefore, with probability at least $1-\frac{\delta}{2}$, we have
\begin{align}\label{eq: coro variance long}
    &2\psi\left(16\Pn[w(h^*;z)^2];\frac{\delta}{C_n}\right)\nonumber\\
    &\leq  2\psi\left(16\Pn[w(\hh_{\text{NMP}};z)^2];\frac{\delta}{C_n}\right)+\Pn\ell(h^*;z)-\Pn\ell(\hh_{\text{NMP}};z)
   \nonumber \\&=  2\psi\left(16\Pn[w(\hh_{\text{NMP}};z)^2];\frac{\delta}{C_n}\right)+\P[\ell(h^*;z)-\ell(\hh_{\text{NMP}};z)]+(\Pn-\P)[\ell(h^*;z)-\ell(\hh_{\text{NMP}};z)]\nonumber\\
    &\leq 2\psi\left(16\Pn[w(\hh_{\text{NMP}};z)^2];\frac{\delta}{C_n}\right)+(\P-\Pn)\hat{f}_{\text{NMP}}\nonumber\\
    &\leq 2\psi\left(16\Pn[w(\hh_{\text{NMP}};z)^2];\frac{\delta}{C_n}\right)+\psi\left(4\Pn[\hat{f}_{\text{NMP}}^2];\frac{\delta}{C_n}\right),
\end{align}
where the first inequality is due to \eqref{eq: property NMP}, the second inequality is due to the fact that $h^*$ minimizes the population risk; and the last inequality is due to \eqref{eq: coro variance first}.

Note that 
\begin{align*}
    4\Pn[\hat{f}_{\text{NMP}}^2]\leq 8\Pn[w({\hh_{\text{NMP}}};z)^2]+8\Pn[w(h^*;z)^2]\\
    \leq 16\Pn[w(\hh_{\text{NMP}};z)^2]\lor 16\Pn[w(h^*;z)^2].
\end{align*}
From the above inequality and \eqref{eq: coro variance long}, with probability at least $1-\frac{\delta}{2}$, we have
\begin{align}\label{eq: coro variance imply the later}
     &2\psi\left(16\Pn[w(h^*;z)^2];\frac{\delta}{C_n}\right)\nonumber\\&\leq \  2\psi\left(16\Pn[w(\hh_{\text{NMP}};z)^2];\frac{\delta}{C_n}\right)+\psi\left(16\Pn[w(\hat{h}_{\text{NMP}};z)^2];\frac{\delta}{C_n}\right)\lor \psi\left(16\Pn[w(h^*;z)^2];\frac{\delta}{C_n}\right).
\end{align}
Whether $\Pn[w(h^*;z)^2]\leq 16\Pn[w(\hh_{\text{NMP}};z)^2]$ or $\Pn[w(h^*;z)^2]> 16\Pn[w(\hh_{\text{NMP}};z)^2]$, the inequality \eqref{eq: coro variance imply the later} always implies
\begin{align}\label{eq: estimate variance before final}
    \psi\left(16\Pn[w(h^*;z)^2];\frac{\delta}{C_n}\right)\leq 2\psi\left(16\Pn[w(\hh_{\text{NMP}};z)^2];\frac{\delta}{C_n}\right)
    =2\psi\left(16\widehat{\V};\frac{\delta}{C_n}\right).
\end{align}
(Note that $\widehat{\V}:=\Pn[w(\hh_{\text{NMP}};z)^2]$.)
We conclude that with probability at least $1-\frac{\delta}{2}$,
\begin{align*}
    \mathcal{E}(\hh_{\MP})&\leq 2\psi\left(16\Pn[w(h^*;z)^2]\lor r^*;\frac{\delta}{C_n}\right)\\
    &=2\psi\left(16\Pn[w(h^*;z)^2];\frac{\delta}{C_n})\lor 2\psi(r^*;\frac{\delta}{C_n}\right)\\
    &\leq 4\psi\left(16\widehat{\V};\frac{\delta}{C_n}\right)\lor\frac{r^*}{8B},
\end{align*}
where the first inequality is due to \eqref{eq: coro variance second} and the last inequality is due to \eqref{eq: estimate variance before final}. 
This completes the proof.
\hfill$\square$
\subsection{Auxiliary lemmata}\label{appendix auxiliary lemma}

\begin{lemma}[{\bf Talagrand's concentration inequality for empirical processes, \cite{bartlett2005local}}]\label{lemma bartlett}
Let $\F$ be a class of functions that map $\pazocal{Z}$ into $[B_1,B_2]$. Assume that there is some $r > 0$ such that for every $f \in \F$, $\textup{Var}[f(z_i)]\leq r$. Then, for every $\delta \in(0,1)$, with probability at least $1-\delta$,
\begin{align*}
    \sup_{f\in\F}(\P-\Pn)f\leq  3\mathfrak{R}\F +\sqrt{\frac{2r\log\frac{1}{\delta}}{n}}+(B_2-B_1)\frac{\log\frac{1}{\delta}}{n},
\end{align*}
and with probability at least $1-\delta$,  \begin{align*}
    \sup_{f\in\F}(\P-\Pn)f\leq  4\mathfrak{R}_n\F +\sqrt{\frac{2r\log\frac{2}{\delta}}{n}}+\frac{9}{2}(B_2-B_1)\frac{\log\frac{2}{\delta}}{n}.
\end{align*}
Moreover, the same results hold for the quantity $sup_{f\in\F}(\Pn-\P)f$.
\end{lemma}

\begin{lemma}[{\bf Bernstein’s inequality, \cite{dirksen2015tail}}]\label{lemma Bernstein} Let $X_1,\cdots,X_n$ be real-valued, independent,
mean-zero random variables and suppose that for some constants $\sigma, B>0$,
\begin{align*}
\frac{1}{n}\sum_{i=1}^n\E|X_i|^k\leq\frac{k!}{2}\sigma^2B^{k-2}, \quad k=2,3,\cdots
\end{align*}
Then, $\forall \delta\in(0,1)$, with probability at least $1-\delta$
\begin{align}\label{eq: bernstein tail bound}
    \left|\frac{1}{n}\sum_{i=1}^n X_i\right|\leq \sqrt{\frac{2\sigma^2\log\frac{2}{\delta}}{n}}+\frac{B\log\frac{2}{\delta}}{n}.
\end{align}
\end{lemma}

\section{Proofs for Section \ref{sec parametric}, Section \ref{sec application fast rate regime} and Section \ref{sec em}}

\subsection{Proof of Lemma \ref{lemma hessian}}
 Fix $u\in\pazocal{B}^d(0,1)$ and $\theta_1,\theta_2\in\Theta$, then  we have
\begin{align*}
   u^T(\nabla\ell(\theta_1;z)-\nabla \ell(\theta_2;z))^T =\int_0^1u^T[\nabla^2\ell(\theta_2+v(\theta_1-\theta_2);z)](\theta_1-\theta_2) dv.
\end{align*}
By Jensen's inequality,
\begin{align*}
    \exp\left(\frac{u^T(\nabla\ell(\theta_1;z)-\nabla\ell(\theta_2;z))}{\beta\|\theta_1-\theta_2\|}\right)
    =\exp\left(\int_0^1u^T[\nabla^2\ell(\theta_2+v(\theta_1-\theta_2);z)]\frac{(\theta_1-\theta_2)}{\|\theta_1-\theta_2\|} dv\right)\nonumber\\
    \leq \int_0^1\exp\left(u^T[\nabla^2\ell(\theta_2+v(\theta_1-\theta_2);z)]\frac{(\theta_1-\theta_2)}{\|\theta_1-\theta_2\|}\right)dv.
\end{align*}
 It is then straightforward to prove the lemma by taking expectation with respect to $z$ in the above inequality and using the condition \eqref{eq: hessian 2}.
 \hfill$\square$
\subsection{Proof of Proposition \ref{prop localized gradient 1}}
 \label{subsec appendix thm parametric}

Take $V=\{v\in\R^d: \|v\|\leq \max\{\Delta_M,\frac{1}{n}\}\}$. We will first prove a  ``uniform localized convergence" argument over all $\theta\in\Theta$ and $v\in V$.
 
 \begin{proposition}[{\bf directional ``uniform localized  convergence" of gradient}]\label{prop localized gradient concentration}   Under Assumption \ref{asm hessian noise}, $ \forall\delta\in(0,1)$,  with probability at least $1-\delta$,  for all $ \theta\in \Theta, v\in V$, either $\|\theta-\theta^*\|^2+\|v\|^2\leq \frac{2}{n^2}$, or
\begin{align*}
&(\P-\Pn) \left[(\nabla \ell(\theta;z)-\nabla \ell(\theta^*;z))^T v\right]\\
   &\leq  c_1 \beta\max\left\{\|\theta-\theta^*\|^2+\|v\|^2,\frac{2}{n^2}\right\}\left( \sqrt{\frac{d+\log\frac{2\log_2 (2n^2\Delta_M^2+2)}{\delta}}{n}}+\frac{d+\log\frac{2\log_2 (2n^2\Delta_M^2+2)\}}{\delta}}{n}\right),
\end{align*}
where $c_1$ is an absolute constant.
 \end{proposition}

\paragraph{Proof of Proposition \ref{prop localized gradient concentration}:} 
for $(\theta, v)\in \Theta\times V$, let $g_{(\theta,v)}=(\nabla \ell(\theta;z)-\nabla \ell(\theta^*;z))^T v$. For $(\theta_1,v_1)$ and $(\theta_2,v_2)\in\Theta\times v$, define the norm on the product space $\Theta\times V$ as \begin{align}\label{eq: norm product space}
\|(\theta_1,v_1)-(\theta_2,v_2)\|_{\pro}=\sqrt{\|\theta_1-\theta_2\|^2+\|v_1-v_2\|^2}.
\end{align}
Denote $\pazocal{B}(\sqrt{r}):=\{(\theta,v)\in\Theta\times V: \|\theta-\theta^*\|^2+\|v\|^2\leq r\}$. Given $(\theta_1,v_1),(\theta_2,v_2)\in\pazocal{B}(\sqrt{r})$, we perform the following re-arrangement and decomposition steps:
\begin{align}\label{eq: interchange decomposition}
    &g_{(\theta_1,v_1)}(z)-g_{(\theta_2,v_2)}(z)\nonumber\\
    &=(\nabla \ell(\theta_1;z)-\nabla \ell(\theta^*;z))^T v_1-(\nabla \ell(\theta_2;z)-\nabla \ell(\theta^*;z))^T v_2\nonumber\\
    &=(\nabla\ell(\theta_1;z)-\nabla \ell(\theta^*;z))^T(v_1-v_2)+(\nabla\ell(\theta_1;z)-\nabla\ell(\theta^*;z))^Tv_2+(\nabla \ell(\theta^*;z)-\nabla\ell(\theta_2;z))^T v_2\nonumber\\
    &=(\nabla\ell(\theta_1;z)-\nabla \ell(\theta^*;z))^T(v_1-v_2)+(\nabla \ell(\theta_1;z)-\nabla\ell(\theta_2;z))^Tv_2
    \end{align}
When $(\theta_1,v_1), 
(\theta_2,v_2)\in\pazocal{B}(\sqrt{r})$, we have
\begin{align*}
    \|\theta_1-\theta^*\|\|v_1-v_2\|\leq \sqrt{r}\|v_1-v_2\|\leq \sqrt{r}\|(\theta_1,v_1)-(\theta_2,v_2)\|_{\pro},
\end{align*}
so from Assumption \ref{asm hessian noise}, $(\nabla\ell(\theta_1;z)-\nabla \ell(\theta^*;z))^T(v_1-v_2)$ is $\beta\sqrt{r}\|(\theta_1,v_1)-(\theta_2,v_2)\|_{\pro}-$sub-exponential. 
Similarly, we can prove $(\nabla \ell(\theta_1;z)-\nabla\ell(\theta_2;z))^T v_2$ to be $\beta\sqrt{r}\|(\theta_1,v_1)-(\theta_2,v_2)\|_{\pro}-$sub-exponential.
From the decomposition \eqref{eq: interchange decomposition}  and Jensen's inequality, for all $(\theta_1,v_1), 
(\theta_2,v_2)\in\pazocal{B}(\sqrt{r})$, we have
\begin{align*}
    &\exp\left(\frac{g_{(\theta_1,v_1)}(z)-g_{(\theta_2,v_2)}(z)}{2\beta\sqrt{r}\|(\theta_1,v_1)-(\theta_2,v_2)\|_{\pro}}\right)\\
    &\leq\frac{1}{2}\exp\left(\frac{(\nabla\ell(\theta_1;z)-\nabla \ell(\theta^*;z))^T(v_1-v_2)}{\beta\sqrt{r}\|(\theta_1,v_1)-(\theta_2,v_2)\|_{\pro}}\right)+\frac{1}{2}\exp\left(\frac{(\nabla \ell(\theta_1;z)-\nabla\ell(\theta_2;z))^T v_2}{\beta\sqrt{r}\|(\theta_1,v_1)-(\theta_2,v_2)\|_{\pro}}\right).
\end{align*}
By taking expectation with respect to $z$ in the above inequality, we prove that $g_{(\theta_1,v_1)}(z)-g_{(\theta_2,v_2)}(z)$ is a $2\beta\sqrt{r}\|(\theta_1,v_1)-(\theta_2,v_2)\|_{\pro}-$sub-exponential random variable, i.e.,
\begin{align*}
    \|g_{(\theta_1,v_1)}(z)-g_{(\theta_2,v_2)}(z)\|_{\text{Orlicz}_1}\leq 2\beta\sqrt{r}\|(\theta_1,v_1)-(\theta_2,v_2)\|_{\pro}.
\end{align*}

From Bernstein inequality for sub-exponential variables (Lemma \ref{lemma Bernstein orlicz}), for any fixed $u\geq 0$ and $(\theta_1,v_1), (\theta_2,v_2)\in\Theta\times V$,
\begin{align*}
    \text{Prob}\bigg\{|(\P-\Pn)[g_{(\theta_1,v_1)}(z)-g_{(\theta_2,v_2)}(z)]\geq2\beta\sqrt{r}\|(\theta_1,v_1)-(\theta_2,v_2)\|_{\pro}\sqrt{\frac{2}{n}}\sqrt{u}&\\ +\frac{2\beta\sqrt{r}\|(\theta_1,v_1)-(\theta_2,v_2)\|_{\pro}}{n}u&\bigg\}\leq 2\exp(-u).
\end{align*}
The above inequality implies that the empirical process $(\P-\Pn)g_{(\theta,v)}$ has a mixed sub-Gaussian-sub-exponential increments with respect to the metrics $(\frac{2\beta\sqrt{r}}{n}\|\cdot\|_{\pro},\frac{2\sqrt{2}\beta\sqrt{r}}{\sqrt{n}}\|\cdot\|_{\pro} )$ (see Definition \ref{def mixed tail}). 

 From Lemma \ref{lemma mixed tail}, there exists an absolute constants $C$ such that $\forall \delta\in(0,1)$,  with probability at least $1-\delta$,
\begin{align*}
    \sup_{\|\theta-\theta^*\|^2+\|v\|^2\leq r} (\P-\Pn) g_{(\theta,v)} \leq C\left(\gamma_2\Bigg(\pazocal{B}(\sqrt{r}),\frac{2\sqrt{2}\beta\sqrt{r}}{\sqrt{n}}\|\cdot\|_{\pro}\right) +\gamma_1\left(\pazocal{B}(\sqrt{r}),\frac{2\beta\sqrt{r}}{n}\|\cdot\|_{\pro}\right)\\
    +\beta r\sqrt{\frac{\log\frac{1}{\delta}}{n}}+\beta r\frac{\log\frac{1}{\delta}}{n}\Bigg).
\end{align*}
 Using Dudley's integral (Lemma \ref{lemma dudley}) to bound the $\gamma_1$ functional and the $\gamma_2$ functional, we obtain that there exist absolute constant $c_1$  such that $\forall \delta\in(0,1)$,  with probability at least $1-\delta$,
\begin{align}\label{eq: localized convergence fthetav}
    \sup_{\|(\theta,\theta')-(\theta^*,\theta^*)\|^2\leq r} \left|(\P-\Pn) g_{(\theta,v)}\right| 
    \leq c_1 \beta r\left( \sqrt{\frac{d+\log\frac{1}{\delta}}{n}}+\frac{d+\log\frac{1}{\delta}}{n}\right).
\end{align}

We set
\begin{align*}
    \psi(r;\delta)=c_1 \beta r\left( \sqrt{\frac{d+\log\frac{1}{\delta}}{n}}+\frac{d+\log\frac{1}{\delta}}{n}\right).
\end{align*} 
Denote $R=2(\Delta_M^2+\frac{1}{n^2})$ and $r_0=\frac{2}{n^2}$. Since $V$ is a $d-$dimensional ball centered at the origin with radius $\max\{\Delta_M,\frac{1}{n}\}$, we know that $\|\theta-\theta^*\|^2+\|v\|^2\leq 2\Delta_M^2+\frac{1}{n^2}\leq R$.  We apply Proposition \ref{prop variant peeling} and obtain: for any fixed $\delta\in(0,1)$, with probability at least $1-\delta$, for all $\theta\in\Theta$ and $v\in V$, 
\begin{align*}
   &(\P-\Pn) \left[(\nabla \ell(\theta;z)-\nabla \ell(\theta^*;z))^Tv\right]=(\P-\Pn) g_{(\theta,v)}\\
   &\leq  \psi\left( \max\left\{\|\theta-\theta^*\|^2+\|v\|^2,\frac{2}{n^2}\right\}; \frac{\delta}{2\log_2({2R}/{\frac{2}{n^2}})} \right)\nonumber\\
    &=  c_1 \beta \max\left\{\|\theta-\theta^*\|^2+\|v\|^2,\frac{2}{n^2}\right\}\left( \sqrt{\frac{d+\log\frac{2\log_2 (n^2R)}{\delta}}{n}}+\frac{d+\log\frac{{2\log_2 (n^2R)}}{\delta}}{n}\right).
\end{align*}
This completes the proof of Proposition \ref{prop localized gradient concentration}.
\hfill$\square$

\paragraph{Proof of Proposition \ref{prop localized gradient 1}:} 

in order to uniformly bound  $\|(\P-\Pn) (\nabla \ell(\theta;z)-\nabla \ell(\theta^*;z))\|$ for all  $\theta\in\Theta$, we  take $$v=\max\left\{\|\theta-\theta^*\|,\frac{1}{n}\right\}\cdot\frac{(\P-\Pn)(\nabla\ell(\theta;z)-\nabla\ell(\theta^*;z))}{\|(\P-\Pn)(\nabla\ell(\theta;z)-\nabla\ell(\theta^*;z))\|}$$ in Proposition \ref{prop localized gradient concentration}. Clearly $\|v\|=\max\{\|\theta-\theta^*\|,\frac{1}{n}\}$. From Proposition \ref{prop localized gradient 1}, we can prove that there exists an absolute constant $c$  such that $ \forall\delta\in(0,1)$, with probability at least $1-\delta$, for all $ \theta\in \Theta$, 
\begin{align*}
&\|(\P-\Pn) (\nabla \ell(\theta;z)-\nabla \ell(\theta^*;z))\|\\&\leq
     c \beta\max\left\{\|\theta-\theta^*\|,\frac{1}{n}\right\}\left( \sqrt{\frac{d+\log\frac{2\log_2 (2n^2\Delta_M^2+2)}{\delta}}{n}}+\frac{d+\log\frac{2\log_2 (2n^2\Delta_M^2+2)}{\delta}}{n}\right)\\
     &\leq 
     c \beta\max\left\{\|\theta-\theta^*\|,\frac{1}{n}\right\}\left( \sqrt{\frac{d+\log\frac{4\log_2 (2n\Delta_M+2)}{\delta}}{n}}+\frac{d+\log\frac{4\log_2 (2n\Delta_M+2)}{\delta}}{n}\right).
\end{align*}
This completes the proof of Proposition \ref{prop localized gradient 1}.

\hfill$\square$

\subsection{Proof of Theorem \ref{thm pl}}

We first prove a proposition on the uniform localized convergence of gradients under Assumption \ref{asm hessian noise} and Assumption \ref{asm bernstein condition para}.

\begin{proposition}[{\bf uniform localized convergence of gradients}]\label{prop localized convergence assumption both}
Let Assumption \ref{asm hessian noise}, Assumption \ref{asm bernstein condition para} hold along with the optimality condition $\P\nabla\ell(\theta^*;z)=0$. Given $\delta\in(0,1)$,  denote
\begin{align*}
    \textup{term \RNum{1}}&:=\sqrt{\frac{2\P[\|\nabla\ell(\theta^*;z)\|^2]\log\frac{4}{\delta}}{n}}+\frac{G_{*}\log\frac{4}{\delta}}{n},
    \\
    \textup{term \RNum{2}}&:=\sqrt{\frac{d+\log\frac{8\log_2(2n\Delta_M+2)}{\delta}}{n}}+\frac{d+\log\frac{\log_2(2n\Delta_M+2)}{\delta}}{n}.
\end{align*}
Then with probability at least $1-\delta$, we have the following:
\begin{align}\label{eq: localized gradient both result 1}
    \|(\Pn-\P)\nabla\ell(\theta^*;z)\|\leq \textup{term \RNum{1}},
\end{align}
and 
\begin{align}\label{eq: localized gradient both result 2}
     \|(\Pn-\P) \nabla \ell(\theta;z)\|\leq  {\textup{term \RNum{1}}}+c_0\beta\max\left\{\frac{1}{n},\|\theta-\theta^*\|\right\}\cdot\textup{term \RNum{2}}
    ,\quad \forall \theta\in\Theta,
\end{align}
where  $c_0$ is an absolute constant.
\end{proposition}

\paragraph{Proof of Proposition \ref{prop localized convergence assumption both}:}
from Proposition \ref{prop localized gradient 1}, 
there exists an absolute constant $c_0$  such that $ \forall\delta_1>0$, with probability at least $1-\frac{\delta}{2}$, for all $\theta\in\Theta$,
\begin{align}
&\|(\Pn-\P) \nabla \ell(\theta;z)\|\nonumber\\
&\leq \|(\Pn-\P)\nabla\ell(\theta^*;z)\|+
     c_0 \beta\max\left\{\|\theta-\theta^*\|,\frac{1}{n}\right\}\cdot \textup{term \RNum{2}}
  \label{eq: localized gradient both 1}.
\end{align}
From  Bernstein's inequality for vectors (Lemma \ref{lemma vector Bernstein}), we have  with probability at least $1-\frac{\delta}{2}$,
\begin{align}\label{eq: localized gradient both 2}
   \|\P\nabla\ell(\theta^*;z)-\Pn\nabla \ell(\theta^*;z)\|\leq \sqrt{\frac{2\P[\|\nabla\ell(\theta^*;z)\|^2]\log\frac{4}{\delta}}{n}}+\frac{G_{*}\log\frac{4}{\delta}}{n}=\textup{term \RNum{1}},
\end{align}
Combining \eqref{eq: localized gradient both 1} and \eqref{eq: localized gradient both 2} by a union bound, we complete the proof of  Proposition \ref{prop localized convergence assumption both}.
\hfill$\square$

\paragraph{}

Let $F$ be $\beta-$smooth, i.e., for all $\theta_1,\theta_2\in\Theta$,
\begin{align*}
    \|\nabla F(\theta_1)-\nabla F(\theta_2)\|\leq \beta\|\theta_1-\theta_2\|.
\end{align*}
 Under this condition and the PL condition, the population risk minimizer $\theta^*$ will be the unique minimizer of $F$ on $\Theta$. We first present the following lemma.

\begin{lemma}
[{\bf relationship between curvature conditions}]\label{lemma curvature conditions} For a $\beta-$smooth function $F$, consider the following conditions:

1. Strong convexity (SC): for all $\theta_1,\theta_2\in\Theta$ we have
\begin{align*}
    F(\theta_1)\geq F(\theta_2)+\nabla F(\theta_2)^T(\theta_1-\theta_2)+\frac{\mu}{2}\|\theta_1-\theta_2\|^2.
\end{align*}

2. Polyak-Lojasiewisz (PL): for all $\theta\in\Theta$ we have 
\begin{align*}
    F(\theta)-F(\theta^*)\leq \frac{1}{2\mu}\|\nabla F(\theta)\|^2.
\end{align*}

3. Error Bound (EB): for all $\theta\in\Theta$ we have
\begin{align*}
    \|\nabla F(\theta)\|\geq \mu\|\theta-\theta^*\|.
\end{align*}

4. Quadratic Growth (QG): for all $\theta\in\Theta$ we have
\begin{align*}
    F(\theta)-F(\theta^*)\geq \frac{\mu}{2}\|\theta-\theta^*\|^2.
\end{align*}
Then, the following hold:
\begin{center}
    (SC)$\implies$(PL)$\implies$(EB)$\implies$(QG),\\
    (EB) with parameter $\mu$ $\implies$ (PL) with parameter $\frac{\mu^2}{\beta}$.
\end{center}
\end{lemma}

Proof of Lemma \ref{lemma curvature conditions} can be adapted from [\cite{karimi2016linear}, Appendix A]. Note that some parameters in the original statements in \cite{karimi2016linear} have typos though the proof ideas are correct. In Lemma \ref{lemma curvature conditions} we fix those typos on the parameters.
As argued in \cite{karimi2016linear}, (PL) and the equivalent (QG) (under the smoothness condition and change of parameters) are the most general conditions that allow linear convergence to a global minimizer.
\paragraph{}
We now prove Theorem \ref{thm pl}.
\paragraph{Proof of Theorem \ref{thm pl}:}
we prove the results on the event $$\mathcal{A}:=\{\text{the results \eqref{eq: localized gradient both result 1} \eqref{eq: localized gradient both result 2} in  Proposition \ref{prop localized convergence assumption both} hold true}\},$$ whose measure is at least $1-\delta$.  We keep the notations ``term \RNum{1}" and ``term \RNum{2}" used in Proposition \ref{prop localized convergence assumption both}, which are defined by
\begin{align*}
    \textup{term \RNum{1}}:=\sqrt{\frac{2\P[\|\nabla\ell(\theta^*;z)\|^2]\log\frac{4}{\delta}}{n}}+\frac{G_{*}\log\frac{4}{\delta}}{n},
    \\
    \textup{term \RNum{2}}:=\sqrt{\frac{d+\log\frac{8\log_2(2n\Delta_M+2)}{\delta}}{n}}+\frac{d+\log\frac{\log_2(2n\Delta_M+2)}{\delta}}{n}.
\end{align*}

The PL condition (Assumption \ref{asm pl}) implies that $\P\nabla\ell(\theta^*;z)=0$.
From the result \eqref{eq: localized gradient both result 1} in Proposition \ref{prop localized convergence assumption both},
\begin{align*}
   \|\Pn\nabla\ell(\theta^*;z)\|= \|(\Pn-\P)\nabla\ell(\theta^*;z)\|\leq \textup{term \RNum{1}}.
\end{align*} So we know that the equation
\begin{align}\label{eq: stationary point appendix}
\|\Pn \nabla\ell(\theta;z)\|\leq \textup{term \RNum{1}}.
\end{align} 
must have a solution within $\Theta$.

The result \eqref{eq: localized gradient both result 2} implies that for all $\theta\in\Theta$ such that $\|\theta-\theta^*\|\leq \frac{1}{n}$,
\begin{align*}
     \|\P \nabla \ell(\theta;z)\|\leq  \|\Pn\nabla\ell(\theta;z)\|+{\textup{term \RNum{1}}}
    +c_0\beta\|\theta-\theta^*\|\cdot{\textup{term \RNum{2}}}.
\end{align*}
Since the PL condition implies (see Lemma \ref{lemma curvature conditions})
\begin{align*}
    \|\P\nabla\ell(\theta;z)\|\geq \mu\|\theta-\theta^*\|,
\end{align*}
for all $\theta\in\Theta$ such that $\|\theta-\theta^*\|\leq \frac{1}{n}$, we have 
\begin{align*}
    \mu\|\theta-\theta^*\|\leq \|\P\nabla \ell(\theta;z)\|\leq \|\Pn\nabla\ell(\theta;z)\|+ {\textup{term \RNum{1}}}
    +c_0\beta\|\theta-\theta^*\|\cdot{\textup{term \RNum{2}}},
\end{align*}
where $c$ is an absolute constant.
Therefore, for all $\theta\in\Theta$, we must have
\begin{align}\label{eq: uniform localized thm pl}
    \mu\|\theta-\theta^*\|\leq \|\P\nabla \ell(\theta;z)\|\leq \|\Pn\nabla\ell(\theta;z)\|+ {\textup{term \RNum{1}}}
    +c_0\beta\|\theta-\theta^*\|\cdot{\textup{term \RNum{2}}}+\frac{\mu}{n}.
\end{align}

Let $\hth\in\Theta$ be an arbitrary solution that satisfies \eqref{eq: stationary point appendix}.
From \eqref{eq: uniform localized thm pl}, we obtain the inequalities for $\|\hth-\theta^*\|$: 
\begin{align}\label{eq: temp nonconvex proof 1}
  \mu \|\hth-\theta^*\|\leq  \|\P \nabla \ell(\hth;z)\|\leq   2\cdot{\textup{term \RNum{1}}}
    +c_0\beta\cdot{\textup{term \RNum{2}}}\cdot \|\theta-\theta^*\|+\frac{\mu}{n}.
\end{align}
Let $c=\max\{4c_0^2,1\}$.  When 
\begin{align*}
    n\geq \frac{c\beta^2(d+\log\frac{4\log (2n\Delta_M+1)}{\delta})}{\mu^2},
\end{align*}
we have $c_0\beta\cdot \textup{term \RNum{2}}\leq \frac{\mu}{2}$ so that from \eqref{eq: temp nonconvex proof 1},
\begin{align*}
    \|\hth-\theta^*\|\leq \frac{2}{\mu}(2\cdot \text{term \RNum{1}}+\frac{\mu}{n})
\end{align*}
and the event $\mathcal{A}$. Plugging in ``$c_0\beta\cdot \textup{term \RNum{2}}\leq \frac{\mu}{2}$" and ``$ \|\hth-\theta^*\|\leq \frac{2}{\mu}(2\cdot \text{term \RNum{1}}+\frac{\mu}{n})$" into the second inequality within \eqref{eq: temp nonconvex proof 1}, we further have 
\begin{align*}
    \|\P\nabla\ell(\hth;z)\|
    \leq  2\cdot\textup{term \RNum{1}}+\frac{\mu}{n}
    +\frac{\mu}{2}\|\hth-\theta^*\|\\
    \leq 4\cdot\textup{term \RNum{1}}+\frac{2\mu }{n}.
\end{align*}
Lastly, since the PL condition implies (see Lemma \ref{lemma curvature conditions})
\begin{align*}
    \P\ell(\hth;z)-\P\ell(h^*;z)\leq\frac{\|\P\ell(\hth;z)\|^2}{2\mu},
\end{align*}
by plugging in ``$\|\P\nabla\ell(\hth;z)\|\leq 4\cdot\textup{term \RNum{1}}+\frac{2\mu }{n}$" we have
\begin{align*}
   \P\ell(\hth;z)-\P\ell(h^*;z)&\leq\frac{\|\P\ell(\hth;z)\|^2}{2\mu}\\
   &\leq \frac{16}{\mu}(\text{term \RNum{1}})^2+\frac{4\mu}{n^2}\\&\leq \frac{64\P[\|\nabla \ell(\theta^*;z)\|^2]\log\frac{4}{\delta}}{\mu n}+\frac{32 G_{*}^2\log^2\frac{4}{\delta}+4\mu^2}{\mu n^2}.
\end{align*}
This completes the proof of Theorem \ref{thm pl}.
\hfill$\square$

\subsection{Proof of Theorem \ref{thm pl algorithm}}\label{subsec proof fast algorithm}

We first prove a simple proposition, which studies how the accumulation of sample approximation errors at every step influences the convergence of the algorithm. 
\begin{proposition}[{\bf localized statistical error of a linearly convergent iterative algorithm}]\label{prop iterative}
Consider a function $F$ (for which we call the ``Lyapunov function") and a parameter $\gamma\in(0,1)$. Assume an algorithm satisfies  for all $t=0,1,\dots$
\begin{align*}
   F( \theta^{t+1})\leq (1-\gamma)F(\theta^t)+\eps^t(n),\\
   \eps^t(n)\leq \alpha(n) F(\theta^t)+\eps^*(n),\\
  \text{and}\quad \theta^t\in\Theta.
\end{align*}
When the sample size $n$ is large enough such that $\alpha(n)\leq \frac{\gamma}{2}$, we have
\begin{align*}
   F(\theta^t)\leq \left(1-\frac{\gamma}{2}\right)^t F(\theta^0)+\frac{2}{\gamma}\eps^*(n), \quad t=0,1,\cdots.
\end{align*}
 \end{proposition}

\paragraph{Proof of Proposition \ref{prop iterative}:}
we have
\begin{align*}
    F(\theta^{t+1})\leq (1-\gamma+\alpha(n))F(\theta^t)+\eps^*(n)\\
    \leq \left(1-\frac{\gamma}{2}\right)F(\theta^*)+\eps^*(n).
\end{align*}
Then by induction we have
\begin{align*}
    F(\theta^t)\leq \left(1-\frac{\gamma}{2}\right)^t F(\theta^0)+\frac{2}{\gamma}\eps^*(n), \quad t=0,1,\cdots.
\end{align*}
This completes the proof of Proposition \ref{prop iterative}.
\hfill$\square$

\paragraph{}

We now prove Theorem \ref{thm pl algorithm}.
\paragraph{Proof of Theorem \ref{thm pl algorithm}:} 
 Assumption \ref{asm hessian noise} implies that the population risk is $\beta-$smooth. Consider the gradient descent algorithm \eqref{eq: gradient descent scheme} with fixed step size $\frac{1}{\beta}$. We have for all $t=0,1,\cdots$,
\begin{align*}
    \theta^{t+1}=\theta^t-\frac{1}{\beta}\Pn\nabla\ell(\theta^t;z).
\end{align*}
 So we have
\begin{align*}
   \P\ell(\theta^{t+1};z)- \P\ell(\theta^t;z) \leq (\P\nabla \ell(\theta^t;z))^T(\theta^{t+1}-\theta^t)+\frac{\beta}{2}\|\theta^{t+1}-\theta^t\|^2\\
   =-\frac{1}{\beta}(\P\nabla\ell(\theta^t;z))^T(\Pn\nabla\ell(\theta^t;z))+\frac{1}{2\beta}\|\Pn\nabla\ell(\theta^t;z)\|^2\\
  = -\frac{1}{\beta}\|\P\nabla\ell(\theta^t;z)\|^2-\frac{1}{\beta}(\P\nabla\ell(\theta^t;z))^T(\Pn\nabla\ell(\theta^t;z)-\P\nabla\ell(\theta^t;z))\\
  +\frac{1}{2\beta}\|\P\nabla\ell(\theta^t;z)+(\Pn\nabla\ell(\theta^t;z)-\P\nabla\ell(\theta^t;z))\|^2\\
  =-\frac{1}{2\beta}\|\P\nabla\ell(\theta^t;z)\|^2+\frac{1}{2\beta}\|\Pn\nabla\ell(\theta^t;z)-\P\nabla\ell(\theta^t;z)\|^2\\
  \leq -\frac{\mu}{\beta}(\P\ell(\theta^t;z)-\P\ell(\theta^*;z))+\frac{1}{2\beta}\|\Pn\nabla\ell(\theta^t;z)-\P\nabla\ell(\theta^t;z)\|^2.
\end{align*}
Rearranging the above inequality, and subtracting $\P\ell(\theta^*;z)$ from both sides, we obtain 
\begin{align}\label{eq: proof pl algorithm stage 1}
    \P\ell(\theta^{t+1};z)-\P\ell(\theta^*;z)\leq \left(1-\frac{\mu}{\beta}\right)(\P\ell(\theta^t;z)-\P\ell(\theta^*;z))+\frac{1}{2\beta}\|\Pn\nabla\ell(\theta^t;z)-\P\nabla\ell(\theta^t;z)\|^2.
\end{align}

 Applying Proposition \ref{prop localized convergence assumption both}, we continue the proof on the event $$\mathcal{A}:=\{\text{the results \eqref{eq: localized gradient both result 1} \eqref{eq: localized gradient both result 2} in  Proposition \ref{prop localized convergence assumption both} hold true}\},$$  whose measure is at least $1-\delta$. We keep the notations ``term \RNum{1}" and ``term \RNum{2}" used in Proposition \ref{prop localized convergence assumption both}, which are defined by
\begin{align*}
    \textup{term \RNum{1}}:=\sqrt{\frac{2\P[\|\nabla\ell(\theta^*;z)\|^2]\log\frac{4}{\delta}}{n}}+\frac{G_{*}\log\frac{4}{\delta}}{n},
    \\
    \textup{term \RNum{2}}:=\sqrt{\frac{d+\log\frac{8\log_2(2n\Delta_M+2)}{\delta}}{n}}+\frac{d+\log\frac{\log_2(2n\Delta_M+2)}{\delta}}{n}.
\end{align*}

The result \eqref{eq: localized gradient both result 2} in Proposition \ref{prop localized convergence assumption both} implies that $\forall \theta\in\Theta$,
\begin{align*}
     \|\Pn \nabla \ell(\theta;z)-\P\nabla \ell(\theta;z)\|\leq  {\text{term \RNum{1}}}
    +c_0\beta\max\left\{\|\theta-\theta^*\|,\frac{1}{n}\right\}\cdot{\text{term \RNum{2}}}
    \\
    \leq \left(\text{term \RNum{1}}+\frac{c_0\beta}{n}\cdot \text{term \RNum{2}}\right)+c_0\beta\cdot \text{term \RNum{2}}\cdot\|\theta-\theta^*\|,
\end{align*}
where  $c_0$ is an absolute constant.
Since the PL condition implies  (see Lemma \ref{lemma curvature conditions}) that
\begin{align*}
    \P\ell(\theta;z)-\P\ell(\theta^*;z)\geq \frac{\mu}{2} \|\theta-\theta^*\|^2, \quad \forall \theta\in\Theta,
\end{align*}
we have 
\begin{align}\label{eq: proof pl algorithm stage 2}
     \|\Pn \nabla \ell(\theta;z)-\P\nabla \ell(\theta;z)\|^2\leq  2\left(\text{term \RNum{1}}+\frac{c_0\beta}{n}\cdot \text{term \RNum{2}}\right)^2\nonumber\\
    +\frac{4c_0^2\beta^2}{\mu}\left(\P\ell(\theta;z)-\P\ell(\theta^*;z)\right)(\text{term \RNum{2}})^2.
\end{align}

Combining \eqref{eq: proof pl algorithm stage 1} and \eqref{eq: proof pl algorithm stage 2}, we have that for all $t=0,1,\dots$,
\begin{align*}
    \mathcal{E}(\theta^{t+1})\leq \left(1-\frac{\mu}{\beta}\right)\mathcal{E}(\theta^t)+\eps^t(n),\\
    \eps^t(n)\leq \alpha(n)\mathcal{E}(\theta^t)+\eps^*(n),
\end{align*}
where 
\begin{align*}
\eps^t(n)=\frac{1}{2\beta}\|\Pn\nabla\ell(\theta^t;z)-\P\nabla\ell(\theta^t;z)\|^2,\\
    \alpha(n)=\frac{2c_0^2\beta}{\mu}(\text{term \RNum{2}})^2,\\
    \eps^*(n)=\frac{1}{\beta}\left(\text{term \RNum{1}}+\frac{c_0\beta}{n}\cdot \text{term \RNum{2}}\right)^2.
\end{align*}

Consider the following two conditions on the sample size (note that they will be satisfied as long as $n$ is large enough):
\begin{align}
\alpha(n)\leq \frac{\mu}{2\beta},\label{eq: requirement always contained 1}\\
\eps^*(n)\leq \frac{\mu^2}{4\beta}\Delta_m^2.\label{eq: requirement always contained 2}
\end{align}

Now consider the condition 
\begin{align*}
    n\geq \frac{c\beta^2}{\mu^2}\left(d+\log\frac{8\log_2(2n\Delta_M+2)}{\delta}\right),
\end{align*}
where $c=\max\{16c_0^2,1\}$ is an absolute constant. Then  \eqref{eq: requirement always contained 1} holds.
Since we also require the sample size $n$ to be large enough such that the ``statistical error" term in \eqref{eq: result theorem algorithm} is smaller than $\frac{\mu}{2}\Delta_m^2$,
the condition \eqref{eq: requirement always contained 2} is also true because
\begin{align*}
   \frac{\mu}{2}\Delta_m^2\geq \frac{16\P[\|\nabla\ell(\theta^*;z)\|^2]\log\frac{4}{\delta}}{\mu n}+\frac{8G_{*}^2\log^2\frac{4}{\delta}+\mu^2}{\mu n^2}\geq\frac{2}{\mu}\left(\text{term \RNum{2}}^2+\frac{\mu}{2n}\right)^2{\geq} \frac{2\beta}{\mu}\cdot \eps^*(n).
\end{align*}
 Therefore, both  condition \eqref{eq: requirement always contained 1} and condition \eqref{eq: requirement always contained 2} hold true under Theorem \ref{thm pl algorithm}'s requirement on the sample size.

Since both  \eqref{eq: requirement always contained 1} and \eqref{eq: requirement always contained 2} are true, we can use induction to prove that with probability at least $1-\delta$, for all $t=0,1,\dots$, $$\mathcal{E}(\theta^t)\leq \frac{\mu}{2}\Delta_m^2.$$
Therefore, for all $t=0,1,\dots,$
\begin{align*}
    \theta^t\in\pazocal{B}^d(\theta^*,\Delta_m)\subseteq\Theta.
\end{align*}

We choose the ``Lyapunov function" in Proposition \ref{prop iterative} to be the excess risk function $\mathcal{E}(\theta)$. Applying Proposition \ref{prop iterative}, we obtain that: when the sample size $n$ is large enough such that the conditions \eqref{eq: requirement always contained 1} and \eqref{eq: requirement always contained 2} hold ture,
we have
\begin{align*}
    \P\ell(\theta^t;z)-\P\ell(\theta^*;z)&\leq\left(1-\frac{\mu}{2\beta}\right)^t\mathcal{E}(\theta^0)+\frac{2\beta}{\mu}\cdot
    \eps^*(n)\\
    &\leq \left(1-\frac{\mu}{2\beta}\right)^t\mathcal{E}(\theta^0)+\frac{2}{\mu}\left(\text{term \RNum{1}}+\frac{\mu}{2n}\right)^2,
    \\
    &\leq \left(1-\frac{\mu}{2\beta}\right)^t\mathcal{E}(\theta^0)+\frac{16\P[\|\nabla\ell(\theta^*;z)\|^2]\log\frac{8}{\delta}}{\mu n}+\frac{8G_{*}^2\log^2\frac{4}{\delta}+\mu^2}{\mu n^2}.
\end{align*}
This completes the proof of Theorem \ref{thm pl algorithm}.
\hfill$\square$

\subsection{Proof of Corollary \ref{coro regression nonlinear}}\label{subsec proof application fast rate}
 
We first verify Assumption \ref{asm hessian noise}. We have
\begin{align*}
    \nabla^2\ell(\theta;z)=2\left(\eta'(\theta^Tx)^2+(\eta(\theta^Tx)-y)\eta''(\theta^Tx)]\right)xx^T.
\end{align*}
Since $x$ is $\tau-$sub-Gaussian, $xx^T$ is a $\tau^2-$sub-exponential. From the fact 
\begin{align*}
    \left|2(\eta'(\theta^Tx)^2+(\eta(\theta^Tx)-y)\eta''(\theta^Tx))\right|\leq C_{\eta}(B+C_{\eta}),
\end{align*}
Assumption \ref{asm hessian noise} holds with $\beta=2C_{\eta}(C_{\eta}+\sqrt{B})\tau^2$.

We then verify Assumption \ref{asm bernstein condition para}. We know
\begin{align*}
    \nabla \ell(\theta^*;z)=2(\eta(x^T\theta^*)-y)\eta'(x^T\theta^*)x.
\end{align*}
So we have for all $z$,
\begin{align*}
    \|\nabla \ell(\theta^*;z)\|\leq \sqrt{d}\|\nabla \ell(\theta^*;z)\|_{\infty}\leq  2C_{\eta}\sqrt{Bd}.
\end{align*}
So Assumption \ref{asm bernstein condition para} holds with $G_{*}=2C_{\eta}\sqrt{Bd}$.

Lastly, by [\citealp{foster2018uniform}, Lemma 5, inequality (16)], Assumption \ref{asm pl} holds with $\mu=\frac{2c_{\eta}^3\tau^2\gamma}{C_{\eta}}$. This completes the proof.
\hfill$\square$

\subsection{Proof of Theorem \ref{thm fo em}}\label{subsec proof em}

Before proving Theorem \ref{thm fo em},
we refer to [\citealp{balakrishnan2017statistical}, Theorem 1] for the following result on the population-based first-order EM update. 

\begin{lemma}[{\bf linear convergence of population-based first-order EM}]\label{lemma population fo em}
Under Assumption \ref{asm em strongly convex}, Assumption \ref{asm em gradient smoothness} and the condition that $\P\ell(\theta;z)$ is $\beta-$smooth, the following update,
\begin{align*}
    \theta^+=\theta-\frac{2}{\beta+\mu_1}\P\nabla\ell_{\theta}(\theta;z)
\end{align*}
satisfies that 
\begin{align*}
    \|\theta^+-\theta^*\|\leq \left(1-\frac{2\mu_1-\mu_2}{\beta+\mu_1}\right)\|\theta-\theta^*\|.
\end{align*}
\end{lemma}

\paragraph{}
We now prove Theorem \ref{thm fo em}.

\paragraph{Proof of Theorem \ref{thm fo em}:} Assumption \ref{asm hessian noise} implies that $\P\ell(\theta;z)$ is $\beta-$smooth, so Lemma \ref{lemma population fo em} holds under the assumptions of Theorem \ref{thm fo em}. Now we turn to analyze the sample-based first-order EM.
Consider the  update of sample-based first-order EM, 
\begin{align*}
    \theta^{t+1}=\theta^t-\frac{2}{\beta+\mu_1}\Pn\nabla\ell_{\theta^t}(\theta^t;z), \quad t=0,1,\dots
\end{align*}
 Fix $t\geq 0$. We have
\begin{align*}
    \|\theta^{t+1}-\theta^*\|\leq \|\theta^t-\frac{2}{\beta+\mu_1}\P\nabla\ell_{\theta^t}(\theta^t;z)\|+\frac{2}{\beta+\mu_1}\|(\P-\Pn)\nabla\ell_{\theta^t}(\theta^t;z)\|
    \\\leq \left(1-\frac{2\mu_1-\mu_2}{\beta+\mu_1}\right)\|\theta^t-\theta^*\|+\frac{2}{\beta+\mu_1}\|(\P-\Pn)\nabla\ell(\theta^t;z)\|.
\end{align*}

Applying Proposition \ref{prop localized convergence assumption both}, we continue the proof on the event $$\pazocal{A}:=\{\text{the results \eqref{eq: localized gradient both result 1} \eqref{eq: localized gradient both result 2} in  Proposition \ref{prop localized convergence assumption both} hold true}\},$$  whose measure is at least $1-\delta$.  We keep the notations ``term \RNum{1}" and ``term \RNum{2}" used in Proposition \ref{prop localized convergence assumption both}, which are defined by
\begin{align*}
    \textup{term \RNum{1}}:=\sqrt{\frac{2\P[\|\nabla\ell(\theta^*;z)\|^2]\log\frac{4}{\delta}}{n}}+\frac{G_{*}\log\frac{4}{\delta}}{n},
    \\
    \textup{term \RNum{2}}:=\sqrt{\frac{d+\log\frac{8\log_2(2n\Delta_M+2)}{\delta}}{n}}+\frac{d+\log\frac{\log_2(2n\Delta_M+2)}{\delta}}{n}.
\end{align*}

Note that we have the optimality condition $\nabla \ell(\theta^*;z)=0$, because the true parameter $\theta^*$ is assumed to minimizes the population risk over $\R^d$ in the problem setting. The result \eqref{eq: localized gradient both result 2} in Proposition \ref{prop localized convergence assumption both} implies that $\forall \theta\in\Theta$,
\begin{align}\label{eq: uniform convergence fo em}
     \|\Pn \nabla \ell(\theta;z)-\P\nabla \ell(\theta;z)\|\leq  {\text{term \RNum{1}}}
    +c_0\beta\max\left\{\|\theta-\theta^*\|,\frac{1}{n}\right\}\cdot{\text{term \RNum{2}}}
    \\
    \leq \left(\text{term \RNum{1}}+\frac{c_0\beta}{n}\cdot \text{term \RNum{2}}\right)+c_0\beta\cdot \text{term \RNum{2}}\cdot\|\theta-\theta^*\|,
\end{align}
where  $c_0$ is an absolute constant.
Therefore, we have that for all $t=0,1,\dots$,
\begin{align*}
    \mathcal{E}(\theta^{t+1})&\leq \left(1-\frac{2\mu_1-\mu_2}{\beta+\mu_1}\right)\mathcal{E}(\theta^t)+\eps^t(n),\\
    \eps^t(n)&\leq \alpha(n)\mathcal{E}(\theta^t)+\eps^*(n),
\end{align*}
where 
\begin{align*}
\eps^t(n)=\frac{2}{\beta+\mu_1}\|(\P-\Pn)\nabla\ell(\theta^t;z)\|,\\
    \alpha(n)=\frac{2c_0\beta}{\beta+\mu_1}\cdot\text{term \RNum{2}},\\
    \eps^*(n)=\frac{2}{\beta+\mu_1}\left(\text{term \RNum{1}}+\frac{c_0\beta}{n}\cdot \text{term \RNum{2}}\right).
\end{align*}

Consider the following two conditions on the sample size (note that they will be satisfied as long as $n$ is large enough):
\begin{align}
\alpha(n)\leq \frac{2\mu_1-\mu_2}{2(\beta+\mu_1)},\label{eq: requirement em 1}\\
\eps^*(n)\leq \frac{2\mu_1-\mu_2}{2(\beta+\mu_1)}\Delta_m.\label{eq: requirement em 2}
\end{align}

When the sample size $n$ is large enough so that both \eqref{eq: requirement em 1} and \eqref{eq: requirement em 2} are true,  we can use induction to prove that with probability at least $1-\delta$, for all $t=0,1,\dots$, $$\|\theta^t-\theta^*\|\leq \Delta_m^2.$$
Therefore, for all $t=0,1,\dots,$
\begin{align*}
    \theta^t\in\pazocal{B}^d(\theta^*,\Delta_m)\subseteq\Theta.
\end{align*}

We choose the  ``Lyapunov function" in Proposition \ref{prop iterative} to be  $\|\theta-\theta^*\|$. Applying Proposition \ref{prop iterative}, we obtain: when the sample size $n$ is large enough such that the conditions \eqref{eq: requirement em 1} and \eqref{eq: requirement em 2} hold ture,
we have
\begin{align}\label{eq: bound on norm fo em}
  \|\theta^t-\theta^*\|  \leq\left(1-\frac{2\mu_1-\mu_2}{2(\beta+\mu_1)}\right)^t\|\theta^0-\theta^*\|+\frac{2(\beta+\mu_1)}{2\mu_1-\mu_2}\cdot
    \eps^*(n)\nonumber\\
    \leq \left(1-\frac{2\mu_1-\mu_2}{2(\beta+\mu_1)}\right)^t\|\theta^0-\theta^*\|+\frac{4}{2\mu_1-\mu_2}\cdot\text{term \RNum{1}}+\frac{2}{n}.
\end{align}

When the sample size is large enough such that
\begin{align}\label{eq: fo em overall condition}
    n\geq  \frac{c\beta^2}{(2\mu_1-\mu_2)^2}\left(d+\log\frac{8\log_2(2n\Delta_M+2)}{\delta}\right) \text{\quad and \quad}  \text{term \RNum{1}}+\frac{2\mu_1-\mu_2}{2n}\leq \frac{(2\mu_1-\mu_2)\Delta_m}{4},
\end{align}
where $c=\max\{64c_0^2,1\}$ is an absolute constant, 
we have \begin{align*}
    \text{term \RNum{1}}\leq \frac{2\mu_1-\mu_2}{4}\left(\Delta_m-\frac{2}{n}\right) \text{\quad and \quad} \text{term \RNum{2}}\leq \frac{2\mu_1-\mu_2}{4c_0\beta},
\end{align*}
which further guarantee that both the condition \eqref{eq: requirement em 1} and the condition \eqref{eq: requirement em 2} are true.
We conclude that when the sample size condition  \eqref{eq: fo em overall condition} is true, we have the bound \eqref{eq: bound on norm fo em}.

Now we use the fact $\mu_1\leq 2\mu_1-\mu_2\leq 2\mu_1$ to simplify the sample size condition  \eqref{eq: fo em overall condition} and the bound \eqref{eq: bound on norm fo em}. It is straightforward to verify that the sample size condition  \eqref{eq: fo em overall condition} will be satisfied when 
\begin{align}\label{eq: sample size em final}
    n\geq \max\left\{ \frac{c\beta^2}{\mu_1^2}\left(d+\log\frac{8\log_2(2n\Delta_M+2)}{\delta}\right),\quad  \frac{128\P[\|\nabla\ell(\theta^*;z)\|^2]\log\frac{4}{\delta}}{\mu_1\Delta_M},\quad \frac{8G_*\log\frac{4}{\delta}+8\mu_1}{\mu_1\Delta_M} \right\};
\end{align}
and the bound \eqref{eq: bound on norm fo em} implies
\begin{align}\label{eq: estimation error bound simplify em}
     \|\theta^t-\theta^*\| 
    \leq \frac{4}{\mu_1}\left(\sqrt{\frac{2\P[\|\nabla\ell(\theta^*;z)\|^2]\log\frac{4}{\delta}}{n}}+\frac{G_{*}\log\frac{4}{\delta}+\mu_1}{n}\right)+\left(1-\frac{2\mu_1-\mu_2}{2(\beta+\mu_1)}\right)^t\|\theta^0-\theta^*\|.
\end{align}
Since we always have  $$\mathcal{E}(\theta^t)\leq \frac{\beta}{2}\|\theta^t-\theta^*\|^2,$$ the bound \eqref{eq: estimation error bound simplify em} will imply 
\begin{align}\label{eq: generalization error em final}
      \mathcal{E}(\theta^t)\leq \frac{16\beta}{\mu_1^2}\left(\sqrt{\frac{2\P[\|\nabla\ell(\theta^*;z)\|^2]\log\frac{4}{\delta}}{n}}+\frac{G_{*}\log\frac{4}{\delta}+\mu_1}{n}\right)^2 +\left(1-\frac{2\mu_1-\mu_2}{2(\beta+\mu_1)}\right)^{2t}\beta\|\theta^0-\theta^*\|^2.
\end{align}
Clearly, the sample size condition \eqref{eq: sample size em final} and the bounds \eqref{eq: estimation error bound simplify em} \eqref{eq: generalization error em final} are identical to those presented in the statement of the theorem. This completes the proof.
\hfill$\square$

\subsection{Proof of Corollary \ref{coro application example em}}\label{subsec proof application fast rate em}
For both examples, verification of Assumptions \ref{asm hessian noise}, \ref{asm bernstein condition para} and \ref{asm em strongly convex} is trivial. The parameters can be specified as $\beta=1$, $G_{*}=\sigma\sqrt{d}$, and $\mu_1=1$.

As for verification of Assumption \ref{asm em gradient smoothness},  we refer to the following results that are direct consequence of \cite{balakrishnan2017statistical}. 

\begin{lemma}[{\bf verification of Assumption \ref{asm em gradient smoothness}}]\label{lemma verification gradient smoothness}
(a) \textup{[\citealp{balakrishnan2017statistical}, Lemma 2]:} Consider  Example \ref{example mixture of Gaussian} under the SNR condition \eqref{eq: snr condition}, where $\eta$ is a sufficiently large constant such that $\eta>\frac{4\sqrt{3}}{3}$ and $c_1(1+\frac{1}{\eta^2}+\eta^2)e^{-c_2\eta^2}<1$. Then   Assumption \ref{asm em gradient smoothness} holds with $\mu_2=c_1(1+\frac{1}{\eta^2}+\eta^2)e^{-c_2\eta^2}$. Here $c_1$ and $c_2$ denote the same absolute constants as in the proof of \textup{[\citealp{balakrishnan2017statistical}, Lemma 2]}. Clearly, we can verify Assumption \ref{asm em gradient smoothness} for all $\eta$ larger than a certain absolute constant. 

(b) \textup{[\citealp{balakrishnan2017statistical}, Lemma 3]:} Consider  Example \ref{example mixture regression}  under the SNR condition \eqref{eq: snr condition}, where $\eta$ is a sufficiently large constant such that  \begin{align*}
    c\eta^{1-\frac{c_\tau^2}{2}}+c_\tau^2\frac{\log \eta}{\eta} +\frac{2}{\eta}\leq 
\frac{1}{8},\\
\sqrt{\frac{\|\theta^*\|}{8\eta}+(4+\frac{2}{31})\frac{C_\tau^2\log \eta}{\eta}+3\eta^{1-\frac{C_\tau^2}{2}}}\leq \frac{1}{8}
\end{align*} hold true for some sufficiently large constants $c_{\tau}, C_\tau$ and an absolute constant $c$. Then Assumption \ref{asm em gradient smoothness} holds with $\mu_2=\frac{1}{4}$. Here $c, c_{\tau}, C_{\tau}$ denote the same  quantity as in the proof of \textup{[\citealp{balakrishnan2017statistical}, Lemma 3]}. Clearly, we can verify Assumption \ref{asm em gradient smoothness} for all $\eta$ larger than a certain absolute constant.
\end{lemma}

To prove the generalization error bound in this corollary, we need to upper bound the problem-dependent parameter $\P[\|\nabla\ell(\theta^*;z)\|^2]$ for  the two examples.

\paragraph{Bounding $\P[\|\nabla\ell(\theta^*;z)\|^2]$ for Example \ref{example mixture of Gaussian}:} we define the function $g:\R^d\rightarrow (0,2)$ as 
\begin{align*}
    g(u)=\frac{2e^{-\frac{\|2\theta^*-u\|^2}{2\sigma^2}}}{e^{-\frac{\|u\|^2}{2\sigma^2}}+e^{-\frac{\|2\theta^*-u\|^2}{2\sigma^2}}}=\frac{2}{e^{\frac{2\|\theta^*\|^2-2u^T\theta^*}{\sigma^2}}+1}.
\end{align*}
In  Example \ref{example mixture of Gaussian},  when conditioned on $w=1$ (i.e., when $z$ is drawn from $N(\theta^*, \sigma^2 I_{d\times d})$), the random gradient $\nabla\ell(\theta^*;z)$ at $\theta^*$ can be shown to be equal to
\begin{align*}
   \left(\nabla\ell(\theta^*;z)|w=1\right)=u\underbrace{\bigg(\frac{e^{-\frac{\|u\|^2}{2\sigma^2}}-e^{-\frac{\|2\theta^*-u\|^2}{2\sigma^2}}}{e^{-\frac{\|u\|^2}{2\sigma^2}}+e^{-\frac{\|2\theta^*-u\|^2}{2\sigma^2}}}\bigg)}_{1-g(u) }+\theta^*\underbrace{\bigg(\frac{2e^{-\frac{\|2\theta^*-u\|^2}{2\sigma^2}}}{e^{-\frac{\|u\|^2}{2\sigma^2}}+e^{-\frac{\|2\theta^*-u\|^2}{2\sigma^2}}}\bigg)}_{g(u)},
   \end{align*}
  where $u=\theta^*-z$ is a random vector drawn from $N(0, \sigma^2I_{d\times d})$. And when conditioned on $w=-1$ (i.e., when $z$ is drawn from $N(0, \sigma^2 I_{d\times d})$), $\nabla\ell(\theta^*;z)$ can be shown to be  equal to
   \begin{align*}
   \left(\nabla\ell(\theta^*;z)|w=-1\right)=v\underbrace{\bigg(\frac{e^{-\frac{\|v\|^2}{2\sigma^2}}-e^{-\frac{\|2\theta^*-v\|^2}{2\sigma^2}}}{e^{-\frac{\|v\|^2}{2\sigma^2}}+e^{-\frac{\|2\theta^*-v\|^2}{2\sigma^2}}}\bigg)}_{1-g(v)}+\theta^*\underbrace{\bigg(\frac{2e^{-\frac{\|2\theta^*-v\|^2}{2\sigma^2}}}{e^{-\frac{\|v\|^2}{2\sigma^2}}+e^{-\frac{\|2\theta^*-v\|^2}{2\sigma^2}}}\bigg)}_{g(v)},
\end{align*}
 where $v=\theta^*+z$ is a random vector drawn from $N(0, \sigma^2I_{d\times d})$. 

Therefore, we have
\begin{align}
    &\P[\|\nabla \ell(\theta^*;z)\|^2]\nonumber
    \\&=\frac{1}{2}\E\left[\|\nabla\ell(\theta^*;z)\|^2|w=1\right] +\frac{1}{2}{\E\left[\|\nabla\ell(\theta^*;z)\|^2|w=-1\right]}\nonumber\\
    &=\frac{1}{2}\E_{u}[\|u \cdot (1-g(u))+\theta^*\cdot g(u)\|^2]+\frac{1}{2}\E_{v}[\|v \cdot (1-g(v))+\theta^*\cdot g(v)\|^2]\nonumber\\
    &=\E_{u}[\|u \cdot (1-g(u))+\theta^*\cdot g(u)\|^2],\label{eq: em optimal point bound expectation}
\end{align}
where the notation $\E_u$ means taking expectation with respect to $u\sim N(0,\sigma^2 I_{d\times d})$, and the notation $\E_v$ means taking expectation with respect to $v\sim N(0,\sigma^2 I_{d\times d})$.

Since $0<g(u)<2$, we have $|1-g(u)|\leq 1$. Thus
\begin{align}
    &\|u \cdot (1-g(u))+\theta^*\cdot g(u)\|^2\nonumber\\
    &\leq 2\|u\|^2\cdot |1-g(u)|^2+2\|\theta^*\|^2\cdot|g(u)|^2\nonumber\\
    &=2\|u\|^2+2\|\theta^*\|^2\cdot g(u)^2.\label{eq: em optimal point bound per sample}
\end{align}
From \eqref{eq: em optimal point bound expectation} and \eqref{eq: em optimal point bound per sample}, we have
\begin{align}
    \P[\|\nabla\ell(\theta^*;z)\|^2]\leq 2\E_{u}[\|u\|^2]+2\|\theta^*\|^2\E_{u}[g(u)^2]\nonumber\\
    =2\sigma^2d+2\|\theta^*\|^2\E_{u}[g(u)^2].\label{eq: em direct bound on problem dependent}
\end{align}

Now we know that $u^T\theta^*$ is a $\|\theta^*\|\sigma-$sub-Gaussian vector with mean $0$. From Markov's inequality, 
\begin{align}\label{eq: mixture Gaussian case 1}
    \textup{Prob}\left(|u^T\theta^*|>\frac{1}{2}\|\theta^*\|^2\right)\leq 2\exp(-\frac{\frac{1}{4}\|\theta^*\|^4}{\|\theta^*\|^2\sigma^2})\nonumber\\
    = \frac{2}{\exp(\frac{\|\theta^*\|^2}{4\sigma^2})}\leq \frac{8\sigma^2}{\|\theta^*\|^2}.
\end{align}
When $|u^T\theta^*|\leq\frac{1}{2}\|\theta^*\|^2$, we have
\begin{align*}
    g(u)=\frac{2}{e^{\frac{\|\theta^*\|^2-u^T\theta^*}{\sigma^2}}+1}\leq \frac{2}{e^{\frac{\|\theta^*\|^2}{2\sigma^2}}}\leq \frac{4\sigma^2}{\|\theta^*\|^2}.
\end{align*}
 Since $0<g(u)<2$, when $|u^T\theta^*|\leq\frac{1}{2}\|\theta^*\|^2$, we have
 \begin{align}\label{eq: mixture Gaussian case 2}
     g(u)^2\leq \frac{8\sigma^2}{\|\theta^*\|^2}.
 \end{align}
As a result,
\begin{align*}
   &\E_{u}[g(u)^2]\\&\leq  \textup{Prob}\left(|u^T\theta^*|>\frac{1}{2}\|\theta^*\|^2\right)\E\left[g(u)^2\bigg||u^T\theta^*|>\frac{1}{2}\|\theta^*\|^2\right]\\&\ \ \  +\textup{Prob}\left(|u^T\theta^*|\leq\frac{1}{2}\|\theta^*\|^2\right)\E\left[g(u)^2\bigg||u^T\theta^*|\leq\frac{1}{2}\|\theta^*\|^2\right]\\
   &\leq 4\cdot\textup{Prob}\left(|u^T\theta^*|>\frac{1}{2}\|\theta^*\|^2\right)+\frac{8\sigma^2}{\|\theta^*\|^2}\\
   &\leq \frac{40\sigma^2}{\|\theta^*\|^2},
\end{align*}
where the second inequality is due to the fact $0<g(u)<2$ and \eqref{eq: mixture Gaussian case 2}, and the last inequality is due to \eqref{eq: mixture Gaussian case 1}. Combining the above result with \eqref{eq: em direct bound on problem dependent}, we have 
\begin{align}\label{eq: bound gradient norm gaussian mixutre}
    \P[\|\nabla\ell(\theta^*;z)\|^2]\leq (2d+40)\sigma^2.
\end{align}

\paragraph{Bounding $\P[\|\nabla\ell(\theta^*;z)\|^2]$ for Example \ref{example mixture regression}:} we define the function 
 $g:\R\times \R^d\rightarrow (0,2)$ as 
\begin{align*}
    g(u,x)=\frac{2e^{-\frac{(2x^T\theta^*-u)^2}{2\sigma^2}}}{e^{-\frac{u^2}{2\sigma^2}}+e^{-\frac{(2x^T\theta^*-u)^2}{2\sigma^2}}}=\frac{2}{e^{\frac{2(x^T\theta^*)^2-2u(x^T\theta^*)}{\sigma^2}}+1}.
\end{align*}
In  Example \ref{example mixture regression}, we have
\begin{align*}
   \left(\nabla\ell(\theta^*;z)|w=1,x\right)=\left[u\underbrace{\bigg(\frac{e^{-\frac{u^2}{2\sigma^2}}-e^{-\frac{(2x^T\theta^*-u)^2}{2\sigma^2}}}{e^{-\frac{u^2}{2\sigma^2}}+e^{-\frac{(2x^T\theta^*-u)^2}{2\sigma^2}}}\bigg)}_{1-g(u,x) }+(x^T\theta^*)\underbrace{\bigg(\frac{2e^{-\frac{(2x^T\theta^*-u)^2}{2\sigma^2}}}{e^{-\frac{u^2}{2\sigma^2}}+e^{-\frac{(2x^T\theta^*-u)^2}{2\sigma^2}}}\bigg)}_{g(u,x)}\right]x,
   \end{align*}
 where $u=x^T\theta^*-y$ is a random vector drawn from $N(0, \sigma^2)$. And we have
   \begin{align*}
   \left(\nabla\ell(\theta^*;z)|w=-1,x\right)=\left[v\underbrace{\bigg(\frac{e^{-\frac{v^2}{2\sigma^2}}-e^{-\frac{(2x^T\theta^*-v)^2}{2\sigma^2}}}{e^{-\frac{v^2}{2\sigma^2}}+e^{-\frac{(2x^T\theta^*-v)^2}{2\sigma^2}}}\bigg)}_{1-g(v,x)}+(x^T\theta^*)\underbrace{\bigg(\frac{2e^{-\frac{(2x^T\theta^*-v)^2\|^2}{2\sigma^2}}}{e^{-\frac{v^2}{2\sigma^2}}+e^{-\frac{(2x^T\theta^*-v)^2}{2\sigma^2}}}\bigg)}_{g(v,x)}\right]x,
\end{align*}
 where $v=x^T\theta^*+y$ is a random vector drawn from $N(0, \sigma^2)$. 
 
 Therefore, we have
\begin{align}
    &\P[\|\nabla \ell(\theta^*;z)\|^2]\nonumber
    \\ &= \frac{1}{2}\E\left[\|\nabla\ell(\theta^*;z)\|^2|w=1\right] +\frac{1}{2}{\E\left[\|\nabla\ell(\theta^*;z)\|^2|w=-1\right]}\nonumber\\
    &=\frac{1}{2}\E_x\left[\|x\|^2\E_{u}[(u \cdot (1-g(u,x))+\theta^*\cdot g(u,x))^2|x]\right]+\frac{1}{2}\E_x\left[\|x\|^2\E_{v}[(v \cdot (1-g(v,x))+\theta^*\cdot g(v,x))^2]\right]\nonumber\\
    &=\E_x\left[\|x\|^2\E_{u}[(u \cdot (1-g(u,x))+\theta^*\cdot g(u,x))^2|x]\right],\label{eq: em optimal point bound expectation regression}
\end{align}
where the notation $\E_u$ means taking expectation with respect to $u\sim N(0,\sigma^2)$, the notation $\E_v$ means taking expectation with respect to $v\sim N(0,\sigma^2)$, and the notation $\E_x$ means taking expectation with respect to $x\sim N(0,I_{d\times d})$.

Similar to the last part (i.e., the proof of \eqref{eq: bound gradient norm gaussian mixutre}), we can prove
\begin{align*}
    \E_u[(u\cdot(1-g(u,x))+\theta^*\cdot g(u,x))^2|x]\leq 42 \sigma^2, \quad \forall x.
\end{align*}
Combine this result with \eqref{eq: em optimal point bound expectation regression}, we obtain
\begin{align*}
    \P[\|\nabla\ell(\theta^*;z)\|^2]\leq 42\sigma^2\E_x[\|x\|^2]=42\sigma^2d.
\end{align*}
This gives an upper bound on $\P[\|\nabla\ell(\theta^*;z)\|^2]$ in Example \ref{example mixture regression}.

Given that we have upper bounded $\P[\|\nabla\ell(\theta^*;z)\|^2]$ by $42\sigma^2d$ in both Example \ref{example mixture of Gaussian} and Example \ref{example mixture regression}, it is straightforward to prove the generalization error bound in   Corollary \ref{coro application example em}.
\hfill$\square$

\subsection{Auxiliary definitions and lemmata}\label{appendix auxiliary lemma parametric}

\begin{definition}[{\bf Orlicz norms, sub-Gaussian, sub-exponential}]\label{def orlicz subgaussian subexponential}
For every $\alpha\in(0,+\infty)$ we define  the Orlicz-$\alpha$ norm of a random $u$:
\begin{align*}
    \|u\|_{\text{Orlicz}_\alpha}=\inf \{K>0:\E\exp\big((|u|/K)^{\alpha}\big)\leq 2\}.
\end{align*}
A random variable/vector $X\in \R^d$ is $K-$sub-Gaussian if $\forall \lambda\in \R^d$, we have
\begin{align*}
    \| \lambda^T X\|_{\text{Orlicz}_2}\leq K\|\lambda\|_2.
\end{align*}
A random variable/vector $X\in \R^d$ is $K-$sub-exponential if $\forall \lambda\in \R^d$, we have
\begin{align*}
    \|\lambda^T X\|_{\text{Orlicz}_1}\leq K\|\lambda\|_2.
\end{align*}
\end{definition}

\begin{lemma}[{\bf Bernstein's inequality for sub-exponential random variables}]\label{lemma Bernstein orlicz}
If $X_1,\cdots,X_m$ are sub-exponential random variables,  then Bernstein's inequality (the inequality \eqref{eq: bernstein tail bound} in Lemma \ref{lemma Bernstein} holds with 
\begin{align*}
\sigma^2=\frac{1}{n}\sum_{i=1}^n\|X_i\|_{\textup{Orlicz}_1}^2, \quad B=\max_{1\leq i\leq n}\|X_i\|_{\textup{Orlicz}_1}.
\end{align*}
\end{lemma}

\begin{lemma}[{\bf vector Bernstein's inequality,  \cite{pinelis1994optimum, pinelis1999correction}} ]\label{lemma vector Bernstein}
Let $\{X_i\}_{i=1}^n$ be a sequence of i.i.d. random variables taking values in a real separable Hilbert space. Assume that $\mathbb{E}[X_i]=\mu$,  $\mathbb{E}[\|X_i-\mu\|^2]= \sigma^2$, $\forall 1\leq i\leq n$. We say that \emph{vector Bernstein's condition} with parameter $B$ holds if for all $1\leq i\leq n$,
\begin{align*}
\mathbb{E}[\|X_i-\mu\|^k]\leq \frac{1}{2}k!\sigma^2B^{k-2}, \quad \forall 2\leq k\leq n.
\end{align*}
If this condition holds, then for all $\delta\in(0,1)$, with probability at least $1-\delta$ we have
\begin{align*}
    \left\|\frac{1}{n}\sum_{i=1}^n X_i-\mu\right\|\leq \sqrt{\frac{2\sigma^2\log{\frac{2}{\delta}}}{n}}+\frac{B\log{\frac{2}{\delta}}}{n}.
\end{align*}
\end{lemma}

The following definitions and lemmata provide some background on generic chaining.

\begin{definition}[{\bf Orlicz-$\alpha$ processes}]
Let $\{X_{f}\}_{f\in \F}$ be a sequence of random variables. $\{X_{f}\}_{f\in \F}$ is called an Orlicz-$\alpha$ process for a metric $\texttt{metr}(\cdot,\cdot)$ on $\F$ if
\begin{align*}
    \|X_{f_1}-X_{f_2}\|_{\text{Orlicz}_\alpha}\leq \texttt{metr}(f_1,f_2), \forall f_1,f_2\in \F.
\end{align*}
In particular, Orlicz-2 process is called ``process with sub-Gaussian increments" and Orlicz-1 process is called ``process with sub-exponential increments".
\end{definition}

\begin{definition}[{\bf mixed sub-Gaussian-sub-exponential increments, \cite{dirksen2015tail}}]\label{def mixed tail}
We say a process $(X_{\theta})_{\theta\in\Theta}$ has mixed sub-Gaussian-sub-exponential increments with respect to the pair $(\texttt{metr}_1, \texttt{metr}_2)$ if for all $\theta_1,\theta_2\in \Theta$, \begin{align*}
    \text{Prob}\left(\|X_{\theta_1}-X_{\theta_2}\|\geq \sqrt{u}\cdot\texttt{metr}_2(\theta_1,\theta_2)+u\cdot\texttt{metr}_1(\theta_1,\theta_2)\right)\leq 2e^{-u}, \forall u\geq 0.
\end{align*}
\end{definition}

\begin{definition}[{\bf Talagrand's $\gamma_{\alpha}-$functional}]
A sequence $F = (\F_n)_{n\geq0}$ of subsets of $\F$ is called admissible if $|\F_0| = 1$ and
$|\F_n| \leq  2^{2^n}$
for all $n \geq 1$. For any $0 < \alpha < \infty$, the $\gamma_{\alpha}-$functional of $(\F, \texttt{metr})$ is defined
by
\begin{align*}
    \gamma_{\alpha}(F, d) = \inf_F
\sup_{f\in\F}\sum_{n=0}^{\infty}
2^{\frac{n}{\alpha}}\metr(f, \F_n),
\end{align*}
where the infimum is taken over all admissible sequences and we write $\texttt{metr}(f, \F_n) =
\inf_{s\in\F_n} \texttt{metr}(f, s)$.
\end{definition}

\begin{lemma}[{\bf Dudley's integral bound for $\gamma_{\alpha}$ functional,  \cite{talagrand1996majorizing}}]\label{lemma dudley}
There exist a constant $C_\alpha$ depending only on $\alpha$ such that 
\begin{align*}
    \gamma_{\alpha}(\F,\texttt{\textup{metr}})\leq C_{\alpha}
    \int_{0}^{+\infty}(\log N(\eps, \F, \texttt{\textup{metr}}))^{\frac{1}{\alpha}}d\eps.
\end{align*}
\end{lemma}

\begin{lemma}[{\bf generic chaining for a process with mixed tail increments, \cite{dirksen2015tail}}]\label{lemma mixed tail}
If $(X_{f})_{f\in\F}$ has mixed sub-Gaussian-sub-exponential increments with respect to the pair $(\metr_1, \metr_2)$, then there are absolute constants $c, C>0$ such that $\forall \delta\in(0,1)$, 
\begin{align*}
    \sup_{\theta\in\Theta}\|X_{f}-X_{f_0}\|
    \leq C(\gamma_2(\F, \texttt{\textup{metr}}_2)+\gamma_1(\F,\texttt{\textup{metr}}_1))+\\c\left(\sqrt{\log\frac{1}{\delta}}\sup_{f_1,f_2\in\F}[\texttt{\textup{metr}}_2(f_1,f_2)]
    +\log\frac{1}{\delta}\sup_{f_1,f_2\in\F}[\texttt{\textup{metr}}_1(f_1,f_2)]\right),
\end{align*}
with probability at least $1-\delta$.
\end{lemma}

\section{Proofs for Section \ref{sec supervised learning strongly convex}}\label{subsec appendix thm sup}


\subsection{Proof of Theorem \ref{thm sup}}\label{subsec sup}

The proof consists of five parts.
Among them, the main purpose of Part \RNum{1} and Part \RNum{4} is to localized the strong convexity parameter. When there is no need to localized the strong convexity parameter (e.g., when one uses the square cost), the proof can be simplified---Part \RNum{1} and Part \RNum{4} will be quite straightforward, and all the ``upper-side" truncation analysis related to $\frac{2\|\xi\|_{L_2}}{\sqrt{c_\kappa}}$,  $\frac{4\|\xi\|^2_{L_2}}{c_\kappa}$ or $\frac{4\|\xi\|^2_{L_2}}{\kappa^2 c_\kappa}$ will be  unnecessary.

\paragraph{\bf Part \RNum{1}: analysis of the  concentrated functions.}\

Denote $T(h)=\|h-h^*\|^2_{L_2}$ and
 \begin{align*}
 v_h=\min \left\{\kappa\|h-h^*\|_{L_2},\frac{2\|\xi\|_{L_2}}{\sqrt{c_\kappa}}\right\}.
 \end{align*}
For every $h\in\Hy$, define 
 \begin{align*}
 f_h(x,y)=\frac{2}{\alpha\left(4\|\xi\|_{L_2}/\sqrt{c_\kappa}\right)}\partial_1\ell_{\sv}(h^*(x),y)(h(x)-h^*(x)),\\
g_h(x,y)=\min\left\{(h(x)-h^*(x))^2, v_h^2\right\}\cdot\mathds{1}\left\{|\xi|\leq \frac{2\|\xi\|_{L_2}}{\sqrt{c_\kappa}}\right\}.
\end{align*} 
 One can view $g_h$ as a truncated version of the quadratic form $(h(x)-h^*(x))^2$. Later we will use concentration to control $(\P-\Pn)(f_h+g_h)$  uniformly.

From Lemma \ref{lemma sup taylor lower bound} (for which we defer to the end of Section \ref{subsec sup}), we can show
\begin{align*}
    \ell_{\sv}(h(x),y)-\ell_{\sv}(h^*(x),y)-\partial_1\ell_{\sv}(h^*(x),y)(h(x)-h^*(x))
    \geq \frac{{\alpha(2v_h)}}{2}\min\left\{(h(x)-h^*(x))^2, v_h^2\right\}\nonumber\\
    \geq \frac{\alpha\left(4\|\xi\|_{L_2}/\sqrt{c_\kappa}\right)}{2}g_h(x,y).
\end{align*}
The above inequality implies that
\begin{align}\label{eq: sup empirical uppper bound}
    \Pn (f_h+g_h)\leq \frac{2}{\alpha\left(4\|\xi\|_{L_2}/\sqrt{c_\kappa}\right)}\Pn [\ell_{\sv}(h(x),y)-\ell_{\sv}(h^*(x),y)].
\end{align}

Recall that $\xi=h^*(x)-y$. By Markov's inequality, 
\begin{align}\label{eq: markov xi}
    \text{Prob}\left(|\xi|\geq \frac{2\|\xi\|_{L_2}}{\sqrt{c_\kappa}}\right)\leq \frac{c_{\kappa}}{4},
\end{align}From the definition of $g_h$ and $v_h$, it is straightforward to show that
\begin{align*}
   &\P g_h=\P\left[\min\left\{(h(x)-h^*(x))^2, v_h^2\right\}\cdot\mathds{1}\left\{|\xi|\leq \frac{2\|\xi\|_{L_2}}{\sqrt{c_{\kappa}}}\right\}\right]\nonumber\\
   &\geq \P \left[\min\left\{(h(x)-h^*(x))^2, v_h^2\right\}\cdot \mathds{1}\left\{  |h(x)-h^*(x)|\geq \kappa\|h-h^*\|_{L_2}\right\}\cdot\mathds{1}\left\{  |\xi|\leq \frac{2\|\xi\|_{L_2}}{\sqrt{c_{\kappa}}}\right\}\right]\nonumber\\
   &\geq\P \left[v_h^2\cdot  \mathds{1}\left\{  |h(x)-h^*(x)|\geq \kappa\|h-h^*\|_{L_2}\right\}\cdot\mathds{1}\left\{|\xi|\leq \frac{2\|\xi\|_{L_2}}{\sqrt{c_{\kappa}}}\right\}\right]\nonumber\\
   &=v_h^2\cdot\text{Prob}\left(  |h(x)-h^*(x)|\geq \kappa\|h-h^*\|_{L_2},\ |\xi|\leq \frac{2\|\xi\|_{L_2}}{\sqrt{c_{\kappa}}}\right)\nonumber\\
   &\geq v_h^2 \cdot\left(\text{Prob}\left(|h(x)-h^*(x)|\geq\kappa\|h-h^*\|_{L_2}\right)-\text{Prob}\left(\|\xi|>\frac{2\|\xi\|_{L_2}}{\sqrt{c_\kappa}}\right)\right)\\&\geq \frac{3c_\kappa}{4}v_h^2,
\end{align*}
where the first inequality is due to $1\geq \mathds{1}\left\{  |h(x)-h^*(x)|\geq  \kappa\|h-h^*\|_{L_2}\right\}$; the second inequality is due to the definition of $v_h$; and the last inequality is due to Assumption \ref{asm moment condition} and \eqref{eq: markov xi}.
From Assumption \ref{asm regularity supervised}, we have
\begin{align}\label{eq: sup population lower bound}
    \P (f_h+g_h)\geq \P g_h\geq \frac{3c_\kappa}{4}v_h^2.
\end{align}

Let us summarize the results from this part. We use the empirical average of the excess loss to upper bound $\Pn (f_h+g_h)$ in \eqref{eq: sup empirical uppper bound}, and use the (truncated) quadratic form to lower bound $\P (f_h+g_h)$ in \eqref{eq: sup population lower bound}. The next steps are to prove concentration of $f_h$ and $g_h$ and establish a ``uniform localized convergence" argument.

\paragraph{\bf Part \RNum{2}: bound the localized empirical process.}\

Given $r>0$, we want to bound the localized empirical process $$\sup_{\frac{r}{\lambda}\leq T(h)\leq r}(\P-\Pn) (f_h+g_h)$$
where $\lambda>1$ is a fixed value that we will specify later.
From the definition of $\varphi_{\text{noise}}(r;\delta)$ in Assumption \ref{asm regularity surrogate sup}, for any $\delta\in(0,1)$, with probability $1-\frac{\delta}{2}$, we have
\begin{align}\label{eq: upper bound fh gh}
   \sup_{\frac{r}{\lambda}\leq T(h)\leq r}(\P-\Pn) (f_h+g_h)\leq  \sup_{\frac{r}{\lambda}\leq T(h)\leq r}(\P-\Pn) g_h + \frac{2}{\alpha\left(4\|\xi\|_{L_2}/\sqrt{c_\kappa}\right)}\varphi_{\text{noise}}\left(r;\frac{\delta}{2}\right).
\end{align}
Given $r>0$, denote the hypothesis class  $\Hy(\frac{r}{\lambda},r)=\{h\in\Hy:\frac{r}{\lambda}\leq T(h)\leq r\}$, and define the function $g_{h,r}$ as $$g_{h,r}(x,y)=\min\left\{ (h(x)-h^*(x))^2, \kappa^2 r, \frac{4\|\xi\|_{L_2}^2}{c_{\kappa}}\right\}\cdot\mathds{1}\left\{|\xi|\leq \frac{2\|\xi\|_{L_2}}{\sqrt{c_\kappa}}\right\}.$$ Recall that $g_h$ is defined by
\begin{align*}
    g_h(x,y)=\min\left\{(h(x)-h^*(x))^2, \kappa^2T(h), \frac{4\|\xi\|_{L_2}^2}{c_\kappa}\right\}\cdot\mathds{1}\left\{|\xi|\leq \frac{2\|\xi\|_{L_2}}{\sqrt{c_\kappa}}\right\}.
\end{align*} For every $h\in\Hy(\frac{r}{\lambda},r)$ and any $(x,y)\in\pazocal{X}\times\pazocal{Y}$,  we have
\begin{align}\label{eq: sandwich 1}
    g_{h,\frac{r}{\lambda}}(x,y) \leq  g_h(x,y) \leq  g_{h,r}(x,y),
\end{align}
and
\begin{align}\label{eq: sandwich 2}
  &g_{h,r}(x,y)-g_{h,\frac{r}{\lambda}}(x,y)  \nonumber\\&\leq \min\left\{(h(x)-h^*(x))^2, \kappa^2r, 
  \frac{4\|\xi\|_{L_2}^2}{c_k}\right\}-\min\left\{(h(x)-h^*(x))^2,\frac{\kappa^2 r}{\lambda}, \frac{4\|\xi\|_{L_2}^2}{c_\kappa}\right\}\nonumber\\
  &\leq \min\left\{\kappa^2r, 
  \frac{4\|\xi\|_{L_2}^2}{c_k}\right\}-\min\left\{ \frac{\kappa^2 r}{\lambda}, \frac{4\|\xi\|_{L_2}^2}{c_\kappa}\right\}\nonumber\\ 
  &\leq \left(1-\frac{1}{\lambda}\right)\min\left\{ \kappa^2 r, \frac{4\|\xi\|_{L_2}^2}{ c_\kappa}\right\}.
\end{align}
From \eqref{eq: sandwich 1} and \eqref{eq: sandwich 2}, for every $h\in\Hy(\frac{r}{\lambda},r)$ and any $(x,y)\in\pazocal{X}\times\pazocal{Y}$,
$$-\left(1-\frac{1}{\lambda}\right){ \min\left\{\kappa^2 r,\frac{4\|\xi\|_{L_2}^2}{c_\kappa}\right\}}\leq g_h(x,y)-g_{h,r}(x,y)\leq 0,$$
which implies
\begin{align*}
  (\P-\Pn)g_h \leq (\P-\Pn) g_{h,r}+\left(1-\frac{1}{\lambda}\right){ \min\left\{\kappa^2 r,\frac{4\|\xi\|_{L_2}^2}{c_\kappa}\right\}}.
\end{align*}
As a result, we have
\begin{align}
    \sup_{\frac{r}{K}\leq T(h)\leq r}(\P-\Pn)g_h\leq \sup_{\frac{r}{K}\leq T(h)\leq r}(\P-\Pn) g_{h,r}+\left(1-\frac{1}{\lambda}\right){ \min\left\{\kappa^2 r,\frac{4\|\xi\|_{L_2}^2}{c_\kappa}\right\}}\nonumber\\
    \leq \sup_{T(h)\leq r}(\P-\Pn) g_{h,r}+\left(1-\frac{1}{\lambda}\right){ \min\left\{\kappa^2 r,\frac{4\|\xi\|_{L_2}^2}{c_\kappa}\right\}}.\label{eq: bound fh by fhr}
\end{align} 

 We know that $g_{h,r}$ is uniformly bounded by $\left[0,\min\left\{\kappa^2 r,\frac{4\|\xi\|_{L_2}^2}{c_\kappa}\right\}\right]$. Form the standard bound for global Rademacher complexity \cite{wainwright2019high}, $\forall \delta\in(0,1)$, with probability at least $1-\frac{\delta}{2}$,
\begin{align}\label{eq: global Rademahcer}
    \sup_{T(h)\leq r}(\P-\P_n)g_{h,r}\leq  2\mathfrak{R} \{g_{h,r}:T(h)\leq r\}+\min\left\{\kappa^2 r,\frac{4\|\xi\|_{L_2}^2}{c_\kappa}\right\}\sqrt{\frac{\log\frac{2}{\delta}}{2n}}.
\end{align}
It is straightforward to verify that for all $h_1,h_2\in\Hy$ and $(x,y)\in\pazocal{X}\times\pazocal{Y}$,
\begin{align*}
    |g_{h_1,r}(x)-g_{h_2,r}(x)|\leq 2\kappa\sqrt{r}|h_1(x)-h_2(x)|.
\end{align*}
From the Lipchitz contraction property of Rademacher complexity (see, e.g., [\citealp{meir2003generalization}, Theorem 7]), we have
\begin{align}\label{eq: after contraction}
    \mathfrak{R}\{g_{h,r}\}\leq 2\kappa\sqrt{r}\mathfrak{R}\{h:T(h)\leq r\}\leq 2\kappa\sqrt{r}\varphi(r),
\end{align}
where $\varphi(r)$ is defined in Assumption \ref{asm regularity surrogate sup}.
Define the $\psi$ function as
\begin{align}\label{eq: choice psi functional sup}
    \psi(r;\delta)=4\kappa\sqrt{r}\varphi(r)+       \left(\sqrt{\frac{\log\frac{2}{\delta}}{2n}}+1-\frac{1}{\lambda}\right)\min\left\{{\kappa^2} r,\frac{4\|\xi\|_{L_2}^2}{c_\kappa}\right\} +\frac{2}{\alpha\left(4\|\xi\|_{L_2}/\sqrt{c_\kappa}\right)}\varphi_{\text{noise}}\left(r;\frac{\delta}{2}\right).
\end{align}
Combining the definition  \eqref{eq: choice psi functional sup} with \eqref{eq: upper bound fh gh} \eqref{eq: bound fh by fhr} \eqref{eq: global Rademahcer} \eqref{eq: after contraction} , for any $ \delta\in(0,1)$, with probability at least $1-\delta$, we have
\begin{align}\label{eq: psi well defined sup}
    \sup_{\frac{r}{K}\leq T(h)\leq r}(\P-\P_n)(f_h+g_h)\leq \psi(r;\delta).
\end{align}

\paragraph{\bf Part \RNum{3}: the ``uniform localized convergence" argument.}\

Applying Proposition \ref{prop peeling stronger}, for any $\delta_1\in(0,1)$ and $r_0\in(0,4\Delta^2)$, with probability at least $1-\delta_1$, for all $h\in\Hy$, either $T(h)\leq r_0$ or 
\begin{align}\label{eq: variant peeling argument}
    &(\P-\Pn) (f_h+g_h)\leq  \psi\left(\lambda T(h);\delta_1({\log_{K}\frac{4K\Delta^2}{r_0}})^{-1}\right)\nonumber\\
    &= 4\kappa\sqrt{\lambda T(h)}\varphi(\lambda T(h))+\frac{2}{\alpha\left(4\|\xi\|_{L_2}/\sqrt{c_\kappa}\right)}\varphi_{\text{noise}}\left(\lambda T(h);\frac{\delta_1}{2\log_{\lambda}\frac{4\lambda\Delta^2}{r_0}}\right)\nonumber\\&\ \ \ +\underbrace{\min\left\{\lambda \kappa^2T(h),\frac{4\|\xi\|_{L_2}^2}{c_\kappa}\right\}\left(\sqrt{\frac{\log\frac{2\log_{\lambda}\frac{4\lambda\Delta^2}{r_0}}{\delta_1}}{2n}}+1-\frac{1}{\lambda}\right)}_{\text{last term in \eqref{eq: variant peeling argument}}}.
\end{align}
We specify
\begin{align*}
    \lambda=\frac{8+2c_\kappa}{8+c_\kappa}.
\end{align*} 
Then when $
    n> \frac{32}{c_\kappa^2}\log\frac{2\log_{\lambda}\frac{4\lambda\Delta^2}{r_0}}{\delta_1}$,
for all $h\in\Hy$ we have
\begin{align*}
   \lambda \left(\sqrt{\frac{\log\frac{2\log_{\lambda}\frac{4\lambda\Delta^2}{r_0}}{\delta_1}}{2n}}+1-\frac{1}{\lambda}\right)< \frac{c_\kappa}{4},
\end{align*}
which implies when $T(h)>0$,
\begin{align} \label{eq: relax middle term}
\text{last term in \eqref{eq: variant peeling argument}} < \frac{c_\kappa}{4} \min\left\{\kappa^2 r,\frac{4\|\xi\|_{L_2}^2}{\lambda c_\kappa}\right\}\leq \frac{c_\kappa}{4}\min\left\{\kappa^2 r, \frac{4\|\xi\|_{L_2}^2}{c_\kappa}\right\}.
\end{align}
Denote $C_{r_0}= 2+\left(\frac{16}{c_\kappa}+2\right)\log\frac{4\Delta^2}{r_0}$, then 
\begin{align*}
   2\log_{\lambda}\frac{4\lambda\Delta^2}{r_0}=2+\frac{\log\frac{4\Delta^2}{r_0}}{\log \lambda}\leq 2+\left(\frac{16}{c_\kappa}+2\right)\log\frac{4\Delta^2}{r_0}=C_{r_0}.
\end{align*}
For any $\delta\in(0,1)$, taking $\delta_1=\frac{2\log_{\lambda}\frac{4\lambda\Delta^2}{r_0}}{C_{r_0}}\delta$, from \eqref{eq: variant peeling argument} \eqref{eq: relax middle term} and the fact $\lambda<2$, we have the following conclusion: when 
$ n>\frac{32}{c_\kappa^2}\log\frac{C_{r_0}}{\delta}$, with probability at least $1-\delta$, for all $h\in\Hy$, either $T(h)\leq r_0$ or
\begin{align}\label{eq: clean peeling bound}
    &(\P-\Pn) (f_h+g_h)\nonumber\\&<  4\kappa\sqrt{2 T(h)}\varphi(2 T(h))+\frac{2}{\alpha\left(4\|\xi\|_{L_2}/\sqrt{c_\kappa}\right)}\varphi_{\text{noise}}\left(2T(h);\frac{\delta}{C_{r_0}}\right)+\frac{c_\kappa}{4}\min\left\{\kappa^2 T(h),\frac{4\|\xi\|^2_{L_2}}{c_\kappa}\right\}.
\end{align}

Let $\hh\in\argmin \Pn \ell_{\sv}(h(x),y)$ be the empirical risk minimizer. From \eqref{eq: sup empirical uppper bound} and the property of $\hh$, we have
\begin{align}
    \Pn (f_{\hh}+g_{\hh})\leq \frac{2}{\alpha\left(4\|\xi\|_{L_2}/\sqrt{c_{\kappa}}\right)}\Pn [\ell_{\sv}(\hh(x)-y)-\ell_{\sv}(h^*(x)-y)]
    \leq 0.\label{eq: sup erm plays 1}
\end{align}
Recall the result \eqref{eq: sup population lower bound} proved in Part \RNum{1},  \begin{align}\label{eq: recall part 1}
\P(f_h+g_h)\geq \frac{3c_\kappa}{4}v_h^2=\frac{3c_\kappa}{4}\min\left\{\kappa^2T(h),\frac{4\|\xi\|_{L_2}^2}{c_\kappa}\right\}.\end{align} 
From \eqref{eq: clean peeling bound} \eqref{eq: sup erm plays 1} \eqref{eq: recall part 1},  when
$ n>\frac{32}{c_\kappa^2}\log\frac{C_{r_0}}{\delta}$, with probability at least $1-\delta$, either $T(\hh)\leq r_0$ or 
\begin{align*}
 &\frac{3c_\kappa}{4}\min\left\{\kappa^2 T(\hh), \frac{4\|\xi\|_{L_2}^2}{c_{\kappa}}\right\}
   \nonumber\\&<  4\kappa\sqrt{2 T(\hh)}\varphi(2 T(\hh))+\frac{2}{\alpha\left(4\|\xi\|_{L_2}/\sqrt{c_\kappa}\right)}\varphi_{\text{noise}}\left(2 T(\hh);\frac{\delta}{C_{r_0}}\right)+\frac{c_\kappa\kappa^2}{4}\min\left\{T(\hh), \frac{4\|\xi\|_{L_2}^2}{c_{\kappa}}\right\},\end{align*} 
i.e., 
\begin{align}\label{eq: tempT3 4}
 &\frac{c_\kappa}{2}\min\left\{\kappa^2 T(\hh), \frac{4\|\xi\|_{L_2}^2}{c_{\kappa}}\right\}
   \nonumber\\&<  4\kappa\sqrt{2 T(\hh)}\varphi(2 T(\hh))+\frac{2}{\alpha\left(4\|\xi\|_{L_2}/\sqrt{c_\kappa}\right)}\varphi_{\text{noise}}\left(2 T(\hh);\frac{\delta}{C_{r_0}}\right).
\end{align}
In the theorem we have asked $ n>\frac{72}{c_\kappa^2}\log\frac{C_{r_0}}{\delta}$. Denote the event 
\begin{align*}
    \mathcal{A}:=\{\text{either $T(\hh)\leq r_0$ or \eqref{eq: tempT3 4} is true}\}.
\end{align*}
Then we have $\text{Prob}(\mathcal{A})\geq 1-\delta$.

\paragraph{Part \RNum{4}: preliminary localization.}\ 

We first prove a preliminary localization result $T(\hh)\leq \max\left\{\frac{4\|\xi\|_{L_2}^2}{\kappa^2 c_{\kappa}}, r_0\right\}$ on the event $\mathcal{A}$. The essential purpose of this step is to localize the strong convexity parameter. If $T(\hh)\in\left( \max\left\{\frac{4\|\xi\|_{L_2}^2}{\kappa^2c_{\kappa}}, r_0\right\}, 4\Delta^2\right]$ is true,  then on the event $\mathcal{A}$ one have 
\begin{align}\label{eq: contradiction 1 sup}
\text{RHS of \eqref{eq: tempT3 4}}> 2\|\xi\|_{L_2}^2.
\end{align}
 In the theorem we ask $ n>\max\left\{\bar{N}_{\delta,r_0},\frac{72}{c_\kappa^2}\log\frac{C_{r_0}}{\delta}\right\}$. According to Assumption \ref{asm regularity surrogate sup}, this implies that 
\begin{align}
     \varphi_{\text{noise}}\bigg(8\Delta^2; \frac{\delta}{C_{r_0}}\bigg)\leq  \frac{\alpha(4\|\xi\|_{L_2}/\sqrt{c_\kappa})\|\xi\|_{L_2}^2}{2}, \label{eq: sample condition sup 1}\\
      \varphi\left(8\Delta^2\right)\leq \frac{\sqrt{2c_\kappa}\|\xi\|^2_{L_2}}{16\Delta},
      \label{eq: sample condition sup 2}
 \end{align} 
 which further imply
 \begin{align}\label{eq: contradiction 2 sup}
     \text{RHS of \eqref{eq: tempT3 4}}\leq \|\xi\|_{L_2}^2+\|\xi\|^2
   =2\|\xi\|_{L_2}^2.
 \end{align}
\eqref{eq: contradiction 1 sup} and \eqref{eq: contradiction 2 sup} result in a contradiction. Therefore,  $T(\hh)$ must be bounded by $\max\left\{\frac{4\|\xi\|_{L_2}^2}{\kappa^2c_{\kappa}}, r_0\right\}$.
  Then on the event $\mathcal{A}$, either $T(\hh)\leq r_0$ or
\begin{align}\label{eq: bound Th by surrogates then}
   \frac{c_\kappa\kappa^2}{2}T(\hh) < \text{RHS of \eqref{eq: tempT3 4}}.
\end{align}

\paragraph{Part \RNum{5}: final steps.}\ 

Let $r_{\text{noise}}^*$ be the fixed point of
\begin{align*}
   \frac{4}{c_\kappa \kappa^2\cdot \alpha\left(4\|\xi\|_{L_2}/\sqrt{c_\kappa}\right)}\varphi_{\text{noise}}\left(2r;\frac{\delta}{C_{r_0}}\right),
\end{align*}
and $r_{\text{ver}}^*$ be the fixed point of 
\begin{align*}
   \frac{8}{c_\kappa\kappa}\sqrt{2r}\varphi(2r).
\end{align*}
 From  the definition of fixed points, 
 when $T(\hh)> \max\{r^*_{\text{ver}},\  r^*_{\text{noise}}\}$,
 we have
 \begin{align*}
    \frac{c_\kappa\kappa^2}{4} T(\hh)>  \frac{2}{\alpha\left(4\|\xi\|_{L_2}/\sqrt{c_\kappa}\right)}\varphi_{\text{noise}}\left(2 T(\hh);\frac{\delta}{C_{r_0}}\right)
 \end{align*}
 and 
 \begin{align*}
     \frac{c_\kappa\kappa^2}{4} T(\hh)> 4\kappa\sqrt{2 T(\hh)}\varphi(2 T(\hh)).
 \end{align*}
 Contrasting the above two inequalities with our previous result \eqref{eq: bound Th by surrogates then},  on the event $\mathcal{A}$ we have
\begin{align*}
T(\hh)\leq \max\{r^*_{\text{ver}},\  r^*_{\xi},\ r_0\}.
\end{align*}
We conclude that when $ n>\max\left\{\bar{N}_{\delta,r_0},\frac{72}{c_\kappa^2}\log\frac{C_{r_0}}{\delta}\right\}$,
 with probability at least $1-\delta$,
\begin{align}\label{eq: sup estimation error bound proof}
    \|\hh-h^*\|^2_{L_2}\leq \max\{r^*_{\text{noise}},\ r^*_{\text{ver}},\  r_0\}.
\end{align}

Finally, from the optimality condition on $h^*$ (Assumption \ref{asm regularity supervised}), it is straightforward to prove that for all $h\in\Hy$, $$\mathcal{E}(h)\leq\frac{\beta_{\sv}}{2}\|h-h^*\|^2_{L_2}.$$
Combining the above inequality with \eqref{eq: sup estimation error bound proof}, we have
\begin{align*}
    \mathcal{E}(\hh)\leq\frac{\beta_{\sv}}{2}\max\left\{ r^*_{\textup{noise}},\ r^*_{\textup{ver}},\ r_0\right\}.
\end{align*}
This completes the proof.
\hfill$\square$

\paragraph{}

\begin{lemma}[{\bf lower bound of the residual of the Taylor expansion}]\label{lemma sup taylor lower bound}
Let $\ell_{\sv}$ be convex with respect to its first argument. Given $v>0$, for all  $u_1,u_2\in\R$ and $y\in\pazocal{Y}$, we have
\begin{align}\label{eq: sup taylor lower bound}
\ell_{\sv}(u_1,y)-\ell_{\sv}(u_2,y)-\partial_1\ell_{\sv}(u_2,y)(u_1-u_2)    \geq \frac{\alpha(2v)}{2}\min\{|u_1-u_2|^2, v^2\}\cdot\mathds{1}\{|u_2-y|\leq v\}.
\end{align}
  \end{lemma}
  
\paragraph{Proof of Lemma \ref{lemma sup taylor lower bound}:}
 we consider the following four cases: (1)  $|u_2-y|> v$; (2)  $|u_2-y|\leq v$ and $|u_1-u_2|\leq v$;  (3)    $|u_2-y|\leq v$ and $u_1-u_2> v$ ; and (4) $|u_2-y|\leq v$ and $u_1-u_2<- v$. It is straightforward to prove \eqref{eq: sup taylor lower bound} in case (1) and case (2). In case (3), because
\begin{align*}
    \ell_{\sv}(u_1,y)-\ell_{\sv}(u_2,y)-\partial_1\ell_{\sv}(u_2,y)(u_1-u_2) =\int_0^1(\partial_1\ell_{\sv}(u_2+t(u_1-u_2))-\partial_1\ell_{\sv}(u_2))(u_1-u_2)dt,
\end{align*}
and $(\partial_1\ell_{\sv}(u_2+t(u_1-u_2))-\partial_1\ell_{\sv}(u_2))(u_1-u_2)\geq 0$ for all $t\in[0,1]$,
 we have
\begin{align*}
    \ell_{\sv}(u_1,y)-\ell_{\sv}(u_2,y)-\partial_1\ell_{\sv}(u_2,y)(u_1-u_2) \geq \int_0^\frac{v}{u_1-u_2}(\partial_1\ell_{\sv}(u_2+t(u_1-u_2))-\partial_1\ell_{\sv}(u_2))(u_1-u_2)dt\\
    =\ell_{\sv}(u_2+v,y)-\ell_{\sv}(u_2,y)-\partial_1\ell_{\sv}(u_2,y)v\geq \frac{\alpha(2v)}{2} v^2.
\end{align*}
Similarly, we can prove \eqref{eq: sup taylor lower bound} in case (4). This completes the proof of Lemma \ref{lemma sup taylor lower bound}.
\hfill$\square$

\subsection{Proof of Corollary \ref{coro sup}}
The proof is nearly identical to the proof of Theorem \ref{thm sup}, but with the following modifications. First, we only need to consider the hypothesis set $\Hy_0$. 
Second, based on the definition of $\varphi_{\text{noise}}$ in Corollary \ref{coro sup}, we modify \eqref{eq: upper bound fh gh} to
\begin{align}
   \sup_{h\in\Hy_0,\frac{r}{\lambda}\leq T(h)\leq r}(\P-\Pn) (f_h+g_h)\leq  \sup_{\frac{r}{\lambda}\leq T(h)\leq r}(\P-\Pn) g_h + \frac{2}{\alpha\left(4\|\xi\|_{L_2}/\sqrt{c_\kappa}\right)}\left(\varphi_{\text{noise}}\left(r;\frac{\delta}{2}\right)-\Phi(h^*)\right).
\end{align}
We also do similar modifications to \eqref{eq: choice psi functional sup} \eqref{eq: psi well defined sup} \eqref{eq: variant peeling argument} \eqref{eq: clean peeling bound}. Third, 
we modify  \eqref{eq: sup erm plays 1} (note that this is the only place we use the property of empirical risk minimization) as follows:
\begin{align}
     &\Pn (f_{\hh}+g_{\hh})\leq \frac{2}{\alpha\left(4\|\xi\|_{L_2}/\sqrt{c_{\kappa}}\right)}\Pn [\ell_{\sv}(\hh(x)-y)-\ell_{\sv}(h^*(x)-y)]
    \nonumber\\&\leq \frac{2}{\alpha\left(4\|\xi\|_{L_2}/\sqrt{c_{\kappa}}\right)}\Phi(h^*),\label{eq: regularized sup}
\end{align}
where the first inequality is due to \eqref{eq: sup empirical uppper bound} and the second inequality is due to the definition \eqref{eq: definition regularized risk minimizer} of the estimator $\hh$. After all these modifications, the inequality \eqref{eq: tempT3 4} still hold true, and the remaining proof is identical to that of Theorem \ref{thm sup}.
\hfill$\square$

\end{document}